\documentclass[acmsmall,authorversion]{acmartmodified}

\usepackage{graphicx}
\usepackage{listings}
\usepackage{multirow}
\usepackage{xcolor}
\usepackage{amsmath}
\usepackage{wrapfig}

\usepackage{xltabular}
\usepackage{caption}

\begin{document}

\title{A Review of Symbolic, Subsymbolic and Hybrid Methods for Sequential Decision Making}

\author{Carlos Núñez-Molina}
\orcid{0000-0003-1450-7323}
\email{ccaarlos@ugr.es}

\author{Pablo Mesejo}
%\affiliation{\institution{University of Granada} \department{Department of Computer Science and AI} \streetaddress{C. Periodista Daniel Saucedo Aranda, s/n} \postcode{18014} \city{Granada} \country{Spain}}
\email{pmesejo@ugr.es}

\author{Juan Fernández-Olivares}
%\affiliation{\institution{University of Granada} \department{Department of Computer Science and AI} \streetaddress{C. Periodista Daniel Saucedo Aranda, s/n} \postcode{18014} \city{Granada} \country{Spain}}
\email{faro@decsai.ugr.es}

\affiliation{\institution{University of Granada} \department{Department of Computer Science and AI} \streetaddress{C. Periodista Daniel Saucedo Aranda, s/n} \postcode{18014} \city{Granada} \country{Spain}}


\begin{abstract}
% NOTA: EN EL ACM AUTHOR GUIDELINES SE DICE EXPLÍCITAMENTE QUE EL ABSTRACT NO DEBERÍA TENER MÁS DE 100 PALABRAS.

% New abstract
{In the field of Sequential Decision Making (SDM), two paradigms have historically vied for supremacy: Automated Planning (AP) and Reinforcement Learning (RL). In the spirit of reconciliation, this paper reviews AP, RL and hybrid methods (e.g., novel learn to plan techniques) for solving Sequential Decision Processes (SDPs), focusing on their knowledge representation: symbolic, subsymbolic or a combination. Additionally, it also covers methods for learning the SDP structure. Finally, we compare the advantages and drawbacks of the existing methods and conclude that neurosymbolic AI poses a promising approach for SDM, since it combines AP and RL with a hybrid knowledge representation.}

% OLD ABSTRACT
%In the field of Sequential Decision Making (SDM), two paradigms have historically vied for supremacy: Automated Planning (AP) and Reinforcement Learning (RL).
%In the spirit of reconciliation, this paper reviews symbolic (e.g., AP), subsymbolic (e.g., RL) and hybrid (e.g., neurosymbolic) methods for  solving Sequential Decision Processes and learning their structure. 
%Additionally, it compares the advantages and drawbacks of the existing approaches and discusses what properties an ideal SDM method should exhibit. Lastly, it is concluded that such method should integrate the AP and RL paradigms and that neurosymbolic AI is the current approach that most closely resembles this ideal method.

\end{abstract}


\maketitle

\section{Introduction}
\label{section:introduction}

% ---- Abbreviatures and acronyms
% SDM, MDP, POMDP, Q(s,a), V(s), $\pi$
% AP, SP, NSP, PP, RL, DRL, RRL
% DNN, CNN, ML, DL (deep learning)
% FOL

% \textit{Definición SDM y SDP. Interés de SDM. Dificultad de resolver SDPs.}
%{\color{blue}MIRAR TRABAJO https://dl.acm.org/doi/pdf/10.1145/3578938!!! PARECE QUE LAS IMÁGENES NO TIENEN QUE OCUPAR TODO EL ANCHO!!!}

Sequential Decision Making (SDM) \cite{littman1996algorithms} is the problem of solving Sequential Decision Processes (SDPs). In an SDP, an agent situated in an environment must make a series of decisions in order to complete a task or achieve a goal. These ordered decisions must be selected according to some optimality criteria, generally formulated as the maximization of reward or the minimization of cost. SDPs provide a general framework which has been successfully applied to solve problems in fields as diverse as robotics \cite{kober2013reinforcement}, logistics \cite{schapers2018asp}, games \cite{silver2018general}, finance \cite{charpentier2021reinforcement} and natural language processing \cite{wang2018deep}, just to name a few.

% \textit{2 acercamientos "en la historia de la IA" para resolver SDPs: AP y RL. Se diferencian en cómo obtienen la solución y el tipo de knowledge representation.}

Throughout the years, many AI methods have been proposed to solve SDPs, i.e., find the sequence of decisions which optimizes the corresponding metric, such as reward or cost. They can be grouped in two main categories: Automated Planning (AP) \cite{ghallab2016automated}  and Reinforcement Learning (RL) \cite{sutton2018reinforcement}. {These two paradigms mainly differ in how they obtain a solution:}
%These two paradigms mainly differ in how they obtain a solution and how they represent their knowledge:

% \textit{Definición AP, haciendo hincapié en cómo obtiene la solución (plan) y el tipo de knowledge representation (explicar diferencia SP vs NSP). Hablar de que AP necesita conocer las dinámicas del entorno (definir action model/planning domain para cuando se mencione más adelante en la introducción).}

% \textit{Definición RL, haciendo hincapié en cómo obtiene la solución (learn from data) y el tipo de knowledge representation (subsymbolic). Hablar también de Deep RL.}

\begin{itemize}
    \item 
    %AP techniques exploit the existing prior knowledge about the environment dynamics, encoded in what is known as the action model or planning domain, to carry out a search and reasoning process in order to find a valid plan, i.e., a sequence of actions that achieves a set of goals starting from an initial state. They can be further grouped up according to the type of knowledge representation employed. Most of them require a symbolic description of the action model, in a formal language based on first-order logic (FOL) such as PDDL \cite{haslum2019introduction}. Other AP techniques do not have such requirement and can represent the action model in a subsymbolic way, usually as a \textit{black-box} which outputs the next state resulting from the application of a given action at the current state. We will refer to the first group of techniques as Symbolic Planning (SP), and use the name Subsymbolic/Non-Symbolic Planning (NSP) for the latter.

    % We now talk about the general case of prob. planning (non-deterministic, no init state) instead of classical planning. CP can be seen as a subset of PP.
    AP techniques exploit the existing prior knowledge about the environment dynamics, encoded in what is known as the action model or planning domain, to carry out a search and reasoning process in order to find a valid plan or policy, i.e., a mapping from states to actions used to achieve a set of goals.
    They can be grouped up according to the type of knowledge representation employed. Many of them require a symbolic description of the action model in a formal language often based on first-order logic (FOL), such as {PDDL \cite{haslum2019introduction} (for deterministic tasks) or PPDDL \cite{younes2004ppddl1} and RDDL \cite{sanner2010relational} (for stochastic tasks)}.
    Other AP techniques do not have such requirement and can represent the action model in a subsymbolic way, usually as a \textit{black-box} which outputs the next state resulting from the application of a given action at the current state. We will refer to the first group of techniques as Symbolic Planning (SP), and use the name Subsymbolic/Non-Symbolic Planning (NSP) for the latter.

    \item 
    % Optimal control/Control Theory es el predecesor de RL. Control theory se centra en soluciones óptimas mientras que RL proporciona aproximaciones, tanto de forma tabular como con neural networks. Control theory se basa en dynamic programming.
    % Control theory -> precursor de RL. Centrado en soluciones óptimas basadas en dynamic programming, en las que existe un conocimiento perfecto de las dinámicas del entorno. RL en cambio, se centra en proporcionar soluciones aproximadas y en las que no se conoce las dinámicas.
    { 
    RL, on the other hand, seeks to learn the policy (mapping from states to actions) that maximizes reward, %RL, on the other hand, learns an optimal policy, i.e., a mapping from states to actions in order to maximize reward, 
    automatically from data with no planning whatsoever. The main precursor of RL is Optimal Control \cite{bertsekas2019reinforcement, sutton2018reinforcement}, a field primarily concerned with providing optimal solutions to SDPs with complete knowledge of the dynamics using dynamic programming methods. Unlike their predecessor, most RL methods, known as model-free RL, focus on obtaining approximate solutions to SDPs where the action model is unknown.} Additionally, the vast majority of RL methods represent their learned knowledge in a subsymbolic way,
    {although a subset of them, known as relational RL, use a symbolic knowledge representation}. 
    Classical RL methods employ a tabular representation, i.e., 
    for every possible state-action pair they store the corresponding policy information, usually the expected future reward. On the other hand, modern RL methods, known as Deep RL (DRL), represent their policy as a Deep Neural Network (DNN) \cite{lecun2015deep}, which allows them to generalize their learned knowledge, not needing to store information about the policy for every single state-action pair.
    
    %Standard RL methods, known as model-free RL, do not need to know the action model. Instead, they are capable of learning the optimal policy, i.e., a mapping from states to actions in order to maximize reward, automatically from data and without performing any planning process whatsoever. Additionally, the vast majority of RL methods represent their learned knowledge in a subsymbolic way. Classical RL methods often employ what is known as a tabular representation, i.e., for every possible state-action pair they store the corresponding policy information, usually the expected future reward. On the other hand, modern RL methods, known as Deep RL (DRL), represent the policy as a Deep Neural Network (DNN) \cite{lecun2015deep}, which allows them to generalize their learned knowledge, not needing to store information about the policy for every single state-action pair.
\end{itemize}

% \textit{Volver a hablar de la diferencia entre SP y RL (aunque el goal de ambos campos es el mismo (resolver SDPs), cada campo ha seguido históricamente un camino diferente...). Hablar de la división entre subsymbolic y symbolic AI en la historia de la IA. Para ello, hacer un paralelismo entre la división entre RL/SP y esta división más general (e.g., "This division between RL and SP is a particular instance of a larger discussion which has taken place throghout the history of AI ... Symbolic AI vs Subsymbolic AI (Deep Learning)". Resumir ventajas y desventajas de symbolic vs subsymbolic AI (en general) y después señalar que estas ventajas/desventajas también se cumplen en el caso del RL y SP (symbolic AI tiene el problema de que requiere mucho conocimiento a priori (lo mismo sucede con SP, al ser una instancia específica de symbolic AI para SDM), subsymbolic AI como DL tiene el problema del data-inefficiency, poor-generalization y falta de interpretabilidad (igual sucede con el (Deep) RL moderno, al ser una instancia de Deep Learning para SDM)). Hablar de que muchos autores han abogado por una reconciliación de symbolic y subsymbolic AI.}

\begin{wrapfigure}{r}{0.5\textwidth}
	\centering
	\includegraphics[width=0.9\linewidth]{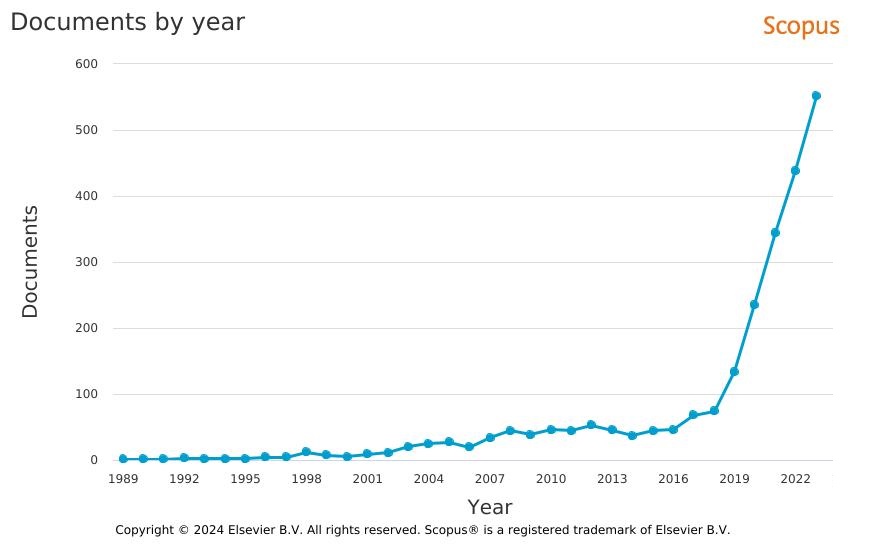}
	\caption{\textbf{Number of publications that integrate AP and RL.} This figure was obtained by introducing the following query in Scopus (search performed on {February 5 2024}): \textit{TITLE-ABS-KEY ( ( "reinforcement learning"  AND  "automated planning" )  OR  ( "model-based reinforcement learning"  OR  "model-based RL" )  OR  ( "relational reinforcement learning"  OR  "relational RL" )  OR  ( "automated planning"  AND  ( "machine learning"  OR  "deep learning" ) )  OR  ( "learn to plan"  OR  "learning to plan" )  OR  ( neurosymbolic  OR  neuro-symbolic  OR  neuralsymbolic  OR  neural-symbolic ) )  AND  ( LIMIT-TO ( SUBJAREA ,  "COMP" )  OR  LIMIT-TO ( SUBJAREA ,  "MATH" )  OR  LIMIT-TO ( SUBJAREA ,  "ENGI" ) ) {AND PUBYEAR > 1979 AND PUBYEAR < 2024} }. © Elsevier B.V.}
	\label{fig:publications_plot}
\end{wrapfigure}

Although RL and AP share the same goal of solving SDPs,
they have historically followed separate paths. This division between RL and AP represents a particular instance of a larger discussion which has taken place throughout the history of AI, confronting two approaches: symbolic and subsymbolic AI. The symbolic approach states that symbol manipulation constitutes an essential part of intelligence \cite{russell2020artificial} and was the predominant view during the majority of the 20th century. In the 21th century however, this claim has been highly contested due to the great success of Machine Learning (ML), specifically Deep Learning (DL) \cite{lecun2015deep}, in many real-world problems such as natural language processing \cite{brown2020language} and computer vision \cite{krizhevsky2017imagenet}, while performing no symbol manipulation whatsoever. Nevertheless, despite the recent successes of subsymbolic AI, current DL methods (including DRL) possess many limitations, mainly their data-inefficiency, poor generalization and lack of interpretability \cite{garnelo2016towards, marcus2018deep, battaglia2018relational}. Since these shortcomings of DL align with the strengths of symbolic AI, several authors have advocated a reconciliation of symbolic and subsymbolic AI \cite{garnelo2019reconciling, marcus2018deep}.

% \textit{Reconciliación de RL y SP (In an analogous way to the attempts to reunify symbolic and subsymbolic AI, many works have tried to bridge the gap between RL and SP...). \textbf{Motivar el interés en estos métodos que integran RL y SP haciendo uso del histograma de Scopus de papers en el campo.} Solo mencionar estos acercamientos (model-based RL, relational RL, uso de DL y Deep RL para aprender información para SP (planning policies/heuristics y action models), etc.). Hacer énfasis en los "novel" neurosymbolic models y definirlos.}

In an analogous way, many works have pursued this unification for the specific case of SDM. As a result, there have been many proposals which try to bridge the gap between RL and AP, with a surge of interest in recent years (see Figure \ref{fig:publications_plot}). A few notable examples are: model-based RL \cite{moerland2023model}, relational RL \cite{tadepalli2004relational}, ML and DL methods for learning the prior knowledge of AP (e.g., planning heuristics and action models) \cite{jimenez2012review}, models which \textit{learn to plan} \cite{moerland2023model}, and novel neurosymbolic methods \cite{garcez2022neural} which combine DNNs with the symbolic representations typically used in AP.

% \textit{Mencionar otras reviews (decir que todas se centran en algún campo específico pero ninguna trata los métodos de SP, RL y los enfoques que hibridan ambos campos de manera unificada). Enlazar con el main contribution del trabajo: la primera review (to the best of our knowledge) que trata métodos simbólicos y subsimbólicos, no solo para resolver SDPs, sino también para aprender la estructura del MDP (que muchos métodos como AP, model-based RL necesitan y otra información extra como landmarks que facilita la resolución del SDP). Decir que limitamos la review a finite (PO)MDPs.}

There exist many examples in the literature that study the application of AI to the problem of SDM from different perspectives.
\cite{ghallab2016automated} summarizes the field of AP,
{
whereas \cite{natarajan2022planning} focuses on Probabilistic Planning (PP) methods.
\cite{sutton2018reinforcement} delves into the foundations of RL and classical methods, whereas \cite{li2018deep} focuses on DRL techniques and \cite{shakya2023reinforcement} reviews both classical RL and DRL.
}
%\cite{ghallab2016automated} summarizes the field of AP whereas \cite{sutton2018reinforcement} does the same for the field of RL and \cite{li2018deep} overviews the subfield of DRL.
Some works focus on the application of ML methods to the field of AP. \cite{jimenez2012review} presents a review of ML methods for learning planning domains and control knowledge, e.g., planning heuristics. \cite{arora2018review} focuses solely on ML techniques for learning planning domains. Other works survey RL methods which incorporate some characteristics of AP.  {\cite{moerland2023model, plaat2023high}} review model-based RL, i.e., RL techniques which harness a model of the environment, \cite{tadepalli2004relational} reviews the field of relational RL, i.e., RL methods which encode their knowledge in a symbolic way,
{and \cite{acharya2023neurosymbolic} surveys neurosymbolic RL techniques.}
Finally, some works focus on the actual integration of AP and RL. \cite{partalas2008reinforcement} surveys methods which combine RL and AP to solve SDPs, but does not cover methods for learning the structure of SDPs such as the action model. \cite{moerland2020framework} presents a framework which unifies RL and AP approaches to solve SDPs. However, as the authors explicitly state, their work leaves out AP techniques which employ a symbolic knowledge representation, i.e., SP methods.

The main contribution of this work is to provide a review of symbolic, subsymbolic and hybrid AI methods for SDM. We propose a novel taxonomy which classifies these techniques according to their purpose, i.e., methods which solve SDPs versus methods which learn the structure of SDPs (e.g., the action model used in AP). 
{Although this connection has been recognized informally in a number of places}, to the best of our knowledge this is the first review which tries to cover all these different approaches and applications as part of the same work. { We intend this review as an opportunity for researchers from both the AP and RL communities to learn from each other, in an effort to unify both fields}. Due to the broad scope of our work, we limit this review to discrete SDPs, which can be formulated as finite Markov Decision Processes (MDPs) \cite{sutton2018reinforcement} and Partially Observable Markov Decision Processes (POMDPs) \cite{lovejoy1991survey}, and leave out SDPs with continuous state and/or action spaces.  Moreover, we do not focus on a specific approach or technique but rather present an overview of the most important methods present in the literature. Throughout this work, we reference several reviews for those readers who would wish to deepen into a specific topic.

% \textit{Decir que, además, esta review es "part position paper" ya que en la sección 5 discutimos las propiedades de un método ideal para resolver SDPs, como consecuencia proponemos un framework teórico que permite analizar los distintos métodos presentados en la review y que, como consecuencia del análisis, argumentamos que los métodos neurosimbólicos son los que se encuentran más cerca de dicho método ideal ("and thus pose a promising approach").}

An additional contribution of this work is provided in Section \ref{section:an_argument_for_the_integration}. Here, we discuss what properties an ideal method for SDM should exhibit. Based on these properties, we then analyse the advantages and disadvantages of the existing techniques for solving SDPs. As a result of this analysis, we argue that an ideal method for SDM should 
{unify the AP and RL paradigms for obtaining a solution of the SDP and combine the symbolic and subsymbolic knowledge representations.
Since neurosymbolic AI is the current approach which most closely resembles this ideal integration, we conclude it poses a very promising line of work.}
%unify the symbolic and subsymbolic paradigms represented by AP and RL, respectively, 
%and that neurosymbolic AI is the current approach which most closely resembles this ideal integration, thus posing a very promising line of work.

This review is organized as follows. Section \ref{section:problem_formulation} presents a formal description of SDPs as (PO)MDPs, used throughout the rest of the review. Sections \ref{section:methods_to_solve_MDPs} and \ref{section:methods_to_learn_the_structure} provide the taxonomy of methods for SDM, the first focusing on methods for solving MDPs whereas the latter focuses on methods for learning the structure of MDPs. Section \ref{section:an_argument_for_the_integration} discusses the characteristics of an ideal method for SDM, as mentioned above. Section \ref{section:future_directions} proposes several directions for future work, based on the integration of symbolic and subsymbolic AI for SDM. Finally, Section \ref{section:conclusions} presents the conclusions of our work.

\begin{table}[h]
\caption{{\textbf{Main acronyms used throughout the paper.} Acronyms appear in bold, next to their definition.}}
\centering
\resizebox{\textwidth}{!}{
\begin{tabular}{|c|c|c|c|}
\hline
\textbf{AP} & Automated Planning & \textbf{NSP} & Non-Symbolic/Subsymbolic Planning\\ \hline
\textbf{CNN} & Convolutional Neural Network & \textbf{POMDP} & Partially Observable MDP \\ \hline
\textbf{CP} & Classical Planning & \textbf{PP} & Probabilistic Planning  \\ \hline
\textbf{DL} & Deep Learning & \textbf{RL} & Reinforcement Learning \\ \hline
\textbf{DNN} & Deep Neural Network & \textbf{RRL} & Relational RL  \\ \hline
\textbf{DRL} & Deep RL &  \textbf{SDM} & Sequential Decision Making \\ \hline
\textbf{FOL} & First-Order Logic &  \textbf{SDP} & Sequential Decision Process \\ \hline
\textbf{MDP} & Markov Decision Process & \textbf{SP} & Symbolic Planning  \\ \hline
\textbf{ML} & Machine Learning &   \textbf{SSP MDP} & Stochastic Shortest Path MDP\\ \hline

\end{tabular}
}
\label{table:acronyms}
\end{table}

\section{Problem Formulation}
\label{section:problem_formulation}

%\textbf{\color{blue} ESTA SECCIÓN ES CASI ENTERAMENTE NUEVA}

In this work, we focus on SDPs with a finite number of states and actions (decisions) and where time is discrete, i.e., after the execution of an action the environment immediately transitions from time instant $t$ to $t+1$. For totally observable environments where the agent has access to full information about the current state, this type of SDPs are commonly described as a finite Markov Decision Process (MDP). However, there exist different alternative formulations for MDPs. In this work, we will use the one given by finite Stochastic Shortest-Path MDPs (SSP MDPs) \cite{natarajan2022planning}, as they provide a general MDP formulation which suits both AP and RL. An SSP MDP is constituted by the following elements:

%In RL, SDPs are usually formulated as reward-based MDPs \cite{sutton2018reinforcement}, where the goal is to maximize the (possibly discounted) sum of rewards over a finite or infinite number of time steps. On the other hand, AP often employs a goal-based MDP \cite{teichteil2011extending} formulation in terms of goals and cost instead of reward. Here, the aim of the agent is to minimize the cost needed to achieve the MDP goals.
%Given these different formulations, we may wonder if there exists some general MDP description which suits both AP and RL, in addition to hybrid methods such as model-based RL. The answer is yes, and this description is given by finite Stochastic Shortest-Path MDPs (SSP MDPs) \cite{natarajan2022planning}. An SSP MDP is constituted by the following elements:

\begin{itemize}
    \item \textbf{State space $S$}, the finite set of states of the system. In some SSP MDPs, at the beginning of the task the agent always starts from a state $s$ randomly sampled from a set of initial states $S_i \subset S$. In other SSP MDPs, the agent may start from any state $s \in S$, i.e., $S_i = S$.
    \item \textbf{Action space $A$}, the finite set of actions the agent can execute. In some SSP MDPs, only a subset of applicable actions $App(s) \subset A$ are available to the agent at a given state $s$. In other SSP MDPs, the agent can execute every action at every state, i.e., $App(s) = A \ \forall s \in S$.
    \item \textbf{Transition function $T: S \times A \times S \rightarrow [0,1]$}. It describes the dynamics of the environment, by specifying the probability $T(s,a,s')=P(s'|s,a)$ of the environment (SSP MDP) transitioning into state $s'$ after the agent executes an (applicable) action $a \in App(s)$ at the current state $s$. If given some state $s \in S$ and action $a \in App(s)$ the environment always transitions into the same state $s'$ (i.e., $P(s'|s,a)=1$ and $P(s''|s,a)=0 \ \forall s'' \neq s'$), the MDP is said to be deterministic. Otherwise, it is stochastic.
    \item \textbf{Cost function $C: S \times A \times S \rightarrow [0,\infty)$}. It gives a finite strictly positive cost $C(s, a, s') > 0$ when the agent goes from state $s$ to $s'$ by executing the (applicable) action $a$. Transitions from a goal state are the only ones with a cost of zero.
    \item \textbf{Goal set $G \subseteq S$}. It contains a finite set of goal states $s_g \in S$, one of which must be reached by the agent. For every goal state $s_g \in G$, action $a \in A$ and non-goal state $s \notin G$, the following conditions are met: $T(s_g, a, s_g)=1$, $T(s_g, a, s)=0$, $C(s_g, a, s_g)=0$. Intuitively, these conditions mean that goal states are terminal, since once reached the agent cannot leave them and no longer incurs in additional costs.
\end{itemize}

% OLD
%A policy $\pi: S \times A \rightarrow [0,1]$ is a partial function which maps every state $s \in S$ to a probability distribution over applicable actions $a \in App(s)$. A solution of an SSP MDP is an optimal policy $\pi^*$, i.e., a policy which minimizes the expected cost needed to reach a goal state $s_g \in G$ from an initial state $s_i \in S_i$. A plan $p$ is the instantiation of a policy $\pi$ starting from an initial state $s_0 \in S_i$, i.e., the sequence of states and actions $s_0, a_0, s_1, a_1, ..., s_{n-1}, a_{n-1}, s_n$ which results from applying at each state $s_j$, where $ 0 \leq j < n$, the action $a \in App(s_j)$ selected by $\pi$. Given an initial state $s_i$, if both the policy $\pi$ and MDP are deterministic, the instantiation of $\pi$ from $s_i$ produces a single plan $p$. If either the policy or MDP are stochastic, the instantiation produces a (possibly infinite) set of plans.

\begin{wrapfigure}{r}{0.31\textwidth}
	\centering
	\includegraphics[width=0.85\linewidth]{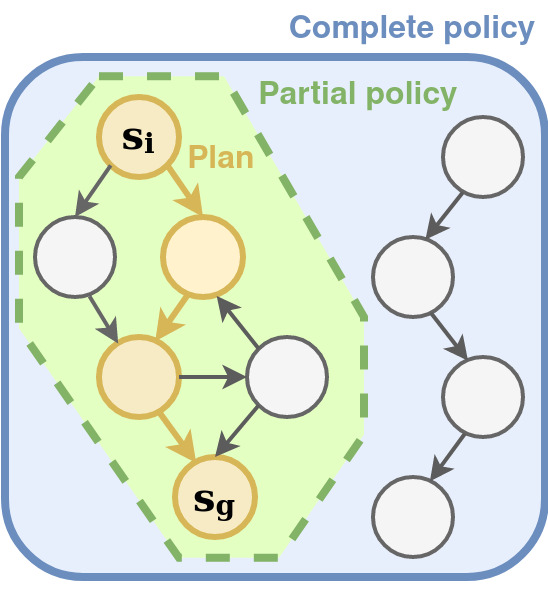}
	\caption{{\textbf{Comparison between complete policies, partial policies and plans.} Nodes in the image represent MDP states, and arrows show the possible transitions between them. A complete policy is defined for the entire MDP state space $S$ (blue square in the picture). A partial policy is defined for a subset $S' \subset S$ of states (dashed green area in the picture). Finally, a plan only stores the action to execute for the states $S'' \subset S'$ of a single trajectory from $s_i$ to some $s_g$ (gold-colored nodes with bold emphasis in the picture).}}
	\label{fig:comparison_policies}
\end{wrapfigure}

% Policy definition
A policy $\pi: S \times A \rightarrow [0,1]$ is a (possibly partial) function that maps states $s \in S$ to probability distributions over applicable actions $a \in App(s)$. 
The value $V^\pi(s)$ of a state $s$ under some policy $\pi$
is equal to the total cost that we expect to obtain if, starting from $s$, we follow policy $\pi$ until $G$ is reached.
Similarly, a Q-value $Q^\pi(s,a)$ represents the expected cost obtained by first executing action $a$ at $s$ and, then, executing actions according to $\pi$ until $G$ is reached.
State values and Q-values are related by the following equation: $V^\pi(s) = \sum_{a \in App(s)} \bigl( \pi(a|s) \cdot Q^\pi(s,a) \bigr) $.
%A solution of an SSP MDP is an optimal policy $\pi^*$ that minimizes the expected cost needed to reach $G$ from every $s \in S$, i.e., a policy whose value function $V^{\pi^*}(s)$ is minimal for every $s \in S$.
A solution of an SSP MDP is an optimal policy $\pi^*$, i.e., a policy that minimizes the expected cost needed to reach $G$.
If $\pi^*$ is optimal, then its value function $V^*$ satisfies the following property:
$V^*(s) \leq V^\pi(s) \ \forall s, \pi$.
A policy is said to be \textit{complete} if it is defined for every MDP state $s \in S$.
However, for MDPs with a single initial state $s_i$ it is often more efficient to only compute a policy $\pi$ for some subset of states $S' \subset S$, e.g., those reachable from $s_i$ by following $\pi$. Such a policy is deemed \textit{partial}.
Finally, if both a policy $\pi$ and MDP are deterministic and the MDP contains a single initial state $s_i$, then $\pi$ can be simply represented as a plan, i.e., as a sequence of actions $a_0, a_1, ..., a_n$ that reach $G$ when executed from $s_i$. This can be done because, in the deterministic setting, the execution of $\pi$ from some state $s_0$ always results in the same sequence (trajectory) of states and actions $s_0, a_0, s_1, a_1, ..., s_{n-1}, a_{n-1}, s_n$.
See Figure \ref{fig:comparison_policies} for a graphical comparison between the three different types of policies.

We previously stated SSP MDPs are suitable for both AP and RL. However, in most RL tasks the goal is to find a policy $\pi^*$ which maximizes the sum of rewards $r$, instead of minimizing the cost needed to reach a goal state $s_g \in G$. In finite-horizon (FH) MDPs, the sum of rewards must be maximized over a finite number of time steps. In infinite-horizon (IFH) MDPs on the other hand, the goal is to maximize the sum of rewards, discounted by a factor $\gamma$, over an infinite number of time steps. It can be proven that both types of reward-based MDPs, FH and IFH MDPs, can be expressed as an equivalent SSP MDP, in terms of goals and costs instead of rewards \cite{natarajan2022planning}. Therefore, the reward-based MDPs usually employed in RL are merely subclasses of the more general SSP MDPs. For this reason, in this work we will interchangeably employ the reward-based formulation of RL and the goal-based formulation of AP, knowing that both of them can be expressed as SSP MDPs.

Finally, all types of MDPs exhibit a very important feature, known as the \textit{Markov property}. This property states that, in any MDP, the cost $C$ (or reward $R$) and transition function $T$ only depend on the current state $s$ of the MDP and the action $a$ executed by the agent at that state. This property allows agents to select actions by only considering the information about their current state and not their past history, i.e., past states and actions. If the environment is partially observable, i.e., the agent lacks information about the current state, the SDP must be described as a POMDP \cite{lovejoy1991survey}. POMDPs share the same formulation as (totally observable) MDPs and also follow the Markov property. However, the current state of POMDPs is hidden and the agent only receives partial information in the form of observations about the state. 
These observations must then be used by the agent to infer the actual state of the environment and solve the POMDP.

\section{Methods to Solve MDPs}
\label{section:methods_to_solve_MDPs}
%\textit{\textbf{Num max. de páginas: 7}}

\begin{table}[h]
\centering
\caption{
{\textbf{Comparison among methods to solve MDPs}.
% We classify methods in Section \ref{section:methods_to_solve_MDPs} according to several criteria.
{The first column groups the methods} in Section \ref{section:methods_to_solve_MDPs} according to their knowledge representation: symbolic, subsymbolic or a combination (\textit{hybrid}).
%\textbf{Knowledge Representation (Rep.)}: symbolic, subsymbolic or a combination (\textit{hybrid}).
\textbf{Solution Representation (Sol.)}:
complete policy (\textit{C}), partial policy (\textit{P}) or plan.
\textbf{Action Model (Model)}: whether the method uses an action model ($\checkmark$), learns it from data ($\bigcirc$) or does not need it ($\times$).
\textbf{Stochasticity (Stoch.)}: can the method tackle stochastic MDPs?
\textbf{Heuristic (Heur.)}: whether the method leverages prior control knowledge in the form of a heuristic or not.
\textbf{Generalizability (Gen.)}: can the policy obtained by the method generalize to novel states and/or goals?
In principle, methods that utilize DNNs learn a policy applicable to any state $s \in S$ (i.e., a complete policy) and can generalize to new states and/or goals not seen during training, although the generalization capability will depend on the particular algorithm.
Among those which can generalize, methods with a symbolic or hybrid knowledge representation often do it better.}
}
\renewcommand{\arraystretch}{1.2}
\begin{tabular}{
|>{\centering\arraybackslash}m{0.4cm}|>
{\centering\arraybackslash}m{5cm}|>
{\centering\arraybackslash}m{1.1cm}|>{\centering\arraybackslash}m{1.1cm}|>{\centering\arraybackslash}m{1.1cm}|>{\centering\arraybackslash}m{1.1cm}|>{\centering\arraybackslash}m{1.1cm}|}
%\begin{tabularx}{\textwidth}{
 % |>{\centering\arraybackslash}X
%  |>{\centering\arraybackslash}X
 % |>{\centering\arraybackslash}X
 % |>{\centering\arraybackslash}X
 % |>{\centering\arraybackslash}X
 % |>{\centering\arraybackslash}X
  %|>{\centering\arraybackslash}X|
%}
\hline
& \textbf{Methods} & \textbf{Sol.} & \textbf{Model} & \textbf{Stoch.} & \textbf{Heur.} & \textbf{Gen.} \\
\hline
\multirow{14}{*}{\rotatebox[origin=c]{90}{Subsymbolic $\ \ \ $}} & PI, VI
\cite{sutton2018reinforcement} & C & $\checkmark$ & $\checkmark$ & $\times$ & $\times$ \\
\cline{2-7}
& LAO* \cite{hansen2001lao}, LRTDP \cite{bonet2003labeled} & P & $\checkmark$ & $\checkmark$ & $\checkmark$ & $\times$ \\
\cline{2-7}
& MCTS \cite{browne2012survey} & P & $\checkmark$ & $\checkmark$ & $\times$ & $\times$ \\
\cline{2-7}
& A* \cite{hart1968formal} & plan & $\checkmark$ & $\times$ & $\checkmark$ & $\times$ \\
\cline{2-7}
& DFS \cite{tarjan1972depth}, BFS \cite{bundy1984breadth} & plan & $\checkmark$ & $\times$ & $\times$ & $\times$ \\
\cline{2-7}
& Q-Learning \cite{watkins1989learning} & C & $\times$ & $\checkmark$ & $\times$ & $\times$ \\
\cline{2-7}
& Deep Q-Learning \cite{mnih2013playing}, REINFORCE \cite{williams1992simple}, A2C \cite{mnih2016asynchronous} & C & $\times$ & $\checkmark$ & $\times$ & $\checkmark$ \\
\cline{2-7}
& Dyna \cite{sutton1991dyna} & C & $\bigcirc$ & $\checkmark$ & $\times$ & $\times$ \\
\cline{2-7}
& AlphaZero \cite{silver2018general} & C & $\checkmark$ & $\checkmark$ & $\times$ & $\checkmark$ \\
\cline{2-7}
& MuZero \cite{schrittwieser2020mastering} & C & $\bigcirc$ & $\checkmark$ & $\times$ & $\checkmark$ \\
\cline{2-7}
& VIN \cite{tamar2016value} & C & $\times$ & $\checkmark$ & $\times$ & $\checkmark$ \\
\cline{2-7}
& UPN \cite{srinivas2018universal} & C & $\times$ & $\times$ & $\times$ & $\checkmark$ \\
\cline{2-7}
& MCTSnet \cite{guez2018learning} & C & $\checkmark$ & $\times$ & $\times$ & $\checkmark$ \\
\cline{2-7}
& IBP \cite{pascanu2017learning}, TreeQN \cite{farquhar2018treeqn}, DRC \cite{guez2019investigation} & C & $\bigcirc$ & $\checkmark$ & $\times$ & $\checkmark$ \\
\hline
\multirow{5}{*}{\rotatebox[origin=c]{90}{Symbolic $\ \ \ \:$}} & SPUDD \cite{hoey1999spudd} & C & $\checkmark$ & $\checkmark$ & $\times$ & $\times$ \\
\cline{2-7}
& Symbolic LAO* \cite{feng2002symbolic}, SSiPP \cite{trevizan2014depth} & P & $\checkmark$ & $\checkmark$ & $\checkmark$ & $\times$ \\
\cline{2-7}
& FF-Replan \cite{yoon2007ff} & plan & $\checkmark$ & $\checkmark$ & $\checkmark$ & $\times$ \\
\cline{2-7}
& FF \cite{hoffmann2001ff}, FD \cite{helmert2006fast} & plan & $\checkmark$ & $\times$ & $\checkmark$ & $\times$ \\
\cline{2-7}
& Relational Q-Learning \cite{dvzeroski2001relational}, Deep Symbolic Policy \cite{landajuela2021discovering} & C & $\times$ & $\checkmark$ & $\times$ & $\checkmark$ \\
\hline
\multirow{4}{*}{\rotatebox[origin=c]{90}{Hybrid $\ \ \ \;$}} & SDRL \cite{lyu2019sdrl} & C & $\checkmark$ & $\checkmark$ & $\times$ & $\checkmark$ \\
\cline{2-7}
& SORL \cite{jin2022creativity} & C & $\bigcirc$ & $\checkmark$ & $\times$ & $\checkmark$ \\
\cline{2-7}
& DRL with Relational Inductive Biases \cite{zambaldi2019deep} & C & $\times$ & $\checkmark$ & $\times$ & $\checkmark$ \\
\cline{2-7}
& ASNet \cite{toyer2018action} & C & $\checkmark$ & $\checkmark$ & $\checkmark$ & $\checkmark$ \\
\hline
\end{tabular}
\label{table:methods_to_solve_mdps}
\end{table}

\begin{figure}[h]
	\centering
	\includegraphics[width=.9\linewidth]{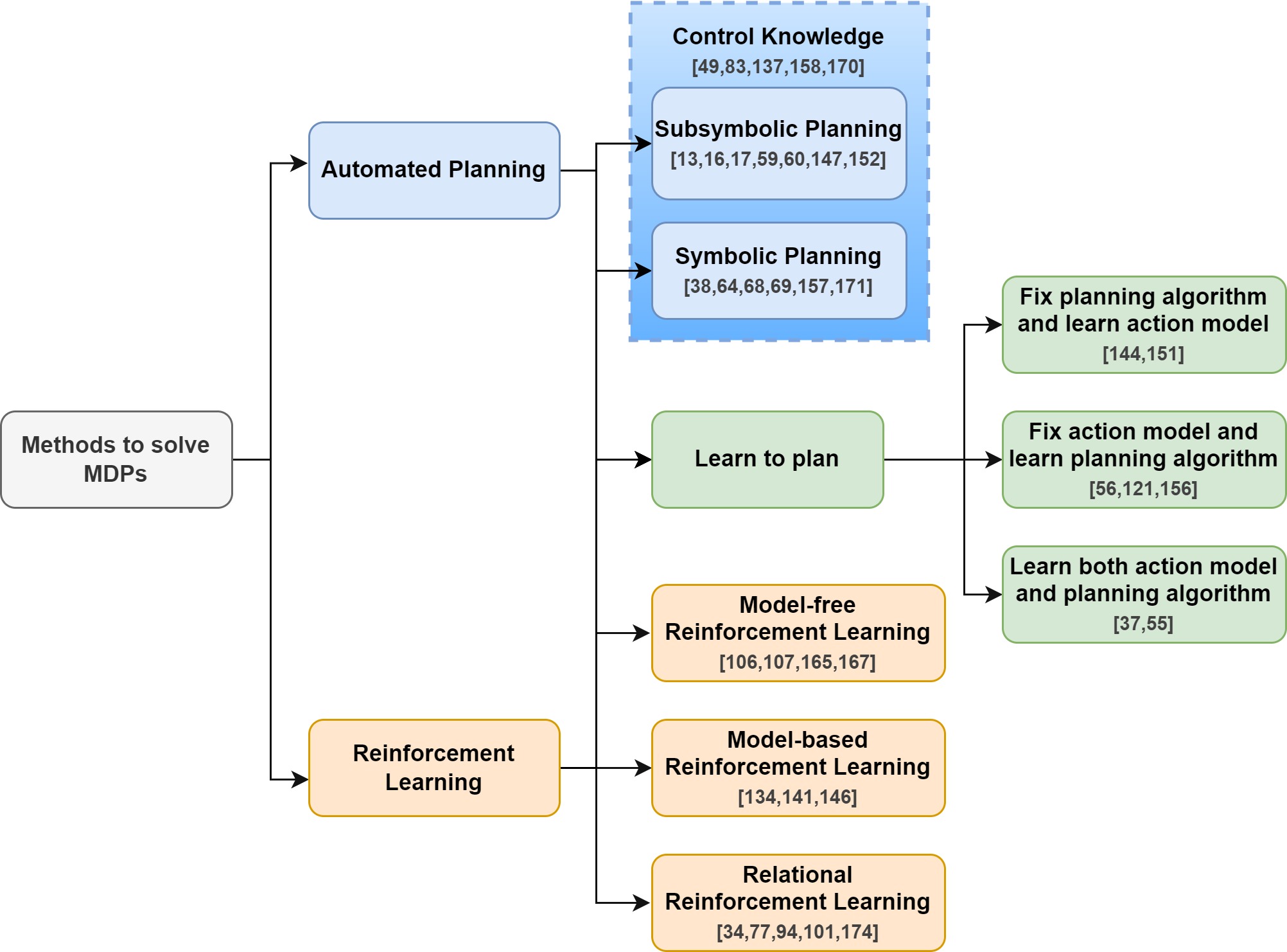}
	\caption{{\textbf{Proposed taxonomy of methods to solve MDPs.} Model-free and model-based RL methods with a symbolic or hybrid knowledge representation are placed in the \textit{Relational Reinforcement Learning} category.}}
	\label{fig:diagram_solve_MDP}
\end{figure}

In this section we present an overview of the main methods to solve MDPs (see Figure \ref{fig:diagram_solve_MDP}). Historically, there have been two competing paradigms. AP \cite{ghallab2016automated} proposes a synthesis-based approach, in which prior information about the MDP is used to carry out a search and reasoning process in order to find its solution. RL \cite{sutton2018reinforcement}, on the other hand, presents a learning-based approach inspired by ML, where the agent does not synthesize an MDP solution but rather learns it from data. In this work, we also discuss a novel family of methods that \textit{learn to plan} \cite{moerland2023model},
thus combining the AP and RL approaches.
In the same manner as AP, these methods carry out a planning process to find a solution of the MDP. However, unlike them, they learn how to actually perform the planning computations automatically from data, similarly to how RL techniques learn the optimal policy also from data.
{Table \ref{table:methods_to_solve_mdps} shows an overview of the methods discussed in this section, classified according to their properties.}

\subsection{Automated Planning}
%\textit{\textbf{Num max. de páginas: 2.5}}

AP \cite{ghallab2016automated} comprises methods that harness information about the environment dynamics to synthesize a solution of the MDP. This information is encoded in the action model, also referred to as the planning domain, which represents the available actions for the agent and how they affect the world.

Some AP techniques, known as Probabilistic Planning (PP) \cite{natarajan2022planning}, solve the general class of stochastic MDPs and are used to find a policy that minimizes the expected cost needed to reach $G$ from $s_i$.
Conversely, Classical Planning (CP) \cite{ghallab2016automated} methods tackle the specific class of deterministic MDPs with {a} single initial state $s_i$ and their goal is to find a minimal-cost plan from $s_i$ to $G$.
%In this review, 
{We group AP methods according to the type of knowledge representation employed in} Subsymbolic/Non-Symbolic Planning (NSP) and Symbolic Planning (SP).
{Additionally, in Section \ref{subsection:planning_policies_and_heuristics}, we explain how both NSP and SP techniques can exploit available knowledge to solve the MDP more efficiently.}

%AP \cite{ghallab2016automated} is a subfield of AI which provides a set of deliberative techniques to solve MDPs. AP methods harness information about the environment dynamics to synthesize a solution of the MDP. This information is encoded in the action model, sometimes also referred to as the planning domain, which represents the available actions for the agent and how they affect the current state of the world. A solution of the MDP is often formulated as the policy or plan that achieves a set of goals while minimizing the cost needed to obtain them. In other cases, it is formulated as the policy that maximizes a reward function. AP methods can be split in two main categories according to the type of knowledge representation employed: Subsymbolic/Non-Symbolic Planning (NSP) and Symbolic Planning (SP).

\subsubsection{Subsymbolic Planning}

We use the name NSP to refer to those AP methods which do not require a symbolic (e.g., FOL-based) description of the MDP. NSP methods only need as action model a transition function that maps a state-action $(s,a)$ pair into its corresponding next state $s'$ and do not care about how such a function is implemented.
For this reason, these techniques are very versatile and can be applied to a wide range of situations where the environment dynamics are known but are not given in a formal, logic-based description.
%
%There exist two main paradigms in NSP: Probabilistic Planning (PP) \cite{natarajan2022planning} and Classical Planning (CP) \cite{ghallab2016automated}.
%PP techniques solve the general class of stochastic MDPs and are used to find a policy that minimizes the expected cost needed to reach $G$ from $S_i$.
%On the other hand, CP algorithms tackle the specific class of deterministic MDPs with a single initial state $s_i$ and their goal is to find a minimal-cost plan from $s_i$ to $G$.

Subsymbolic PP algorithms can be grouped according to whether they obtain complete or partial policies. 
Two foundational algorithms that find complete policies are Policy Iteration (PI) and Value Iteration (VI) \cite{sutton2018reinforcement}. 
PI is a dynamic programming algorithm that involves two steps: policy evaluation,
where the value $V^\pi(s)$ of the current policy $\pi$ is computed for every state $s \in S$, and policy improvement, where $\pi$ is updated by selecting, for each $s$, the action $a$ that optimizes $Q^\pi(s,a)$. These two steps are repeated until $\pi$ converges to the optimal policy $\pi^*$.
VI is another dynamic programming algorithm. It directly estimates the value $V^*(s)$ of $\pi^*$ by iteratively updating the current estimate of $V^*(s)$ using the Bellman Optimality Equation. Once $V^*(s)$ is computed, $\pi^*$ can be obtained by simply selecting, for each state $s$, the action $a$ with best $Q^*(s,a)$.
Some PP methods that obtain partial policies leverage a heuristic (see Section \ref{subsection:planning_policies_and_heuristics}) to direct the search.
LAO* \cite{hansen2001lao} gradually expands states reachable from $s_i$, initializes their value with a heuristic and backs up this information by running VI or PI on those states whose value could have changed.
LRTDP \cite{bonet2003labeled} repeatedly samples trajectories from $s_i$ using the current policy $\pi$, initializes the value of new states with a heuristic and 
updates $\pi$ and the values of states in the trajectories using the Bellman Optimality Equation.
Unlike these methods, MCTS \cite{browne2012survey} does not require a heuristic.
This algorithm gradually builds a search tree from $s_i$,
estimating state values from the results of sampled trajectories.
MCTS was originally developed for deterministic environments,
but it has been extended to manage stochasticity.
% OLD
%Unlike the two previous methods, MCTS\footnote{Although originally deterministic, MCTS has been extended to stochastic environments.} \cite{browne2012survey} does not require a heuristic. It gradually builds a search tree from $s_i$, estimating state values from the results of sampled trajectories and balancing the exploitation of promising states with the exploration of less visited states.
%
Finally, subsymbolic CP methods can also be grouped according to the use of heuristics.
A widely-known CP algorithm that leverages heuristics is A* \cite{hart1968formal}, which can in fact be viewed as a specialization of LAO* for deterministic MDPs.
Other CP methods like depth-first search (DFS) \cite{tarjan1972depth} and breadth-first search (BFS) \cite{bundy1984breadth} explore $S$ without heuristics. DFS does so by exploring as far as possible along the current trajectory before moving to the next, whereas BFS expands all states with depth $n$ before exploring those with depth $n+1$.

\begin{wrapfigure}{r}{0.5\textwidth}
%\begin{figure}[h]
	\centering
        \includegraphics[width=0.95\linewidth]{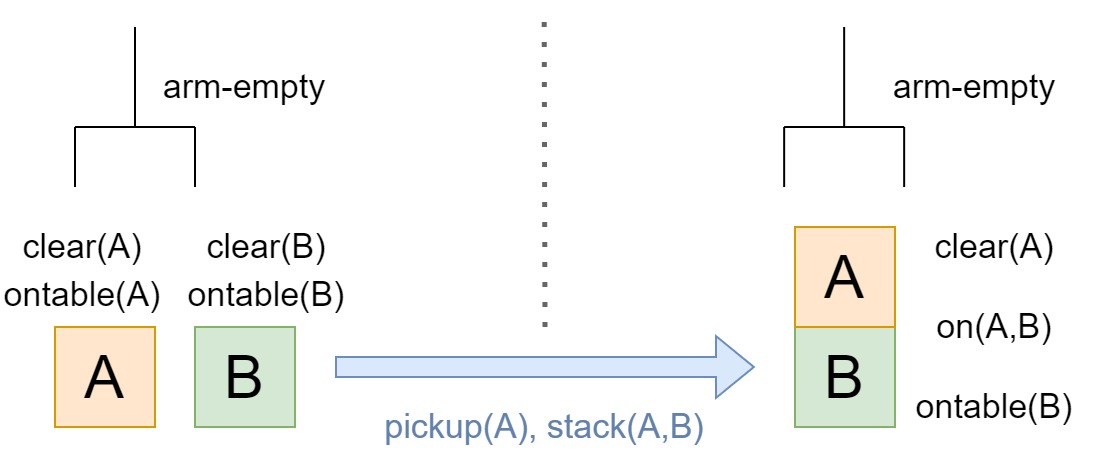}
	\caption{\textbf{CP task encoded using PDDL.} The task belongs to the PDDL domain known as \textit{blocksworld}, consisting of blocks that can be stacked one upon another with a gripper arm. The blue arrow represents a plan that achieves the goal state (right) starting from the initial state (left).}
	\label{fig:figure_blocksworld}
%\end{figure}
\end{wrapfigure}

\subsubsection{Symbolic Planning}
% Nuevo esquema

% Unlike NSP, SP methods require symbolic models. Example of languages: (P)PDDL, RDDL. Explain (P)PDDL: Probabilistic vs deterministic PDDL. Domain (MDP dynamics) vs problem (init state and goal). RDDL: explain differences with PPDDL.
% REFERENCIAR FIG 4
% Several advantages of symbolic action models: 1) reduce human effort (use same domain for different problems, these formal languages result convenient for specifying many types of MDPs) 2) leverage symbolic description to obtain heuristics (see Section 3.1.3) (unlike NSP).

% Methods
% Algebraic Decision Diagrams: Symbolic LAO* and SPUDD
% Probabilistic planners: SSiPP and FF-Replan
% Classical planners: FF, FD, LAMA
% IPC: periodic competitions for comparing classical and probabilistic planners on a set of (P)PDDL/RDDL domains serving as benchmarks.

% Mencionar extensiones: generalized planning, temporal planning, etc.? -> Creo que no porque no aporta mucho y así me puedo ahorrar referencias (que ocupan espacio)

We use the name SP to refer to those methods which, unlike NSP, require a symbolic description of the MDP in a formal language,
such as (P)PDDL or RDDL (see Section \ref{section:task_general_action_models}).
These languages split the MDP description into two different items: a planning \textit{domain}, which encodes the MDP dynamics, and a planning \textit{problem}, which describes the particular objects in the MDP, its initial state and goal(s) to achieve. 
%A common choice for modeling MDPs is to use PDDL, for CP tasks, and Probabilistic PDDL (or PPDDL) \cite{younes2004ppddl1}, for PP tasks.
%
%(P)PDDL is a declarative, FOL-based language that models planning tasks using two different items: a planning \textit{domain}, which encodes the MDP dynamics, and a planning \textit{problem}, which encodes the particular initial state and goal(s) to achieve. 
Figure \ref{fig:figure_blocksworld} shows an example CP task encoded in PDDL.
%
%RDDL \cite{sanner2010relational} is another language that also models tasks as domain-problem pairs. In RDDL, everything (including actions) is represented as a parameterized variable of a particular type. This modeling approach is better suited than PPDDL for domains where actions have many uncorrelated effects, as often occurs in systems with many objects that mostly evolve independently from each other.
%
Symbolic MDP descriptions provide several advantages over subsymbolic ones. Firstly, languages such as (P)PDDL and RDDL make it possible to specify a wide variety of MDPs with ease.
Secondly, these symbolic descriptions can be exploited by SP algorithms to speed up planning.

% Methods
% Algebraic Decision Diagrams: Symbolic LAO* and SPUDD
% Probabilistic planners: SSiPP and FF-Replan
% Classical planners: FF, FD, LAMA
% IPC: periodic competitions for comparing classical and probabilistic planners on a set of (P)PDDL/RDDL domains serving as benchmarks.

Many symbolic PP algorithms leverage MDP descriptions by using \textit{algebraic decision diagrams} (ADDs). 
ADDs represent MDP elements (e.g., values, transitions and costs) as functions of boolean variables (e.g., FOL atoms), using a special type of decision tree.
They encode the inner MDP structure and group similar states together (like those that share the same value), which allows for more efficient implementations of the NSP algorithms seen in the previous section.
For example, SPUDD \cite{hoey1999spudd} integrates VI with ADDs, whereas Symbolic LAO* \cite{feng2002symbolic} does so for LAO*.
Other PP techniques use symbolic descriptions to compute heuristics (see Section \ref{subsection:planning_policies_and_heuristics}).
FF-Replan \cite{yoon2007ff} solves the all-outcomes determinization (see Section \ref{subsection:planning_policies_and_heuristics}) of the MDP with the classical planner FastForward (FF) \cite{hoffmann2001ff}. Then, it executes the plan on the original, stochastic MDP until it fails, at which point FF is called again.
SSiPP \cite{trevizan2014depth} works by successively decomposing the MDP into subproblems and solving them. For each subproblem, it computes a policy that can be executed for at least $t$ time steps before a new policy is needed.
Finally, much effort has been devoted to developing planners for the symbolic CP case, such as FF and FastDownward (FD) \cite{helmert2006fast}, which mainly differ in the search algorithms and heuristics employed.
New classical and probabilistic planners are periodically compared with each other in the International Planning Competitions \cite{vallati20152014}, which evaluate planners on a set of benchmark (P)PDDL/RDDL domains.

\subsubsection{Control knowledge}
\label{subsection:planning_policies_and_heuristics}

% Decir que las domain-independent heuristics se obtienen a partir del symbolic action model

% Cost-to-go equivalent to $V(s)$

% Skip planning policies, focus on planning heuristics

% Nuevo esquema

% Planning is expensive. Complexity of planning (PP and CP).
% We need to use control knowledge. Often, represented as heuristics which estimate the cost-to-go from the current state $s$ to $G$ (equivalent to $V(s)$).

% domain-independent heuristics widely used by SP techniques: many heuristics for CP (delete-relaxation), leverage symbolic MDP description. Domain-independent heuristics for PP: leverage CP heuristics. Determinization (all-outcomes).
% domain-specific heuristics: example, manhattan distance.
% Sometimes not available, they can be learned from data. Example methods, neurosymbolic.
% Review about control knowledge.

\begin{wrapfigure}{r}{0.41\textwidth}
	\centering
        \includegraphics[width=0.625\linewidth]{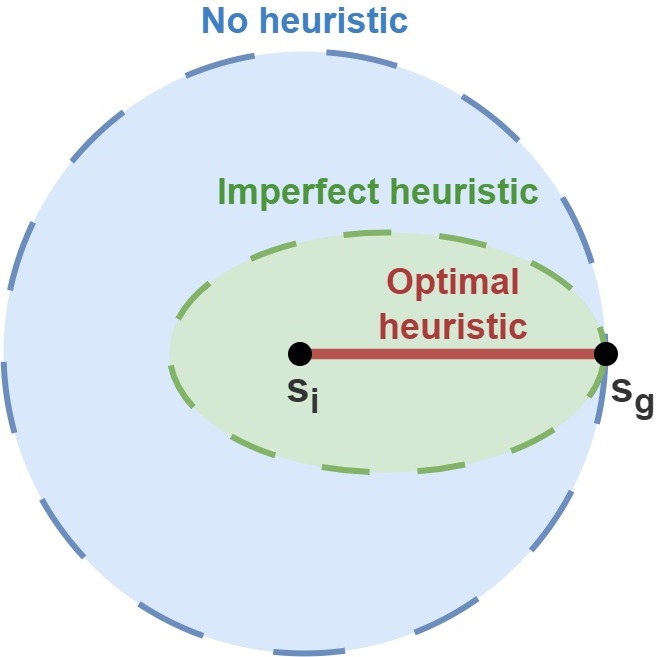}
	\caption{{\textbf{Planning with a heuristic.}
        {The figure} illustrates how heuristics help reduce planning effort.
        For simplicity, we depict the case where the MDP is deterministic and search is carried out from the initial state $s_i$ to the goal $s_g$. 
        When no heuristic is employed, the planning algorithm needs to explore the state space in all directions until $s_g$ is finally found (see blue circle in the image).  
        A heuristic can prevent this by providing guidance and 
        reducing the number of states that are explored (green ellipse in the image).
        Finally, if this heuristic is optimal/perfect (i.e., it predicts the optimal cost $V^*(s)$ for every state $s \in S$), only those states on the optimal plan(s) from $s_i$ to $s_g$ need to be explored.
        Analogously, for stochastic MDPs, the only states that need to be explored are those reachable from $s_i$ by following the optimal policy $\pi^*$.
 }}
	\label{fig:figure_heuristics}
\end{wrapfigure}

Planning is computationally expensive. Solving the general class of (stochastic) SSP MDPs (as in PP) is EXPTIME-complete, whereas solving deterministic SSP MDPs with {a} single initial state (as in CP) is PSPACE-complete \cite{natarajan2022planning}. For this reason, if we hope to apply AP methods {(from both NSP and SP)} to real-world problems{,} we need to harness \textit{control knowledge} to direct the search (see Figure \ref{fig:figure_heuristics}). This knowledge often comes in the form of planning \textit{heuristics}, which estimate the expected cost from a state $s$ to the goal $G$ (in other words, they provide an initial estimate for $V^*(s)$).

Most SP techniques exploit symbolic MDP descriptions to compute powerful domain-independent heuristics, which can be applied to any planning domain.
There exist a wide variety of domain-independent heuristics for CP. One of the most popular approaches are the so-called \textit{delete-relaxation} heuristics \cite{keyder2014improving}, which estimate the cost from the current state to the goal in a simplified version of the planning task where the \textit{delete} effects of actions are ignored.
Probabilistic planners often leverage heuristics by obtaining a deterministic version (\textit{determinization}) of the MDP being solved and computing a CP heuristic on it, which is then used as a heuristic for the original, stochastic MDP.
The most widely-used one is known as the \textit{all-outcome determinization} \cite{yoon2007ff}. Given a stochastic MDP $M$, it obtains a deterministic version $M'$ of it which contains a separate deterministic action for each probabilistic effect of every action of $M$.
Alternatively, some methods utilize domain-specific heuristics to solve a particular planning domain in a very efficient way. For instance, the L1 norm or \textit{Manhattan distance} provides a straightforward heuristic which has been successfully applied to many pathfinding problems \cite{cazenave2006optimizations}.
However, in some situations we may not have access to this type of prior knowledge. In these cases, heuristics can be learned from example trajectories, using different ML methods such as linear regression \cite{yoon2006learning}, regression trees and support vector machines \cite{us2013learning}.
Neurosymbolic methods have also been employed.
\cite{shen2020learning} applies Graph Neural Networks \cite{battaglia2018relational} to predict the heuristic value from the delete-relaxation representation of the problem, 
whereas \cite{gehring2022reinforcement} leverages domain-independent heuristics to efficiently learn domain-specific heuristics with a Neural Logic Machine \cite{dong2019neural} trained using DRL.
A more comprehensive review of methods to learn control knowledge can be found in  \cite{jimenez2012review}.

\subsection{Reinforcement Learning}
%\textit{\textbf{Num max. de páginas: 2.5}}

RL \cite{sutton2018reinforcement} is a subfield of ML that provides an alternative approach to AP for solving MDPs. 
Instead of employing an action model to synthesize a solution of the MDP, RL techniques use the data gathered from the environment to learn the optimal policy that maximizes reward. To do so, they must balance the exploration of the environment, i.e., the process of trying out new actions and observing their outcomes, with the exploitation of the learned knowledge, i.e., selecting the best action found so far (which might not be the optimal one). This is known as the \textit{exploration-exploitation tradeoff}. Most RL methods, known as model-free RL, do not need a model of the environment to learn the optimal policy. However, some methods known as model-based RL harness the information contained in the action model to facilitate the learning process. In addition, although the vast majority of RL techniques represent the information in a subsymbolic way, some of them known as relational RL use a symbolic knowledge representation analogous to that of SP.

\subsubsection{Model-free RL}

Model-free RL provides methods to learn the optimal policy when the model of the world, i.e., the action model, is unknown. These methods can be further grouped in (model-free) value-based and policy-based RL. {Value-based techniques learn the optimal state values $V^*(s)$ or Q-values $Q^*(s,a)$, which are then used to obtain the optimal policy $\pi^*$.}
%Value-based RL learns to estimate the expected cumulative reward $\mathbb{E}_{\pi^*}[R]$, referred to as the \textit{return}, associated with a state (value function $V(s)$) or a state-action pair (action-value function $Q(s,a)$). Once this function has been learned, the optimal policy simply corresponds to selecting in each state the action with the maximum return.
A classical algorithm for value-based RL is Q-Learning \cite{watkins1989learning}, 
{which can be seen as an adaptation of VI to the model-free setting where the environment dynamics are unknown.
It uses the Bellman Optimality Equation to estimate the $Q^*(s,a)$ values from the rewards and state transitions observed when executing actions in the environment.}
%Q-Learning utilizes the Bellman Optimality Equation to learn the optimal Q-value $Q(s,a)$ associated with each state-action pair $(s,a)$, i.e., the maximum return that can be obtained for each $(s,a)$ pair.
These Q-values need to be stored for every possible combination of states and actions, which results infeasible for MDPs with large state spaces. Deep Q-Learning \cite{mnih2013playing} solves this problem by using a DNN to
approximate the Q-values. Since this DNN is capable of generalizing to new states, it does not need
to memorize the Q(s, a) value for every (s, a) pair, thus making it possible to apply this algorithm
to real-world problems with large state spaces.

In contrast to value-based methods, policy-based RL explicitly learns $\pi^*$ without needing to estimate $V^*(s)$ or $Q^*(s,a)$. One of the most well-known algorithms in this category is REINFORCE \cite{williams1992simple}. REINFORCE uses a DNN to approximate the policy, i.e., given an input state it returns a probability distribution over the actions. This DNN is then trained using gradient-based methods to maximize the probability of selecting actions with a large $Q^*(s,a)$ associated. One main issue of REINFORCE is that $Q^*(s,a)$ is not always a good measure of action optimality, since it does not only depend on the action $a$ but also on the state $s$ it is executed. Advantage Actor Critic (A2C) \cite{mnih2016asynchronous} solves this problem by substituting $Q^*(s,a)$ for the advantage $A^*(s,a) = Q^*(s,a) - V^*(s)$, which measures how good $a$ is when compared to the average action at $s$. A2C 
trains two separate DNNs: the actor and the critic. The critic learns to predict the value $V^*(s)$ of a given state, which is then used to calculate $A^*(s,a)$. The actor learns $\pi^*$ by using the same method as REINFORCE. However, it utilizes the advantage $A^*(s,a)$ predicted by the critic to measure action optimality, instead of $Q^*(s,a)$. Thus, A2C entails a hybrid approach which integrates both value and policy-based RL.
A deeper view into classical RL and DRL methods can be found in 
{\cite{sutton2018reinforcement, li2018deep, shakya2023reinforcement}.
}
%A deeper view into classical RL and DRL methods can be found in \cite{sutton2018reinforcement} and \cite{li2018deep}, respectively.

\subsubsection{Model-based RL}

Model-based RL tries to combine the fields of RL and AP. It enhances standard, model-free RL algorithms with a model of the world, which can be employed in two main ways. The first alternative is to use the action model as a simulator to obtain data for training the policy, thus minimizing the required amount of interaction with the real world. For this reason, model-based RL techniques are more sample-efficient, i.e., need less data to learn the optimal policy, than model-free methods \cite{kaiser2020model}. A second alternative is to integrate a deliberative process into the decision-making cycle of RL. Instead of simply selecting the best action according to the policy, we can use the action model to carry out a planning process, guided by the learned policy/value function, in order to select the next action to execute. This approach blurs the line between model-based RL and AP with a learned heuristic. Finally, we can differentiate between model-based RL methods which require the action model to be given a priori and those which do not. This second category of methods use the data collected from the environment to learn the action model in addition to a policy, employing some of the techniques commented in Section \ref{subsection:action_model_learning}.

\begin{figure}[h]
	\centering
	\includegraphics[width=.95\linewidth]{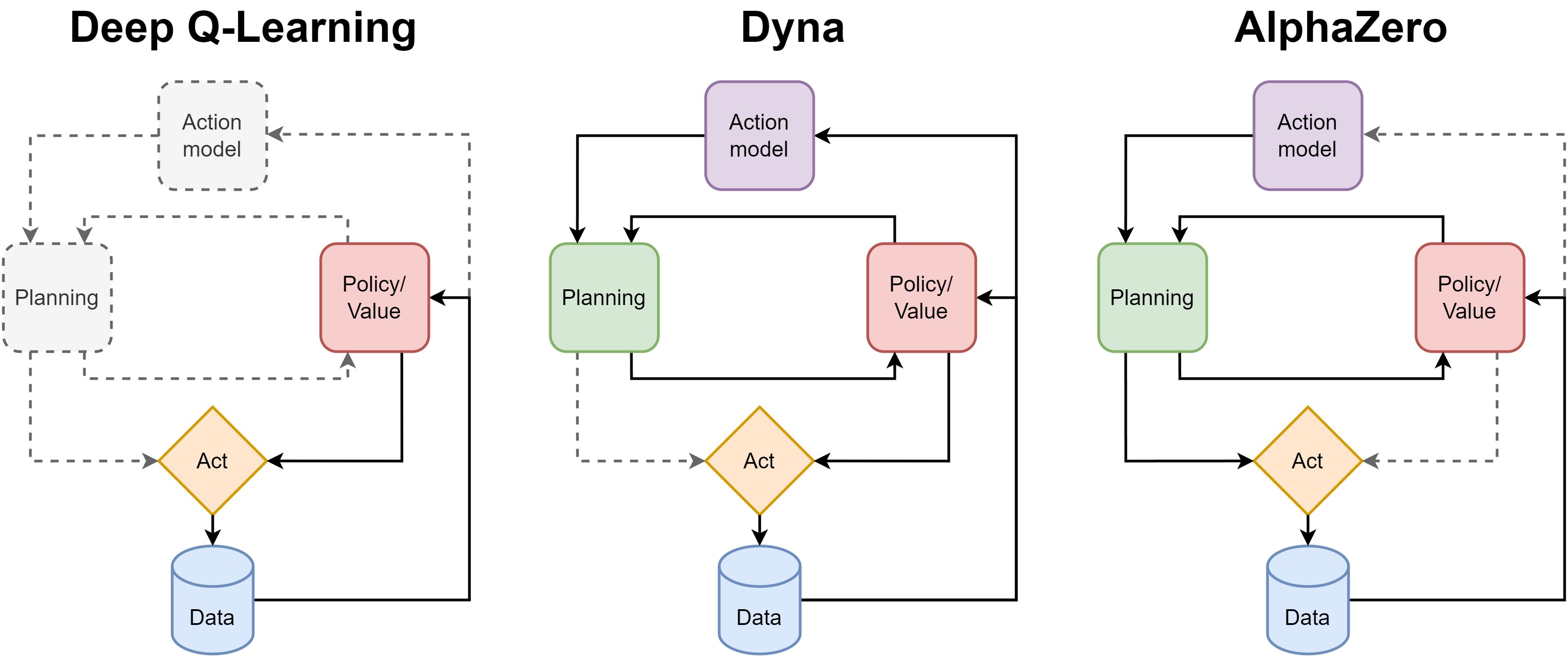}
	\caption{{\textbf{Model-free vs model-based RL.} Adapted from \cite{moerland2023model} with permission by the authors. The figure compares the architectures of a model-free RL algorithm, Deep Q-Learning, and two model-based RL methods, Dyna and AlphaZero. Thick lines and colored elements are used by the algorithm, whereas grayed out elements and dotted lines are not, and are displayed just for comparative purposes. Deep Q-Learning learns a value function from data, which is then used to select the action to execute with no planning whatsoever.
    On the other hand, Dyna utilizes data to learn both a value function and action model. This model is used to obtain extra data to train the value function, which ultimately decides the action to execute.
    AlphaZero utilizes data to train both a value function and a policy. These are used to guide a planning process over an action model provided in advance, in order to decide the action to execute.
 %It can be observed that Deep Q-Learning does not use an action model or perform any planning whatsoever. Conversely, AlphaZero uses an action model (given a priori) to perform a planning process (guided by a policy and value function) that selects the next action to execute.
 }}
	\label{fig:figure_comparison_model-free_model-based_RL}
\end{figure}

We will use Dyna \cite{sutton1991dyna} and AlphaZero \cite{silver2018general} as illustrative examples of the different existing model-based RL algorithms. One of the oldest examples of model-based RL can be found in Dyna. This method learns a subsymbolic, black-box model of the world. Then, it trains the Q-Learning algorithm on both experience obtained from the real world and data sampled using the learned action model. Dyna performs reactive execution, i.e., the action to execute is selected according to the value function trained with Q-Learning, with no planning involved whatsoever. AlphaZero is a novel model-based RL method which has been successfully applied to play the games of chess, shogi and Go at superhuman level. Unlike Dyna, it requires a prior model of the world, although a newer version of this method known as MuZero \cite{schrittwieser2020mastering} overcomes this limitation. AlphaZero trains both a value function and a policy, implemented as a single deep Convolutional Neural Network (CNN) \cite{krizhevsky2017imagenet}, which together guide the planning process performed by the MCTS algorithm. This planning process outputs a probability distribution from which the action to execute is sampled. The value function is trained to predict the game winner from the current state whereas the policy is trained to match the probabilities obtained by MCTS. Thus, in AlphaZero there exists a clear synergy between RL and AP: RL is used as a heuristic to guide the planning process, which in return allows to obtain data to train the RL policy, acting as a \textit{policy improvement operator}. A comparison between Deep Q-Learning, Dyna and AlphaZero is shown in Figure \ref{fig:figure_comparison_model-free_model-based_RL}. Finally, a more comprehensive review of model-based RL is provided in {\cite{moerland2023model, plaat2023high}}.

\subsubsection{Relational RL}

Relational Reinforcement Learning (RRL) \cite{tadepalli2004relational} can be considered the intersection of RL and Relational Learning \cite{koller2007introduction}. It is comprised of methods that combine RL with the symbolic representations employed in SP. These symbolic representations are well suited for MDPs that can be naturally described in terms of objects and their interactions, known as object-oriented MDPs \cite{diuk2008object}. 
RRL methods often leverage lifted representations, i.e., symbolic representations with variables, to abstract from concrete objects and, thus, naturally generalize to tasks with varying number of objects. In addition, the knowledge learned by RRL techniques is more amenable to interpretation than the one typically learned in DRL.

\begin{wrapfigure}{r}{0.37\textwidth}
%\begin{figure}[h]
	\centering
        \includegraphics[width=0.8\linewidth]{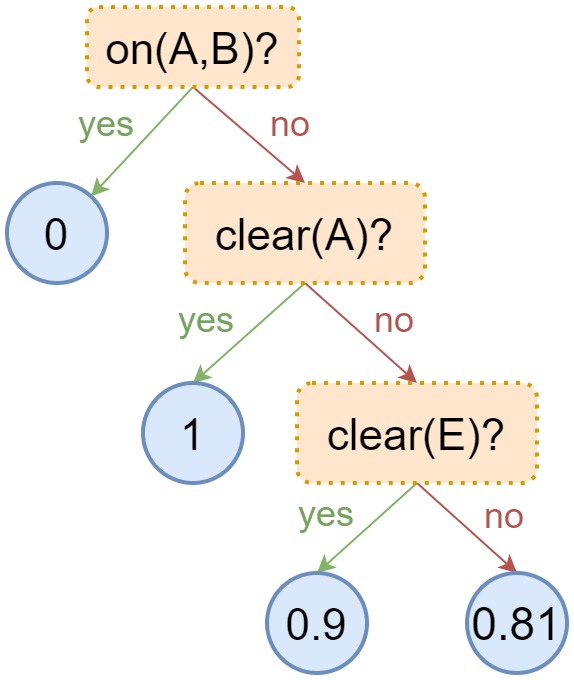}
	\caption{\textbf{Regression tree learned with relational Q-Learning} (adapted from \cite{dvzeroski2001relational}). The example corresponds to the blocksworld domain (see Figure \ref{fig:figure_blocksworld}). The tree encodes the $Q^*(s,a,g)$ value for the goal $g=on(A,B)$ and action $a=stack(D,E)$. Inner nodes (orange) check the truth value of a grounded predicate of the state $s$, whereas leaf nodes (blue) return the corresponding Q-value.}
	\label{fig:figure_relational_RL}
%\end{figure}
\end{wrapfigure}

One of the best known RRL algorithms is relational Q-Learning \cite{dvzeroski2001relational}. This method utilizes a symbolic knowledge representation for the goal-conditioned Q-values $Q^*(s,a,g)$, which generalize Q-values $Q^*(s,a)$ to different goals $g$. This function is learned with a relational regression tree \cite{kramer1996structural} which, given the current state $s$, goal $g$ to achieve and action $a$ to execute, predicts the associated Q-value $Q^*(s,a,g)$ (see Figure \ref{fig:figure_relational_RL}). 
%The relational P-function \cite{dvzeroski2001relational} is a form of policy-based RRL which explicitly encodes the policy represented by the relational Q-function. It also uses a relational tree but, instead of returning $Q^*(s,a,g)$, it outputs whether $a$ is optimal given $s$ and $g$.
% Better not to add boutilier2001symbolic as it should be in Symbolic Planning section (as SPUDD)
%\cite{boutilier2001symbolic} is an example of model-based RRL. It proposes a symbolic dynamic programming method which adapts the VI algorithm to the relational setting. 
In recent years, there has been a renewed interest in RRL. 
% Este trabajo es más antiguo que el que hemos añadido ahora (PEORL) y generaliza este último.
%\cite{yang2018peorl} proposes PEORL, a model-based RL architecture which combines SP and Hierarchical RL. It utilizes a symbolic action model (given a priori) to generate plans that guide the RL algorithm, and leverages the learned experience to enrich the symbolic knowledge and improve planning.
{
\cite{lyu2019sdrl} proposes a model-based, symbolic DRL framework which integrates SP with hierarchical DRL. It leverages a symbolic action model containing a high-level description of the environment dynamics to compute plans composed of subtasks to achieve. Then, DRL is used to learn a low-level policy for each subtask.
In \cite{jin2022creativity}, the previous neurosymbolic framework is augmented with the ability to learn the symbolic action model from trajectories. To do so, a function that maps subsymbolic, low-level states to symbolic, high-level states is required.
\cite{landajuela2021discovering} utilizes a recurrent neural network to generate symbolic policies represented as concise mathematical expressions. Generated policies are then evaluated on the environment, and the obtained rewards are used to train the policy generator with RL itself. 
}
On the other hand, \cite{zambaldi2019deep}
proposes a Deep RRL algorithm that departs from the classical RRL definition, since it does not employ a symbolic representation.
{Instead, the scene objects are detected with a CNN and a deep attention-based model \cite{vaswani2017attention} is used to reason about their interactions. The authors show their approach improves the efficiency, generalization and interpretability of conventional DRL.}
%On the other hand, \cite{zambaldi2019deep} proposes a Deep RRL algorithm that departs from the classical RRL definition, since it does not use a symbolic representation. Instead, the authors opt for a deep attention-based model \cite{vaswani2017attention} which improves the efficiency, generalization capacity and interpretability of conventional DRL. It works by detecting the objects of the scene with a CNN and reasoning about their interactions with a self-attention mechanism. 
{More information about RRL techniques can be found in \cite{tadepalli2004relational, yu2023reinforcement, acharya2023neurosymbolic}.}

\subsection{Learn to plan}
%\textit{\textbf{Num max. de páginas: 2}}

\begin{wrapfigure}{r}{0.41\textwidth}
	\centering
	\includegraphics[width=0.65\linewidth]{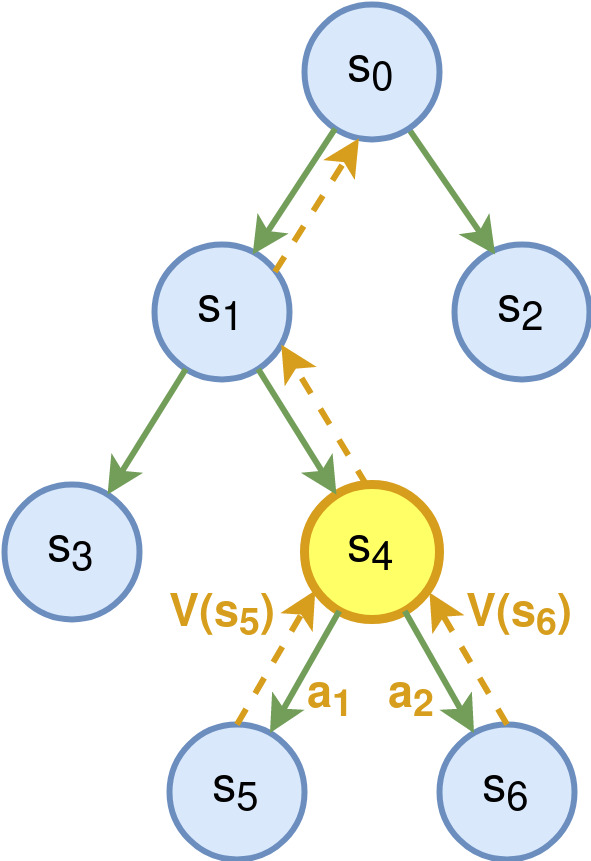}
	\caption{{\textbf{Learning to plan approaches.} 
 Green arrows represent the action model which determines the transitions between states (we assume determinism for simplicity purposes).
 Yellow items represent the different aspects of a planning algorithm, 
 which mainly comprises 1) expanding some state ($s_4$ in the image example),
 2) executing one or more actions ($a_1$, $a_2$ in the example),
 3) evaluating the resulting states ($s_5$, $s_6$)
 and 4) propagating this new information to the other states (yellow dashed lines).
 %mainly comprising 1) which state to expand ($s_4$ in the image example), 2) which action(s) to execute ($a_1$, $a_2$ in the example), 3) how to evaluate the new states ($s_5$, $s_6$) and 4) how to propagate this new information to the other states (yellow dashed lines).
 Some methods only learn the action model (green items) for a given planning algorithm  \cite{tamar2016value,srinivas2018universal}, other techniques only learn the planning computations (yellow items) \cite{guez2018learning,pascanu2017learning,toyer2018action} and, finally, some learn both \cite{farquhar2018treeqn, guez2019investigation}.
}}
	\label{fig:diagram_learn_to_plan}
\end{wrapfigure}

As we have seen previously, many MDP-solving techniques use an action model, which can be either known or learned from data. This model of the world can be used in several ways. The first option is to plan over it. We can use a symbolic planner, e.g., FF, or an NSP procedure, e.g., MCTS, depending on the type of knowledge representation used by the model, in order to simulate different courses of action and find the best one. The second option is to leverage the action model to obtain a policy or value function/heuristic. We can follow the model-based RL approach and employ the model to obtain data for training a value function or policy, or do as in SP and compute a domain-independent heuristic using the planning domain and problem. In addition to these two options, there exists a third alternative which has not been discussed yet: to \textit{learn to plan} \cite{moerland2023model}. This idea combines the RL and AP approaches and is inspired by the novel area of algorithmic reasoning \cite{cappart2021combinatorial}, which studies how to teach DNNs to compute algorithms. In the case of SDM, instead of considering the planning procedure as an external process, we can integrate it into the computational graph of our DNN, i.e., the graph containing the sequence of operations performed by the model to transform the inputs into outputs. There exist three main ways to embed the planning procedure into a DNN. Firstly, we can learn an action model that is compatible with a planning algorithm chosen a priori. Secondly, given an action model, we can learn how to perform the actual planning computations on it. Finally, we can jointly learn the action model and planning algorithm at the same time.
{These different approaches are compared in Figure \ref{fig:diagram_learn_to_plan}.}

\subsubsection{Fix the planning algorithm and learn the action model}

Some methods choose a planning algorithm a priori and embed it into the computational graph of the decision-making model. If the planning procedure is differentiable, 
meaning
the gradient of the learned policy/value function can be backpropagated through the planning operations,
it is possible to learn an action model compatible with it by using gradient-based optimization techniques. The action model is trained to output the optimal policy or value function as a result of the iterative computations performed by the chosen planning algorithm. Therefore, the action model learns a representation of the dynamics tailored to the task at hand and planning algorithm employed, as opposed to most action models which represent the environment dynamics in a general, task-independent way. This special type of action models belong to the family of value-equivalent action models (see Section \ref{subsection:task-specific_action_models}).

Value Iteration Networks (VINs) \cite{tamar2016value} are a good example of this. The authors of this work propose a differentiable implementation of the classical VI algorithm using recurrent CNNs.
%(see Figure \ref{fig:figure_VIN}).
This planning procedure is then embedded into a DNN architecture which can be trained end-to-end to predict the optimal policy by using standard RL and Supervised Learning, i.e., ML methods which are trained on labeled samples corresponding to \textit{(input, output)} pairs. The obtained model generalizes better than reactive policies, i.e., those that do not perform planning, when applied to new problems of the same domain. Universal Planning Networks (UPNs) \cite{srinivas2018universal} follow a similar approach. This work proposes a DNN architecture which learns a policy for a continuous task. The architecture integrates a differentiable planning module, which performs planning by gradient descent, and is trained end-to-end with Supervised Learning to predict the optimal policy. The resulting action model exhibits a state representation suitable for gradient descent planning, which can be adapted to other tasks.

% We remove figures that only depict a single method
%\begin{figure}[h]
	%\centering
	%\includegraphics[width=0.95\linewidth]{Figure_VIN.jpg}
	%\caption{\textbf{Value Iteration Network (VIN)}. The picture on the left shows the global VIN architecture, which receives as input a state observation $\phi(s)$ and outputs the policy $\pi_{re}$. Picture on the right shows the Value Iteration module, a recurrent CNN that performs the computations of the Value Iteration algorithm. Source: \cite{tamar2016value}.}
	%\label{fig:figure_VIN}
%\end{figure}

These methods are similar to those discussed in Section \ref{subsection:task-specific_action_models} like MuZero, since they also learn a value-equivalent action model that is suitable for planning. The difference between both types of methods is how the planning process is implemented. In MuZero, planning is performed explicitly, outside the computational graph. It learns a state representation useful for predicting the rewards, value function and policy for any given state. In VINs and UPNs, however, planning is performed implicitly, as a differentiable process embedded into the computational graph to predict the optimal policy. Thus, they may learn a slightly different state representation to that employed by MuZero and other models where planning is performed as a separate process.

\subsubsection{Fix the action model and learn the planning algorithm}

Other methods follow the opposite approach. Given an action model, which can be either known a priori or learned as a first step, they learn how to plan over it in order to solve the corresponding SDM task. The planning algorithm obtained will be able to harness the information contained in the action model to predict the optimal policy or value function. The existing methods in the literature provide different amounts of freedom to the planning algorithm, by controlling which parts of it are fixed a priori and which ones must be learned.

\cite{guez2018learning} proposes MCTSnet, a DNN architecture that embeds the MCTS algorithm into its computational graph. This model learns how to perform the different MCTS operations (selection, expansion, simulation and backpropagation) in order to play the \textit{Sokoban} game. Other works give even more freedom to the planning process. \cite{pascanu2017learning} proposes Imagination-Based Planner (IBP), a DNN architecture capable of constructing, evaluating and executing plans. It learns when to plan, which states to expand and when to stop planning. Both MCTSnet and IBP utilize a subsymbolic representation for their learned knowledge. Instead, \cite{toyer2018action} proposes Action Schema Networks (ASNets), a neurosymbolic model for learning to plan. ASNets correspond to DNNs specialized to the structure of planning problems. 
%(see Figure \ref{fig:figure_ASNet}).
They are composed of alternating action and proposition layers, with the specific network topology given by the associated planning domain and problem to solve, and a final layer which outputs the policy. Policies learned by ASNets are shown to generalize to different problems of the same domain.

\subsubsection{Learn both the action model and planning algorithm}

Lastly, some methods combine the two previous approaches. They embed a differentiable action model and differentiable planning procedure into the same computational graph, and then jointly optimize both parts to solve a particular SDM task. Although this idea represents the most end-to-end approach for learning to plan, the resulting model is hard to optimize, due to its great complexity and the interdependence between the action model and planning algorithm, as the quality of one depends on the quality of the other and vice versa.

In \cite{farquhar2018treeqn}, RL is used to jointly learn an action model and how to plan over it. The proposed method, called TreeQN, employs the learned model to try all possible action sequences up to a predefined depth, learning to predict the state values and rewards along the simulated trajectories. These values are then backed up the search tree to estimate the Q-values at the current state. \cite{guez2019investigation} proposes the Deep Repeated ConvLSTM (DRC) model, a powerful DNN architecture capable of learning to plan even though it incorporates no inductive bias for that purpose beyond its iterative nature. The model is composed of stacked ConvLSTM \cite{shi2015convolutional} blocks which are repeatedly unrolled to predict the policy and value function. The authors show that DRC exhibits characteristics of planning, such as an increase in performance when given additional thinking time.

\section{Methods to Learn the Structure of MDPs}
\label{section:methods_to_learn_the_structure}
%\textit{\textbf{Num max. de páginas: 7}}

\begin{figure}[h]
	\centering
	\includegraphics[width=.75\linewidth]{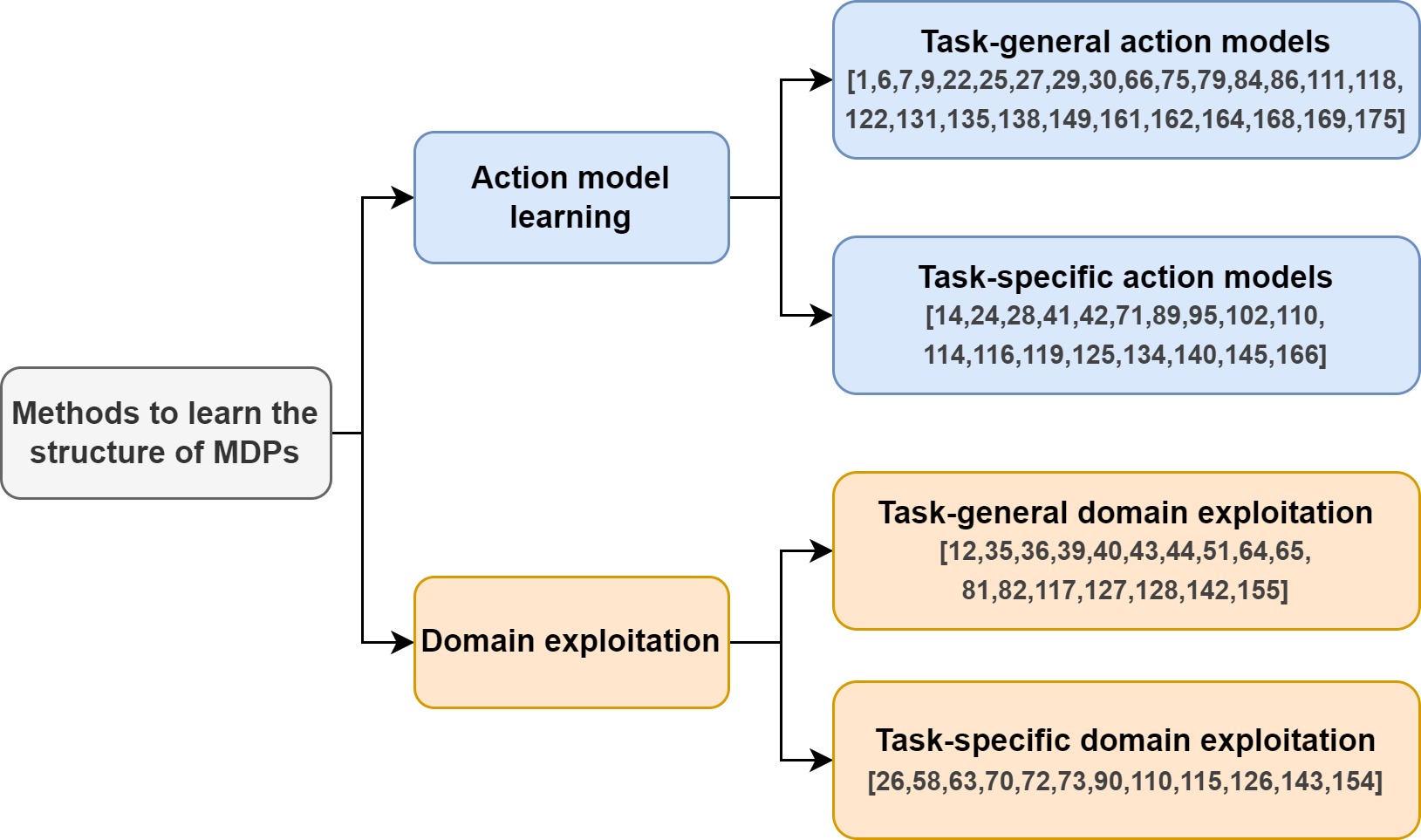}
	\caption{\textbf{{Proposed taxonomy of methods to learn the structure of MDPs.}}}
	\label{fig:diagram_structure_MDP}
\end{figure}

Every MDP has a different underlying structure. Its most important features are encoded in the action model, which describes the environment dynamics and is required by many of the MDP-solving methods seen in the previous section, e.g., AP and model-based RL. Additionally, there exist specific aspects of the MDP structure, e.g., landmarks \cite{hoffmann2004ordered} and state invariants \cite{fox1998automatic}, which if known can facilitate its resolution and provide insight into the properties of the MDP. Moreover, some learning methods \cite{shen2020learning, balduccini2011learning, hogg2008htn} require training data in the form of problem instances and example trajectories, which often need to be provided by domain experts. In this section, we discuss the main methods (see Figure \ref{fig:diagram_structure_MDP})  for learning these different aspects of the MDP structure.
%in case they are not given a priori.

%In this section, we summarize the main existing methods for learning the structure of MDPs. This information is required to apply many MDP-solving methods such as AP, which needs to know the environment dynamics. Even those methods which do not need information about the structure of the MDP can benefit from this extra knowledge. For instance, it has been shown that RL techniques which use a model of the world, i.e., model-based RL, are more sample-efficient than those which do not, i.e., model-free RL \cite{kaiser2020model}. In addition to the environment dynamics, there exist other aspects of the MDP structure which are useful for solving them, e.g., landmarks \cite{hoffmann2004ordered}, and for understanding its underlying properties, e.g., domain invariants \cite{kalyanam2013formal}. Finally, there exist methods for automatically generating data for a given MDP. For example, some methods are used to generate planning problems which, after being solved, provide a set of plan traces that serve as training data for ML and DL techniques, such as those which learn planning heuristics. However, in many real-world problems this information is not available. In this case, we can use methods for learning those structural aspects of the MDP which are required but not given a priori.

\subsection{Action model learning}
\label{subsection:action_model_learning}
%\textit{\textbf{Num max. de páginas: 4}}

The action model, which also receives the names of \textit{planning domain} and \textit{world model}, represents the most important aspect of the MDP structure. It encodes the dynamics of the environment and how the agent can affect them, i.e., the available actions for the agent and the effect each action has on the world state. The action model is essential for AP techniques, which need it to carry out their deliberative process. It is also advantageous for RL as it has been shown that those techniques which use an action model, i.e., model-based RL, are more sample-efficient than those which do not, i.e., model-free RL \cite{kaiser2020model}. Here, we discuss methods for automatically learning the action model from data. These techniques can be categorized according to the scope of the learned model in methods which learn task-general action models and those which learn task-specific action models.

\subsubsection{Task-general action models}
\label{section:task_general_action_models}
%\textit{\textbf{Num max. de páginas: 2}}

% Classification of methods to learn task-general action models
% TODO: do picture
% In the caption, analyze findings from the image. Examples:
% Almost no symbolic action models for both stochasticity and Partial Observability
% Symbolic and hybrid models tend to generalize better
% Every hybrid model is for deterministic MDPs
% >>> Note that models for POMDPs are applicable to MDPs and models applicable to stochastic environments are also applicable to deterministic ones.

%Subsymbolic
%Det MDP: \cite{deisenroth2011pilco}
%Stoch. MDP: \cite{sutton2008dyna, hester2013texplore, khansari2011learning, abbeel2004learning, asadi2018towards}
%Det POMDP: \cite{wahlstrom2015pixels, chiappa2017recurrent}
%Stoch. POMDP: \cite{depeweg2017learning, chrisman1992reinforcement}

%Symbolic
%Det MDP: \cite{shen1989rule, wang1996learning, walsh2008efficient}
%Stoch. MDP: \cite{oates1996searching, pasula2007learning, jimenez2008pela, safaei2007incremental}
%Det POMDP: \cite{yang2007learning, zhuo2014action, mourao2008using, segura2021discovering}
%Stoch. POMDP: \cite{yoon2007towards}

%Hybrid
%Det MDP: \cite{battaglia2016interaction, chang2017compositional, kansky2017schema, kipf2020contrastive, asai2022classical}
%% Stoch. MDP:
%% Det POMDP:
%% Stoch. POMDP:

\begin{figure}[h]
	\centering
	\includegraphics[width=.7\linewidth]{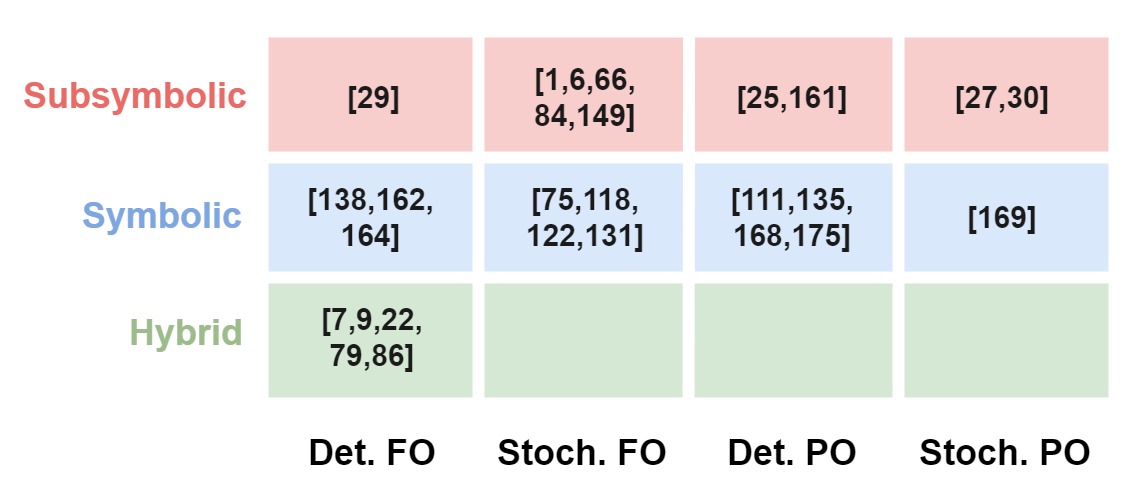}
	\caption{{\textbf{Comparison among methods to learn task-general action models.} Methods are grouped according to their knowledge representation (Y axis) and the type of MDPs they can be applied to (X axis): deterministic MDPs (\textit{Det. FO}), stochastic MDPs (\textit{Stoch. FO}), deterministic POMDPs (\textit{Det. PO}) and stochastic POMDPs (\textit{Stoch. PO}). Methods for learning (stochastic) POMDPs are the most general, as they can be applied to every other MDP category. Methods that use a symbolic or hybrid representation for their learned action model tend to generalize better to states not seen during training. Finally, we note that all the hybrid methods presented in this section tackle the deterministic fully observable case. These works focus on improving the properties (e.g., data-efficiency, interpretability and generalizability) of subsymbolic methods via their proposed hybrid representation, leaving more complex settings (e.g., stochastic and PO) for future work.}}
	\label{fig:diagram_task_general_models}
\end{figure}

% Summarized
These methods try to learn a general model that represents the dynamics of the environment as accurately as possible
and which can, in theory, be applied to solve any task of the corresponding domain.
These techniques can be further grouped according to the type of knowledge representation employed by the learned model ({see Figure \ref{fig:diagram_task_general_models}}).

%These methods try to learn a model which represents the dynamics of the environment as accurately as possible, in what is known as a task-general action model. Instead of learning a model tailored for a specific task/problem, i.e., a specific reward function in RL or a particular set of goals in AP, they obtain a general model which can, in theory, be applied to solve any task of the corresponding domain. These techniques can be further grouped according to the type of knowledge representation employed by the learned model.

\begin{figure}[h]
	\centering
	\includegraphics[width=.49\linewidth]{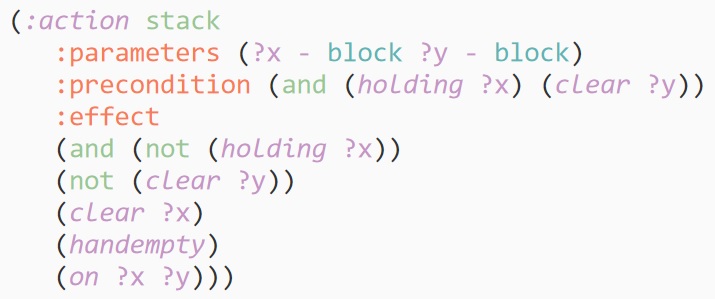}
	\includegraphics[width=.49\linewidth]{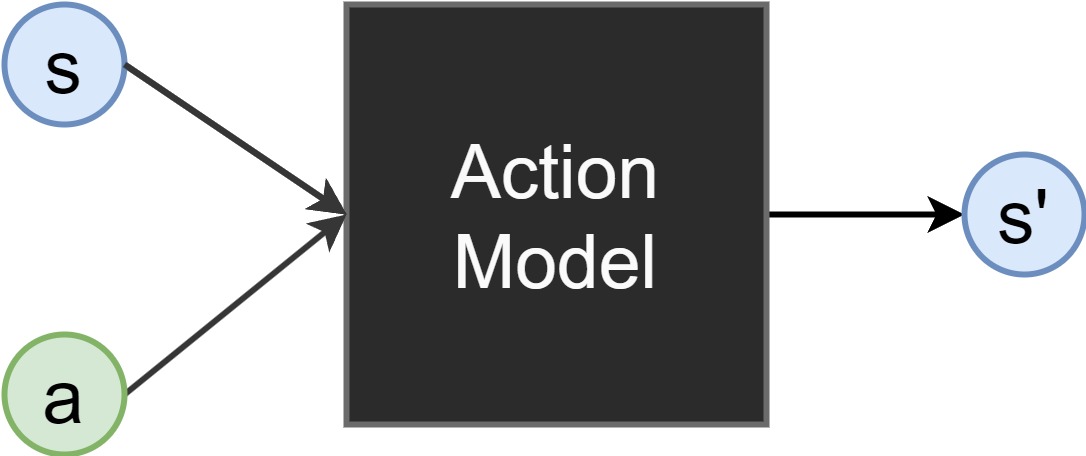}
	\caption{\textbf{Symbolic and subsymbolic action models.} \textbf{Left:} Extract from a symbolic action model, corresponding to the PDDL description of action \textit{stack} in the blocksworld domain. { Given a symbolic description of a state $s$, as a set of objects and atoms, and a \textit{grounded applicable} action $a$, i.e., an action with parameters instantiated on objects at $s$ and whose preconditions are met at $s$, it allows to obtain the next state $s'$ by applying the effects of $a$ to $s$.
    \textbf{Right:} Subsymbolic action model, corresponding to a black-box. It receives a subsymbolic representation of $s$ (e.g., as an image or a set of state variables) and $a$ as inputs, and outputs $s'$. The symbolic action model is amenable to interpretation, whereas the subsymbolic one is usually not.}
    %that receives $s$ and $a$ as inputs and outputs $s'$ and $r$. The symbolic action model is amenable to interpretation, whereas the subsymbolic one is not.
    }
	\label{fig:comparison_action_models}
\end{figure}

\noindent \textbf{Task-general models with subsymbolic knowledge representation.} These models represent the environment dynamics in a subsymbolic way, often as a black-box which receives as inputs a state $s$ of the world and an action $a$ to execute, and outputs the next state $s'$ %and often also the associated reward $r$ 
(see Figure \ref{fig:comparison_action_models}, right). They are usually trained with ML and DL techniques, e.g., linear regression \cite{sutton2008dyna}, 
%nearest neighbours \cite{jong2007model},
random forests \cite{hester2013texplore}
%Gaussian processes \cite{wang2005gaussian} 
and DNNs \cite{wahlstrom2015pixels}, in a supervised manner on samples of the form {$(s,a,s')$} collected from the environment.

For an effective learning of subsymbolic action models, some aspects which contribute to the uncertainty of the model must be considered. Firstly, we need to take into account the estimation errors which occur when the model is applied to regions of the state-space not seen during training. Most works, such as PILCO \cite{deisenroth2011pilco}, address this problem by estimating the uncertainty in the model predictions. Secondly, most environments are non-deterministic, so the learned action model must reflect this stochasticity in some way. Two possible solutions are to approximate the entire next state distribution \cite{khansari2011learning} and to learn a generative model from which we can draw samples \cite{depeweg2017learning}. Thirdly, some environments exhibit partial observability, i.e., they are POMDPs. Action models for POMDPs need to incorporate information about previous states using methods such as belief states \cite{chrisman1992reinforcement} 
or recurrent neural networks \cite{chiappa2017recurrent}.
%or neural turing machines \cite{gemici2017generative}. 
Finally, in order to perform a multi-step look-ahead, the predicted next state $s'$ must be repeatedly fed into the model as input. To prevent prediction errors from accumulating, some works use multi-step prediction losses for training \cite{abbeel2004learning} whereas others learn different models for each \textit{n}-step prediction \cite{asadi2018towards}. A deeper view into subsymbolic action models can be found in {\cite{moerland2023model, plaat2023high}}.

\noindent \textbf{Task-general models with symbolic knowledge representation.} These models use an interpretable, formal language often based on FOL to represent objects and their relations. 
% Note, this reference is for PDDL (Classical planning) and not for PPDDL (prob. planning), which is younes2004ppddl1
One example is PDDL \cite{haslum2019introduction}, the standard language in {CP} for representing planning domains and used by the majority of {classical} planners. In PDDL, the planning domain encodes the existing types of objects, predicates and actions. Each action has a series of parameters (variables that can be instantiated with objects), preconditions (predicates which must be true or false before executing the action) and effects (see Figure \ref{fig:comparison_action_models}, left).
The effects correspond to those grounded predicates (i.e., atoms) which, after applying the action, will become true (\textit{add} effects) and those which will turn false (\textit{delete} effects).
{
PDDL is extended to the general PP case by the PPDDL \cite{younes2004ppddl1} language, which adds support for stochastic action effects that occur with a given probability.
RDDL \cite{sanner2010relational} is an alternative language for modeling PP tasks.
In RDDL, everything (including actions) is represented as a parameterized variable of a particular type. This modeling approach is better suited than PPDDL for domains where actions have many uncorrelated effects, as often occurs in systems with many objects that mostly evolve independently from each other.
}

Methods for learning symbolic action models usually receive as input a set of trajectories $(s_0, a_0, s_1, ..., a_{n-1}, s_n)$, obtained by solving planning problems of the corresponding domain. Then, they try to find the planning domain, i.e., the preconditions and effects of each action in the domain, which best fits the given trajectories. The simplest scenario corresponds to learning planning domains for totally-observable, deterministic environments. This is a well-studied problem which has been solved in numerous ways \cite{shen1989rule, wang1996learning, walsh2008efficient}. Action preconditions are inferred by analysing the predicates which appear at the states preceding an action whereas action effects are learned by comparing the predicates of the states before and after applying the action, in what is known as the \textit{delta-state}. Other works learn planning domains for environments with uncertainty, which can come in the form of non-deterministic actions or partial observability of states (POMDPs). \cite{oates1996searching, pasula2007learning, jimenez2008pela} tackle the case of non-determinism whereas \cite{yang2007learning, zhuo2014action, mourao2008using} do the same for partially observable environments. \cite{segura2021discovering} also tackles partially observable environments but is able to learn more expressive domains than the previous methods, with numerical variables and relations. The hardest case corresponds to learning action models for environments which present both non-determinism and partial observability. This problem has been poorly studied, with just one preliminary work trying to address it \cite{yoon2007towards}. Finally, we can classify the existing methods according to the type of algorithm used to learn the planning domain, e.g., RL \cite{safaei2007incremental},
%relational RL \cite{rodrigues2012active}, 
%surprise-based learning \cite{molineaux2014learning},
Supervised Learning \cite{mourao2008using},
inductive rule learning \cite{segura2021discovering},
MAX-SAT \cite{yang2007learning}
and transfer learning \cite{zhuo2014action}.
A deeper view into symbolic action models can be found in \cite{jimenez2012review, arora2018review}.

\begin{figure}[h]
	\centering
	\includegraphics[width=0.95\linewidth]{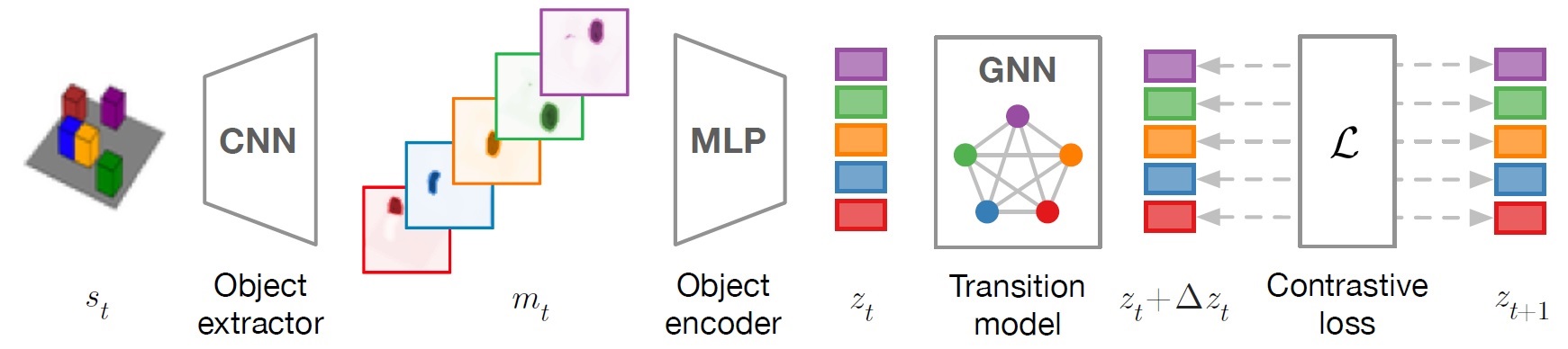}
	\caption{\textbf{Action model with a hybrid knowledge representation}. Reproduced with permission by the authors of \cite{kipf2020contrastive}. Firstly, the \textit{object extractor} (a CNN) receives an image representation of the current state $s_t$ and outputs a set of object masks $m_t$, used to extract the objects in $s_t$ (each one associated with a different color in the image). Secondly, the \textit{object encoder} (a multilayer perceptron or MLP) receives $m_t$ and returns a set of abstract object states $z_t$. Thirdly, the \textit{transition model} (a Graph Neural Network) receives $z_t$ and the action to apply to each object as inputs, and predicts the resulting abstract state $z_t + \Delta z_t$. This prediction is then compared with the ground truth $z_{t+1}$, in order to train the model. © Thomas Kipf}
	\label{fig:figure_hybrid_action_model}
\end{figure}

\noindent \textbf{Task-general models with hybrid knowledge representation.} { Several works use a hybrid knowledge representation for the action model, one that sits between the black-box representation usually employed by subsymbolic methods and the logic-based representation of symbolic methods. They encode the action model in terms of objects and their relations, which results in better interpretability and generalization to novel situations (e.g., different number of objects) than purely subsymbolic models.}
%Several works try to combine the representations used by symbolic and subsymbolic action models. They represent the action model in terms of objects and their relations, in a way that resembles the FOL-based representations employed by symbolic methods. This type of object-centered representations are more interpretable than purely subsymbolic models and also generalize better to novel situations, such as scenes with a different number of objects. 
\cite{battaglia2016interaction} learns a physics simulator which receives as input a graph encoding a set of objects and interactions to consider, and applies a DNN to predict the new states of the objects. \cite{chang2017compositional} also learns a physics simulator but implements it as an encoder-decoder architecture, where the encoder summarizes each pair-wise interaction of an object with its neighbours and the decoder predicts the future state of the object.  \cite{kansky2017schema} learns a set of abstract \textit{schemas} which encode local cause-effect relationships between objects; these schemas are instantiated with the objects at the scene to form the schema network, a probabilistic model used to predict the reward obtained by executing a given sequence of actions. \cite{kipf2020contrastive} uses a CNN to extract objects from images, obtains an embedding for each object with an encoder and predicts the interactions between objects with a Graph Neural Network (see Figure \ref{fig:figure_hybrid_action_model}). 
{ Lastly, \cite{asai2022classical} proposes Latplan, a neurosymbolic method which uses Variational Autoencoders \cite{Kingma2014auto} to learn a planning domain from image pairs describing environment transitions. The learned domain is represented as \textit{grounded} PDDL, i.e., as a PDDL model with only constants and no variables, thus corresponding to propositional logic instead of FOL.}
% Aprender lifted PDDL models en Latplan es dejado como future work.

% Lastly, \cite{asai2022classical} proposes Latplan, a neurosymbolic method which uses Variational Autoencoders (VAE) \cite{Kingma2014auto} to learn a grounded PDDL model, i.e., a PDDL model with only constants and no variables, from image pairs representing environment transitions. This makes it possible to apply a symbolic, off-the-shelf planner to solve an MDP for which its symbolic description is unknown.

\subsubsection{Task-specific action models}
\label{subsection:task-specific_action_models}
%\textit{\textbf{Num max. de páginas: 2}}

\begin{figure}[h]
	\centering
	\includegraphics[width=0.9\linewidth]{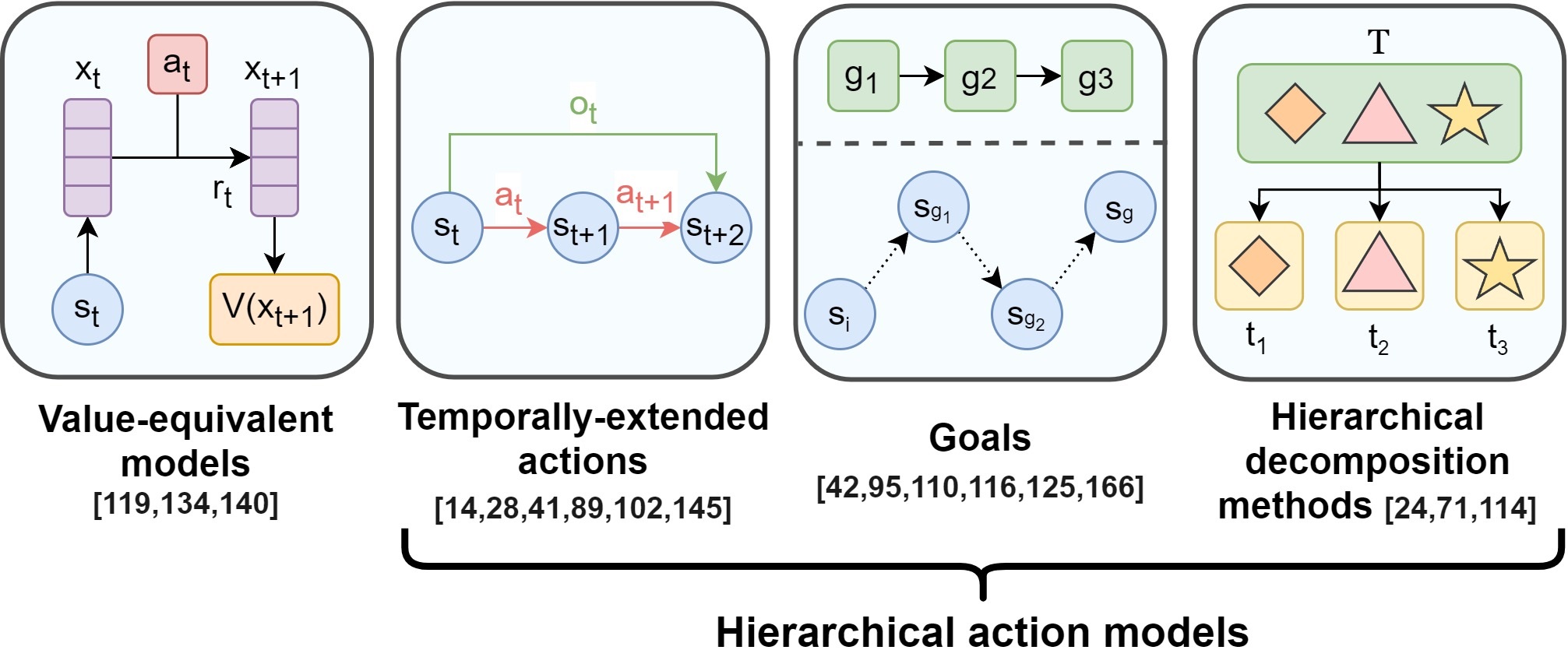}
	\caption{{\textbf{Comparison among methods to learn task-specific action models}. Methods are grouped according to how they integrate task-specific knowledge into the action model.
    \textbf{Value-equivalent models:} 
    in the image example, the current state $s_t$ is encoded into a task-specific latent representation $x_t$, which is then unrolled with action $a_t$ to predict the next latent state $x_{t+1}$, reward $r_t$ and value $V(x_{t+1})$.
    %the current MDP state $s_t$ is encoded into a latent representation $x_t$ which only describes those aspects of the state useful for solving the task at hand. The model learns the state transitions in this latent representation (in the image, $x_{t+1}$ results from applying $a_t$ to $x_t$) and to predict rewards ($r_t$) and state-values ($V(x_{t+1})$).
    \textbf{Temporally-extended actions:} 
    in the image example, applying the temporally-extended action $o_t$ to $s_t$ is equivalent to executing the sequence of primitive actions $a_t, a_{t+1}$.
    %these are high-level actions that take several time-steps to execute. In the image example, applying the temporally-extended action $o_t$ to $s_t$ is equivalent to executing the primitive actions $a_t$ and $a_{t+1}$.
    \textbf{Goals:} 
    in the image example, the sequence of goals $g_1, g_2, g_3$ guides the low-level policy (first from $s_i$ to $s_{g_1}$, then from $s_{g_1}$ to $s_{g_2}$ and, finally, to $s_{g_3}=s_g$). Dotted arrows abstract several state transitions.
    % Goal Reasoning methods deliberate about the (sub)goals $g_1, g_2, ..., g_n$ needed to achieve the task goal $G$. In the image example, the high-level goal sequence $g_1, g_2, g_3$ guides the low-level policy (first from $s_i$ to $s_{g_1}$, then from $s_{g_1}$ to $s_{g_2}$ and, finally, to $s_g$). Pointed arrows represent multiple state transitions $s_0, s_1, ..., s_n$.
    \textbf{Hierarchical decomposition methods:} 
    in the image example, the high-level task $T$ is decomposed into three subtasks (each represented by a different shape): $t_1$, $t_2$ and $t_3$.
    % they describe how to decompose a complex task into a series of simpler subtasks. In the image example, the high-level task $T$ is decomposed into three subtasks (each represented by a different shape): $t_1$, $t_2$ and $t_3$.
    }}
	\label{fig:task_specific_models}
\end{figure}

Unlike the techniques previously discussed, these methods learn a model specifically tailored for a task or set of tasks, i.e., a task-specific action model. This model represents the dynamics of the environment and, additionally, encodes task-related knowledge which is useful for solving the corresponding tasks. Techniques for learning task-specific action models can be further categorized according to how they integrate this extra knowledge inside the action model ({see Figure \ref{fig:task_specific_models}}).

\noindent \textbf{Value-equivalent action models.}
The dynamics of complex environments are very difficult to model accurately (as task-general action models try to do), since states are composed of a large number of interrelated elements. However, in most cases only a subset of these state features are actually relevant for the task at hand. Value-equivalent models \cite{grimm2020value} are a type of subsymbolic action models which, instead of being trained to predict the next state as accurately as possible, learn a state representation useful for predicting the value, i.e., reward, of future states. Thanks to this, they learn to only focus on task-relevant state features and abstract away those aspects of the environment not useful to solve the task. One of the most successful implementations of this idea can be found in MuZero \cite{schrittwieser2020mastering}. This work extends the AlphaZero \cite{silver2018general} algorithm to the scenario where no action model is provided a priori. MuZero learns a value-equivalent action model which is used by the MCTS planning procedure to predict the value of future states and select the best action. MuZero achieved state-of-the-art results on the Atari video game environment and matched the performance of AlphaZero on Go, chess and shogi. Value Prediction Networks \cite{junhyuk2017value} follow a similar approach, training an action model to predict future rewards and values,
%(see Figure \ref{fig:figure_VPN}),
which is then used by a search algorithm.
%This model is then used by a \textit{b}-best, \textit{d}-depth search algorithm instead of MCTS, where \textit{b} and \textit{d} are hyperparameters.
In a similar fashion, the Predictron \cite{silver2017predictron} learns an action model which can be repeatedly rolled forward to predict the value of the state received as input.

%\begin{figure}[h]
%	\centering
%	\includegraphics[width=0.9\linewidth]{Figure_VPN.jpg}
%	\caption{\textbf{Value Prediction Network}. { \textbf{Left:} value-equivalent action model. It receives as inputs a state $x$ and an option $o$ (instead of an action). The encoding module $f^{enc}$ maps $x$ to an abstract state $s$, the outcome module $f^{out}$ predicts the reward $r$ and discount factor $\gamma$ associated with executing $o$ at $s$, the transition module $f^{trans}$ predicts the next abstract state $s'$ and, lastly, the value module $f^{value}$ estimates the value $V^*(s')$ of $s'$.}
%    %that receives a state $x$ and option $o$ (instead of an action) and learns to predict the reward $r$, discount factor $\gamma$, next state $s'$ and its value $V(s')$. 
%    \textbf{Right:} the action model is unrolled multiple times to perform planning. Source: \cite{junhyuk2017value}.}
%	\label{fig:figure_VPN}
%\end{figure}

\noindent \textbf{Hierarchical action models.}
These models distribute knowledge across multiple levels of hierarchy or abstraction. The bottom level represents the environment dynamics in a general, task-independent way akin to task-general action models. Then, one or more higher levels encode specific, task-dependent knowledge which facilitates the resolution of the corresponding task(s). The main methods for representing hierarchical knowledge are temporally-extended actions \cite{sutton1999between}, goals \cite{aha2018goal} and hierarchical decomposition methods \cite{georgievski2015htn}.

Temporally-extended actions \cite{sutton1999between}, also known as macroactions \cite{korf1985macro} and options \cite{sutton1999between}, are a type of abstract, high-level actions whose execution extends across several time steps, as opposed to primitive, low-level actions which are applied in just one step. A set of options defined over an MDP constitutes a semi-Markov Decision Process \cite{sutton1999between}, a special type of MDP where actions take variable amounts of time to finish. 
%(see Figure \ref{fig:figure_options}).
Options consist of three components: a policy, an initiation set and a termination condition. In order to start executing an option, the current state must belong to the initiation set. Once started, the agent selects actions according to the option policy until the termination condition is met. Options make it possible to group actions which are often executed together. In AP, this translates into a reduction in the depth of the search tree \cite{jimenez2012review} whereas, in RL, options facilitate exploration and value propagation, which in turn speeds up learning \cite{mcgovern1998macro}. Nevertheless, options also increase the number of alternatives to choose from at each state. For this reason, it is important to carefully consider how many options to use. Options have been successfully learned and applied to both the field of AP \cite{fikes1972learning, botea2005macro, coles2007marvin} and RL \cite{stolle2002learning, machado2017laplacian, konidaris2007building}. 

% La imagen no aporta mucho, es un poco confusa y ocupa espacio.
%\begin{figure}[h]
%	\centering
%	\includegraphics[width=.4\linewidth]{Figure_options.jpg}
%	\caption{\textbf{Options, MDPs and SMDPs}. { \textbf{Top:} MDP, where an action (black dot in the image) is executed at each time step. \textbf{Middle:} SMDP, where the execution of options (empty dots in the image) takes several time steps. \textbf{Bottom:} options over MDP, where the execution of an option (empty dot) results in the execution of several actions (black dots) until the option terminates. Source: \cite{sutton1999between}.}
    %In an MDP, actions are executed in just one time step. In an SMDP, options are executed across several time steps.
%    }
%	\label{fig:figure_options}
%\end{figure}

Goal Reasoning \cite{aha2018goal} provides a design philosophy for agents whose behaviour revolves around goals. It makes it possible to design agents which not only learn how to obtain a particular goal, but also reason about what should be the goal to achieve in the first place. This is especially important for dynamic environments where unexpected events (discrepancies) may require a change in the goals to pursue.
%may render the current goals unsuitable for the new situation and require new goals to pursue.
Goal-Driven Autonomy  \cite{molineaux2010goal} provides a general framework for goal-reasoning agents that detect discrepancies, generate possible explanations for them, formulate new goals and manage (prioritize) the pending goals.
%\cite{jaidee2012learning} uses Case-based Reasoning \cite{kolodner2014case} and RL to learn to detect discrepancies, associate them to new goals and learn policies to achieve the selected goals.
\cite{weber2012learning} proposes an agent which learns to formulate goals using expert demonstrations for the StarCraft game. \cite{pozanco2018learning} learns to predict future goals based on current and past states and performs an anticipatory planning process which considers both current and future goals. \cite{nunez2022learning} proposes a neurosymbolic method which combines DRL with SP to select and achieve goals in order to reduce planning times. Finally, in RL goals have also been utilized as a method to improve exploration \cite{forestier2022intrinsically, laversanne2018curiosity}. These methods learn a goal space from which goals are sampled in order to direct exploration towards interesting regions of the state space.

%\begin{wrapfigure}{r}{0.42\textwidth}
%	\centering
%	%\includegraphics[width=.55\linewidth]{Figure_HTN.jpg}
 %       \includegraphics[width=0.9\linewidth]{Figure_HTN.jpg}
%	\caption{\textbf{Example of an HTN decomposition method}. The depicted method decomposes the compound task \textit{transport} into four subtasks, two of which are also compound (\textit{dispatch} and \textit{return}) and two of which are primitive (\textit{load} and \textit{move}). Source: \cite{nau2003shop2}.}
%	\label{fig:figure_HTN}
%\end{wrapfigure}

Hierarchical decomposition methods make it possible to split a complex problem into a series of subproblems which are simpler to solve, in order to reduce computational effort.
% so that the total computational effort of solving these subproblems is smaller than if the original problem had been directly addressed.
The most common way to perform this decomposition is provided by Hierarchical Task Networks (HTN) \cite{georgievski2015htn}. HTN domains contain a set of tasks, which can be grouped in either primitive or compound. Primitive tasks correspond to actions that can be directly executed in the environment, in a similar fashion to PDDL actions. Conversely, compound tasks cannot be directly executed and need to be decomposed into a series of primitive or compound tasks. Every compound task has associated one or more decomposition methods, representing different ways to achieve the same task. Each decomposition method has some preconditions which must be true in order to be applied and defines a sequence of subtasks the original task can be decomposed into.
%(see Figure \ref{fig:figure_HTN}).
HTN planners receive as inputs an HTN domain and a compound task to achieve, known as the goal task.
Then, they find a valid decomposition of the goal task as a sequence of primitive tasks by repeatedly applying the decomposition methods available. There exist a wide variety of HTN planners such as NOAH \cite{sacerdoti1975nonlinear}, Nonlin \cite{tate1977generating}, SHOP2 \cite{nau2003shop2} and SIADEX \cite{castillo2006efficiently}. One issue of HTN planning is the substantial time investment needed to encode HTN domains.
% One issue of HTN planning is the substantial time investment needed to encode HTN domains, which is why it has been widely adopted in autonomous spacecraft control \cite{gancet2005task, rudnick2016scalable}, a field where this effort is justified. 
% No añado esta parte por falta de espacio (y no creo que aporte mucho)
To alleviate this burden, several methods have been developed to automatically learn HTN domains from data \cite{hogg2008htn, nejati2006learning, chen2021learning}. This learning data is comprised of a set of trajectories obtained from experts, often along with some additional information such as annotated tasks or partial method definitions.

\subsection{Domain exploitation}

An action model defines the structure of the MDP. Some aspects of this structure are explicitly encoded in the action model description, e.g., the available actions and environment dynamics. However, other properties of the MDP, such as state invariants and landmarks, do not appear in this description. Throughout the years, many techniques have tried to learn this additional structural information with different purposes, such as facilitating the resolution of the MDP or generating data for training some of the methods discussed in Section \ref{section:methods_to_solve_MDPs}. Unlike methods for learning action models, this field lacks a proper structure. Thus, we propose to group these techniques under the name \textit{domain exploitation}. In addition, we further categorize them in methods
that learn information about the entire domain and those which only learn information about a specific task.

\subsubsection{Task-general domain exploitation}

\begin{figure}[h]
	\centering
	\includegraphics[width=0.9\linewidth]{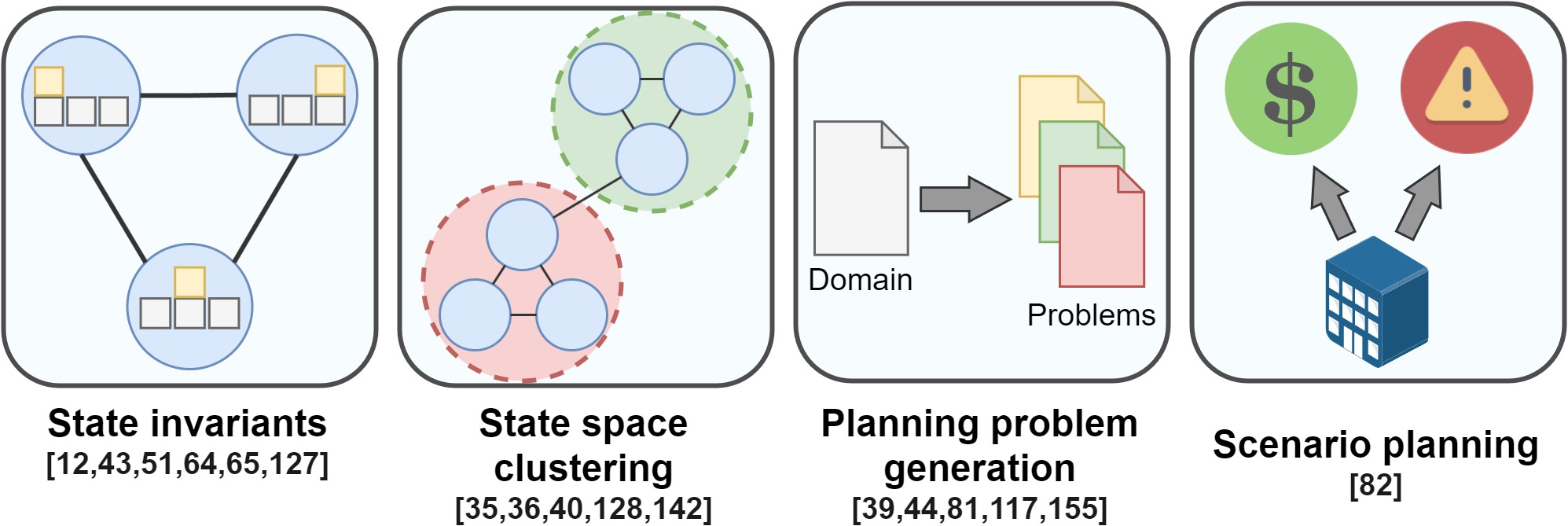}
	\caption{{\textbf{Comparison among task-general domain exploitation methods.}
 \textbf{State invariants:} 
 {the figure} shows an example \textit{exactly-1} invariant where, at each state (blue node), the yellow block is on top of exactly one other block.
 \textbf{State space clustering:}
 {the figure} shows an example state clustering method which groups together states that are densely connected (red and green circles in the image).
 \textbf{Planning problem generation:}
 these methods generate a set of planning problems pertaining to some particular domain.
 \textbf{Scenario planning:}
 in the image example, a company (represented by a blue building) wants to foresee possible future scenarios, both good (green circle) and bad (red circle), {to} select the best course of action.
  }} 
	\label{fig:task_general_domain_exploitation}
\end{figure}

These methods learn information about the entire domain and do not focus on any particular task. We discuss four different approaches: state invariants, state space clustering, planning problem generation and scenario planning ({see Figure \ref{fig:task_general_domain_exploitation}}).

\noindent \textbf{State invariants.} {These are properties which hold true for every valid, i.e., reachable, state of the MDP. 
The most widely-known invariant type corresponds to \textit{mutex constraints} \cite{blum1997fast}, which declare that several state properties are mutually exclusive, i.e., cannot be true at the same time. For example, in the blocksworld domain, a block $X_1$ can never be on top of more than one other block at the same time, i.e., the set of atoms $\{on(X_1,X_2), ..., on(X_1, X_n)\}$ are pair-wise mutually exclusive for all $n$ blocks at the state. Another invariant is given by the \textit{exactly-n} constraint which, given a set $P = \{p_1, p_2, ..., p_m\}$ of properties, states that exactly $n$ properties from $P$ must be true in every MDP state. For example, at each blocksworld state $s$, the arm must either be holding one block $X_i$ (i.e., $holding(X_i) \in s$) or be empty (i.e., $handempty() \in s$). This corresponds to an \textit{exactly-1} invariant where $P=\{holding(X_1),...,holding(X_n),handempty()\}$. There exist many methods for automatically extracting state invariants given a symbolic (e.g., PDDL) domain description. Most methods \cite{helmert2009concise, gerevini1998inferring, rintanen2008regression} extract invariants via inductive reasoning: they prove that, if a given invariant is true at some state $s$, it will remain true at all successor states $s'$ of $s$ (i.e., those obtained by executing an applicable action at $s$). Some methods follow a different aproach. 
For instance, TIM \cite{fox1998automatic} models the behavior of objects in the domain using finite state machines. It forms classes of objects with shared behavior, {from which state invariants are inferred.}
%For instance, TIM \cite{fox1998automatic} represents the behaviour of objects in the domain using finite state machines and forms classes of objects that share the same behaviour. 
%{These classes are then used to infer state invariants.}
%These classes are then used to infer a rich type structure along with different state invariants.
Finally, state invariants have many applications in SP. For example, the FD planner \cite{helmert2006fast} needs to extract mutex constraints in order to translate the planning task from PDDL to a different encoding in terms of multi-valued variables, whereas the STAN \cite{long1999efficient} planner leverages the state invariants extracted by TIM in order to enhance system performance.}

\noindent \textbf{State space clustering.}  Some methods apply clustering techniques to the state space of the MDP. They group states together according to a notion of similarity or distance between states, which must be properly defined in order to obtain clusters with the desired qualities. This definition can incorporate information about the task at hand or be completely task-agnostic. Thus, some state space clustering techniques are task-general and obtain a clustering for the entire domain while others are task-specific and focus on obtaining a clustering suitable for a concrete task. Additionally, several methods require a symbolic description of the MDP whereas others do not impose this restriction.
\cite{singh1994reinforcement} proposes a RL algorithm that groups together states $s$ with similar value $V^*(s)$ using a form of soft aggregation, where a state belongs to each cluster with a certain probability. The RL agent learns a value function at the cluster level, and calculates the value of a given state as the weighted sum of the values of the clusters it belongs to.
%\cite{singh1994reinforcement} uses clustering as a form of function approximation, i.e., generalization, for classical RL. In this approach, the RL agent observes the primitive states but can only update the value function for the clusters. A form of soft state aggregation is used, where a state belongs to each cluster with a certain probability. Then, the value of a cluster generalizes to all the states that belong to it, in proportion to the clustering probabilities. These probabilities are calculated with a task-specific method that tries to minimize the value function estimation error, i.e., states with similar values are grouped together. 
\cite{feyzabadi2017planning} partitions the MDP state space into a smaller number of abstract states or clusters
%(see Figure \ref{fig:figure_state_clustering})
and uses the resulting abstract MDP to compute the optimal policy in a more efficient manner. The used clustering algorithm groups together states that are connected and have similar value. 
In SP, the most widely used approach for state clustering selects a subset $P$ of state variables (e.g., atoms), called a \textit{pattern}, and assigns any two MDP states $s_i, s_j$ to the same abstract state (i.e., cluster) if they share the same value for all variables in $P$. The clustering obtained is then often leveraged to calculate a planning heuristic, known as a \textit{pattern database heuristic}, based on distances computed over the abstract state space induced by the pattern $P$ \cite{edelkamp2001planning, edelkamp2007automated, rovner2019counterexample}.

\noindent \textbf{Planning problem generation.} { Given a planning domain, we may be interested in obtaining a set of planning problems pertaining to that particular domain. Among other applications, they can be used as training data for methods that apply ML to SP (e.g., \cite{shen2020learning, balduccini2011learning}) {and as benchmarks to compare planning performance, as done in the International Planning Competitions}. In most situations, these problems need to be created by hand
or produced by hard-coded, domain-specific problem generators, which requires great effort from the human designers. Nonetheless, there exist several methods for automatically generating planning problems. \cite{fern2004learning} generates problems through random walks. It randomly generates an initial state $s_i$ and executes $n$ random actions to arrive at another state $s_n$. Then, it selects a subset of the atoms of $s_n$, which constitutes the goal $g$, and returns the corresponding planning problem $(s_i, g)$. Problems generated with this method are always solvable but they may be inconsistent, i.e., the initial state $s_i$ generated may correspond to an impossible situation of the world (e.g., in blocksworld, a state where a block is simultaneously on top of two blocks).
\cite{fuentetaja2012planning} also employs a random walk approach but is able to generate problems that are valid, i.e., both solvable and consistent. To achieve this, it receives as inputs the domain description along with some additional information that determines the characteristics of the problems generated, in order to preserve consistency.
\cite{torralba2021automatic} proposes Autoscale, a method that leverages domain-specific generators in order to obtain problems that are valid, diverse and of graded difficulty, for their use in planning competitions.
\cite{nunez2023nesig} proposes NeSIG, a neurosymbolic method that uses DRL to learn to generate problems for a given PDDL domain, so that they are valid, diverse and difficult to solve.
%(see Figure \ref{fig:figure_problem_generation}).
Lastly, \cite{katz2020generating} generates complete planning tasks (i.e., domain-problem pairs) that are difficult, diverse and of a particular structure specified by the user.}

\noindent \textbf{Scenario planning.} This is a decision support technique where the goal is to generate a variety of possible future scenarios to help organizations foresee the future and adapt to it. \cite{katz2021scenario} proposes a semi-automatic, neurosymbolic method for performing scenario planning in the real world. It uses DNNs to extract forces and their causal relations from a set of documents encoded in natural language. These forces are the elements of study in scenario planning, e.g., \textit{pandemic}, \textit{lockdown} and \textit{loss of benefits}. The causal relations between forces determine \textit{what causes what}, e.g., \textit{pandemic} may cause \textit{lockdown} which in turn may cause \textit{loss of benefits}. Once this information has been extracted, it is translated into a PDDL planning domain and problem, which can then be solved with a symbolic planner. The solutions (plans) found by the planner correspond to possible future scenarios, which can be provided to human experts for their analysis.

%\cite{katz2021scenario}. IBM Scenario Planning. Generate alternative futures to help decision making and risk management (prepare for what's about to come).They use neural causal extraction techniques to derive hidden causal relations from documents.Scenarios can be derived from sets of plans for the corresponding planning problem.They apply neural causal extraction to collection of documents and produce a planning domain and problems collection.In previous work, Mind Maps were created manually. Mind Maps describe the causal relation between all the forces. In this work, they eliminate the need for Mind Maps (causal relation between forces are extracted from documents using neural networks).\< They use a neurosymbolic approach. They use neural causal extraction techniques to derive causal relations between forces from a collection of documents and build a symbolic AI planning model (they produce a planning domain and collection of problems from the collection of documents) from the causal relations to capture the solutions to the scenario planning problem.\>%

\subsubsection{Task-specific domain exploitation}

\begin{figure}[h]
	\centering
	\includegraphics[width=0.7\linewidth]{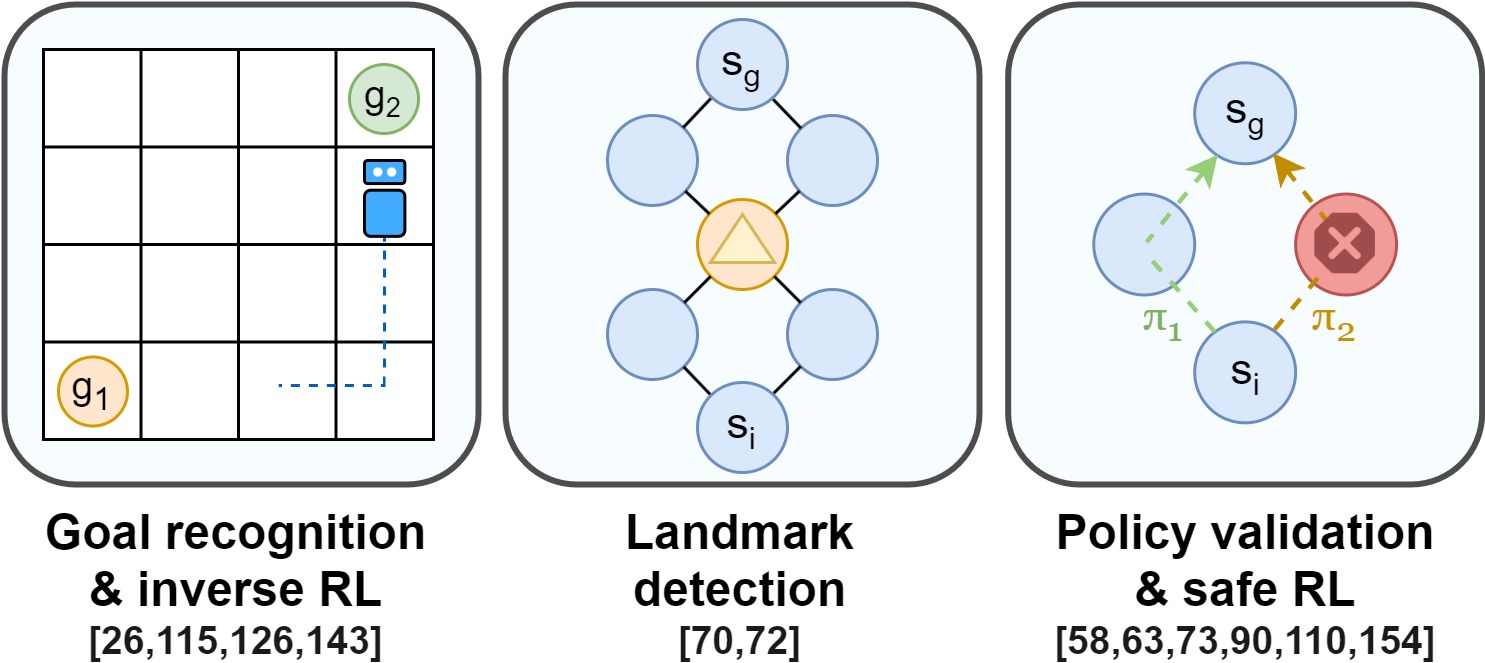}
	\caption{{\textbf{Comparison among task-specific domain exploitation methods.}
    \textbf{Goal recognition \& inverse RL:} {the figure} shows an example goal recognition task, where we want to determine whether the agent (represented by a robot) is pursuing goal $g_1$ or $g_2$. From the agent's actions (blue dotted line in the image), it is obvious that it is pursuing $g_2$.
    \textbf{Landmark detection:} {the figure} shows an example state landmark (yellow node with a triangle), as every trajectory from $s_i$ to $s_g$ must necessarily traverse this state.
    \textbf{Policy validation \& safe RL:} image depicts an example safe RL task. Policy $\pi_2$ is unsafe, as it traverses an unsafe state (red crossed node), whereas $\pi_1$ is safe since it does not traverse this state.}} 
	\label{fig:task_specific_domain_exploitation}
\end{figure}

We now present methods that, instead of extracting information about the entire domain, focus on a specific problem/task and its solution. In this work, we discuss three different approaches: goal recognition/inverse RL, landmark detection and policy validation/safe RL ({see Figure \ref{fig:task_specific_domain_exploitation}}).

\noindent \textbf{Goal recognition \& inverse RL.} Goal recognition, also referred to as plan recognition, is the problem of finding the goals that best explain the observed behaviour of an agent. 
%This has many applications such as risk management \cite{sohrabi2019ibm}, video games \cite{synnaeve2011bayesian} and network monitoring \cite{sohrabi2013hypothesis}.
\cite{ramirez2009plan} follows a \textit{plan recognition as planning} approach. Instead of using a library of plans, the proposed method only needs to know the planning domain, the possible set $G$ of goals and a partially observed plan $p$ representing the behaviour of the agent. Then, the authors use standard SP techniques to find those goals $g \in G$ for which the optimal plan that achieves them is compatible with the observations in $p$.
%(see Figure \ref{fig:figure_goal_recognition}).
Several works have extended this approach, such as \cite{sohrabi2016plan}, which provides a relaxation of the problem formulation that allows it to consider noisy and missing observations. 

%\begin{wrapfigure}{r}{0.45\textwidth}
%	\centering
%        \includegraphics[width=0.5\linewidth]{Figure_goal_recognition.jpg}
%	\caption{\textbf{Plan recognition in a navigation domain}. Room A (marked with a circle) is the initial position of the agent, while rooms C, I and K (marked with a square) are its possible destinations. { The partially observed plan $p$ contains two observations, $A \rightarrow B$ and $F \rightarrow G$, corresponding to the movements of the agent from room A to B and from room F to G, respectively. Given the observed behaviour, the only possible destinations are I and K but not C. This is because the optimal plan that reaches room C from A is not compatible with the observation $F \rightarrow G$, i.e., the agent does not need to go from F to G in order to reach C. Source: \cite{ramirez2009plan}.}}
%	\label{fig:figure_goal_recognition}
%\end{wrapfigure}

In the context of RL, goal recognition is known by the name of inverse RL. Here, a different formulation is employed, in terms of rewards instead of goals. The aim of inverse RL is to infer the reward function being optimized by an agent, given its policy or some observations about its behaviour. 
% Some example applications are helicopter control \cite{abbeel2006application}, path planning in robotics \cite{kim2016socially} and fuel-efficient driving \cite{vogel2012improving}.
\cite{ng2000algorithms} provides a foundational method to address this problem. It takes an expert's policy as input and utilizes linear programming to obtain the reward function that maximally differentiates the input policy from others, in terms of their optimality. \cite{choi2011map} follows a different approach. It uses Bayesian inference to estimate the posterior probability of the reward functions, given some prior distribution over them and the observed behaviour data. 
%The authors resort to gradient-based optimization in order to efficiently calculate the reward function that maximizes this posterior probability distribution.
A comprehensive survey of inverse RL is provided in \cite{arora2021survey}.

%\begin{wrapfigure}{r}{0.45\textwidth}
%	\centering
%        \includegraphics[width=0.5\linewidth]{Figure_landmarks.jpg}
%	\caption{\textbf{Landmarks in a navigation domain}. The agent starts at cell $S$ and must get to cell $G$. The dotted arrows represent different possible paths (plans). No matter the chosen path, the yellow cells will always need to be traversed in order to reach $G$. Thus, they correspond to (state) landmarks in this planning task.}
%	\label{fig:figure_landmarks}
%\end{wrapfigure}

\noindent \textbf{Landmark detection.} In SP, landmarks are properties (or actions) that must be true (or executed) at some point for every plan that solves a particular planning problem. 
%(see Figure \ref{fig:figure_landmarks}).
Landmarks have been successfully applied to a wide range of problems, such as computing planning heuristics \cite{karpas2009cost} and performing goal recognition \cite{pereira2020landmark}.
% > In this review we focus on single-agent MDPs <
%and planning in multi-agent environments (both cooperative \cite{maliah2017collaborative} and competitive \cite{pozanco2018counterplanning}).
One of the most widely used methods for automatically obtaining planning landmarks can be found in \cite{hoffmann2004ordered}. Given the description of a CP task, this work is able to extract as landmarks those facts (propositions) which must necessarily be made true during the execution of any solution plan. The proposed method also approximates the order in which these landmarks must be achieved, and encodes this information as a directed graph where the nodes correspond to landmarks and the edges to order restrictions between them. Then, this graph is used to decompose the planning task into smaller sub-tasks. An iterative algorithm is used where, at each step, the leaf nodes of the graph (corresponding to those landmarks that can be directly achieved) are handed to the planner as goals and deleted from the graph. This process is repeated until all the landmarks have been achieved. The experiments carried out by the authors show this method can significantly reduce planning times when used in conjunction with state-of-the-art planners. \cite{holler2021landmark} presents a method to automatically extract landmarks for HTN planning. The proposed method represents the HTN task with an AND/OR graph which is used to obtain different types of landmarks corresponding to facts, actions and methods. The authors show this technique is able to extract more than twice the number of landmarks obtained by other methods, which can then be employed to improve the time performance of HTN planners.

\noindent \textbf{Policy validation \& safe RL.} Given an MDP and a policy (or plan), we may be interested in testing whether the policy actually corresponds to a valid solution of the MDP or not. This problem receives the name of policy (or plan) validation. In SP, one of the most important plan validation systems is VAL \cite{howey2004val}, which was initially developed to automatically validate the plans produced as part of the 3rd International Planning Competition. It can check whether a given plan, represented in PDDL, is executable and achieves the corresponding goals. In case the plan is flawed, VAL gives advice on how the user can fix it. Additionally, VAL provides different visualization options. Another form of plan validation can be found in the plan monitoring process carried out by online planning architectures. These systems, which interleave planning and execution, must be able to detect \textit{discrepancies}, i.e., unexpected events that require a modification in the behaviour of the agent. For example, the goal reasoning framework proposed in \cite{molineaux2010goal} allows agents to detect discrepancies, infer their causes and generate new goals to pursue, which may require a new plan.

Policy safety is a very important aspect of policy validation. Given a policy, we may be interested in determining if there exists some state in the MDP for which our policy performs very poorly. Safe RL tries to address this issue. It comprises RL techniques which, in addition to maximizing reward, satisfy certain criteria regarding the performance (safety) of the system during the learning and/or deployment processes. This is especially important for real-world environments with critical safety requirements, such as helicopter flight \cite{koppejan2011neuroevolutionary} and gas turbine control \cite{hans2008safe}. One possible approach to safe RL is to transform the reward function so that it includes some notion of risk. For example, \cite{heger1994consideration} proposes a pessimistic version of the classical Q-Learning algorithm. Instead of predicting the expected total reward $Q^*(s,a)$ associated with a state $s$ and action $a$, it estimates the minimum possible total reward (under the optimal policy) for $(s,a)$. This way, it learns the policy that maximizes reward for the worst-case scenario. Other works focus on adapting the exploration process followed by the agent in order to learn the policy. In \cite{thomaz2006reinforcement}, a human user is allowed to guide an RL agent. The human teacher can provide feedback to the agent in the form of a reward function and, additionally, restrict the set of actions the agent can take at any given moment. Thanks to this guidance, the RL agent is able to learn the optimal policy more efficiently while exploring the state space in a safer way. A comprehensive survey of safe RL can be found in \cite{garcia2015comprehensive}.

\section{Towards an Ideal Method for SDM}
\label{section:an_argument_for_the_integration}
%\textit{\textbf{Num max. de páginas: 7}}

%\textbf{\color{blue} Esta sección es enteramente nueva}

%\textcolor{red}{Si sobra espacio, añadir una imagen con las 5 ideal properties for SDM: Applicability, Ease of use, Efficiency, Intepretability, Generalizability. Puedo dibujar un círculo/pentágono dividido en cada parte. Cada parte representa una propiedad, tiene un dibujo y el nombre de la propiedad.}

In this section, we try to provide further insight into the existing methods for solving MDPs. Firstly, we propose to categorize these methods along two main dimensions: how they solve the MDP and how they represent their knowledge. Secondly, we discuss what properties an ideal method for SDM should exhibit. Based on these properties, we then analyse the advantages and disadvantages of the different approaches for solving MDPs.

% In this section, we propose a categorization of the different existing methods that solve MDPs. In addition, we discuss what properties an ideal method for SDM should exhibit. Based on these desirable properties, we discuss the main advantages and disadvantages of the different approaches for solving MDPs.

%In this work, we have presented many different approaches for solving MDPs. In light of all these different alternatives, we may wonder what advantages and disadvantages each method provides. Additionally, we may wonder what properties an ideal method for SDM would 

% In this work, we have presented many different approaches for solving MDPs. In this section, we try to provide further insight into the differences between these alternative methods. Firstly, we propose a categorization of these methods along two different dimensions: how they solve the MDP and how they represent their knowledge. Secondly, we discuss what properties an ideal method for SDM should exhibit. Based on these desirable properties, we analyse the main advantages and disadvantages of the different approaches for solving MDPs. 

% Classification of methods to solve MDPs according to two axes. Show diagram. Explain diagram (main methods for each category, including neurosymbolic methods).

We note that the existing methods for solving MDPs differ from one another in two main aspects. Firstly, they differ in the approach employed to obtain a solution of the MDP. Some methods (e.g., AP algorithms) use an action model to synthesize their solution, often by performing a reasoning process over it. Conversely, other methods (e.g., model-free RL algorithms) do not require a model of the MDP and, instead, learn their solution using the data obtained by interacting with the environment. Secondly, methods for solving MDPs differ in the type of knowledge representation employed. Whereas some methods represent their knowledge symbolically using a formal, logic-based language (e.g., PDDL), other methods encode their knowledge subsymbolically, usually into the weights of a DNN.

\begin{figure}[h]
	\centering
	\includegraphics[width=0.97\linewidth]{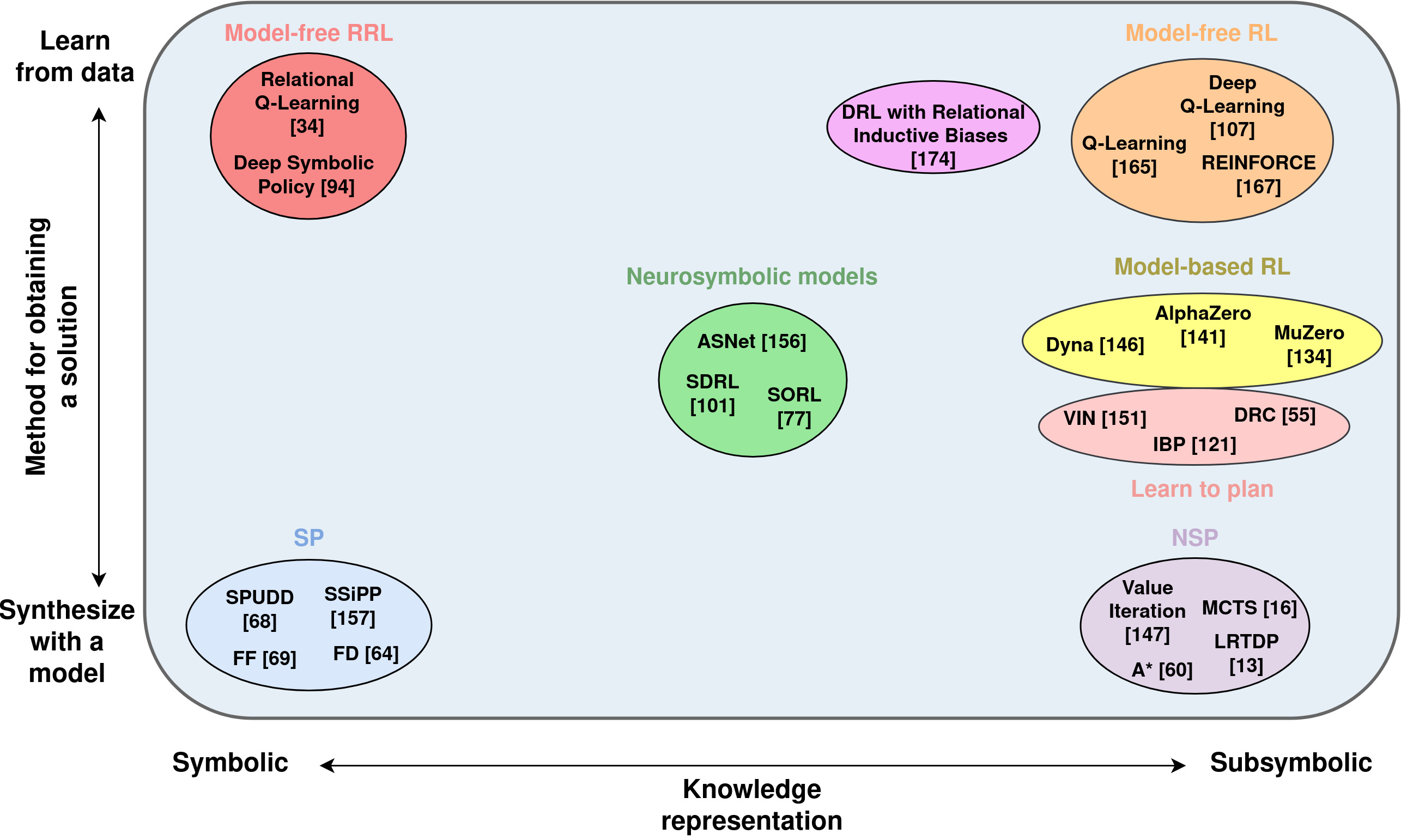}
	\caption{\textbf{Diagram of methods according to their knowledge representation and method for obtaining their solution.} Methods placed in the middle of an axis represent hybrid approaches that combine the two extremes of the corresponding axis, i.e., methods that both learn from data and synthesize their solution with a model (Y axis) and methods that integrate symbolic and subsymbolic knowledge representations (X axis). Colored bubbles are used to group similar methods together.}
	\label{fig:diagram_2D_methods}
\end{figure}

Figure \ref{fig:diagram_2D_methods} shows a diagram with different methods for solving MDPs, organized according to how they solve the MDP (Y axis) and how they represent their knowledge (X axis).
The bottom of the diagram contains methods that employ an action model to synthesize a solution of the MDP, i.e., AP methods. AP methods that employ a symbolic knowledge representation, i.e., SP methods, are placed on the bottom left corner of the diagram, whereas those with a subsymbolic representation, i.e., NSP methods, are placed on the bottom right corner.
The top of the diagram corresponds to methods that do not require an action model and instead learn the MDP solution from the data obtained by interacting with the environment, i.e., model-free RL methods. Most model-free RL methods, from both tabular RL and DRL, represent their knowledge subsymbolically and, thus, are placed on the top right corner of the diagram. Some of them, known as model-free RRL, do employ a symbolic knowledge representation and, hence, are placed on the top left corner. 
Some RL methods, known as model-based RL, leverage an action model in order to learn their solution of the MDP. Since they combine the two alternative methods for obtaining the MDP solution, they are placed in the middle of the Y axis on the diagram.
% El unico model-based RRL method que estaba en el diagrama descubri que era SP en realidad (como SPUDD)
%Model-based RL methods that employ a subsymbolic knowledge representation are placed on the middle right part of the diagram, whereas those with a symbolic representation, known as model-based RRL, are placed on the middle left.
{Since all the model-based RL methods discussed in the paper employ a subsymbolic knowledge representation, they are placed on the right of the diagram.}
The middle right part of the diagram also corresponds to those methods that \textit{learn to plan}, as they represent their knowledge subsymbolically and both employ an action model and learn their solution from data.
The Deep RRL method proposed in \cite{zambaldi2019deep} corresponds to a model-free RL algorithm (thus placed on the top of the diagram) that employs a DNN with a relational inductive bias in the form of an attention mechanism. Therefore, this work uses a subsymbolic knowledge representation that shares some properties with symbolic ones and, hence, is placed between the middle and right part of the X axis.
Finally, neurosymbolic models combine both methods for obtaining a solution, the \textit{learn from data} of RL and \textit{synthesize with a model} of AP, and additionally integrate the symbolic and subsymbolic knowledge representations. This is why they are placed in the middle of our diagram.

% Properties of an ideal method for SDM. Explain what a user would desire. Explain the five properties.

% Explicar las cinco propiedades de manera escueta, ya compararé las ventajas de AP y RL y symbolics vs subsymbolic en el analysis.

In light of so many different approaches for solving MDPs, we may wonder what are the advantages and disadvantages of the existing methods. In order to answer this question, we first discuss what properties an ideal AI for SDM should exhibit. We frame this question from the perspective of a user who \textit{simply} wants to solve a particular SDM task in the best way possible. Firstly, an ideal SDM method should be \textbf{applicable} to any task posed by the user. Secondly, the method should be \textbf{easy to use}, requiring as little human effort as possible. Furthermore, it should solve the task \textbf{efficiently}. In addition, it should be \textbf{interpretable} by the user.
Finally, the solution obtained by the method should \textbf{generalize} to other tasks different from the one it was obtained for. These requirements give shape to five different properties that result desirable for MDP-solving methods: %({see Figure \ref{fig:figure_ideal_method_SDM}}):

\begin{itemize}
    \item \textbf{Applicability.} An ideal method for SDM should be capable of solving all different sorts of MDPs. Nonetheless, there exist many characteristics that make MDPs more difficult to solve and, thus, limit the applicability of SDM methods. In this work, we have focused on two of them: stochasticity and partial observability (i.e., POMDPs). A few other MDP features that limit applicability are: continuity (i.e., MDPs with continuous state spaces $S$ and/or action spaces $A$), multiple agents (e.g., games like chess where an opponent must be beaten), noisy state observations (e.g., a robot with faulty sensors) and/or noisy actions (e.g., a robot with faulty actuators), the size of the state space, and the complexity of the MDP dynamics.

    % Poner en el análisis (siguiente parte):
    % For instance, CP algorithms cannot solve stochastic nor partially-observable MDPs.
    % Tabular RL methods cannot be applied to MDPs with big state spaces whereas DRL methods can
    % Solving some MDPs requires reasoning with numerical and/or temporal aspects that surpass the capabilities of CP methods
    
    % In this work, we have focused on two of them: stochasticity and partial observability (i.e., POMDPs). For instance, CP algorithms cannot solve stochastic nor partially-observable MDPs. A few other MDP features that limit applicability are: continuity (i.e., MDPs with continuous state spaces $S$ and/or action spaces $A$), multiple agents (e.g., games like chess where an opponent must be beaten), noisy state observations (e.g., a robot with faulty sensors) and/or noisy actions (e.g., a robot with faulty actuators), the size of the state space (e.g., tabular RL methods cannot be applied to MDPs with big state spaces whereas DRL methods can) and the complexity of the MDP dynamics (e.g., solving some MDPs requires reasoning with numerical and/or temporal aspects that surpass the capabilities of CP methods).

    \item \textbf{Ease of use.} This property refers to the amount of human effort needed to adapt an SDM method to a particular task, so the larger the effort, the harder it is to use. An ideal method for SDM should require as little human effort as possible. The amount of effort is directly proportional to the quantity of prior knowledge (about the MDP to solve) required by the method and how difficult it is to encode this knowledge in a suitable representation for it.

    \item \textbf{Efficiency.} An ideal method for SDM should obtain a solution of the MDP as efficiently as possible. In Computer Science, efficiency is usually measured in terms of the time and space (i.e., memory) an algorithm needs to solve a problem. Analogously, efficiency in AP is usually measured as the time a method spends to find the MDP solution or, alternatively, as the number of tasks from a set that it is able to solve given some time and memory limits, what is known as \textit{coverage}. On the other hand, efficiency in RL is often measured as the amount of data a method needs to achieve a certain performance, what is known as \textit{data-efficiency}.

    \item \textbf{Interpretability.} In recent years, there has been great interest in developing interpretable/explainable AI (XAI) \cite{gunning2019darpa} systems for many different applications, including SDM {\cite{chakraborti2020emerging, hickling2023explainability}}. An ideal SDM method should be fully interpretable, i.e., humans should be able to understand how it works, the characteristics of the solution obtained by the method and the reasons behind the actions it takes. In addition to understanding and verifying the system, interpretability allows to modify its behaviour, thus opening the door to building collaborative systems where humans and SDM methods cooperate to solve problems.
    
    \item \textbf{Generalizability.} In some cases, we may be interested in solving not one but a set of different (although similar) MDPs. Solving each MDP separately may be computationally intractable if the number of MDPs in the set is very large (or even infinite). For this reason, an ideal method for SDM should be able to generalize, i.e., the solution obtained by the method for a particular MDP should also be applicable to solve other similar MDPs.  
    %to reuse the solution obtained for one MDP to solve other similar MDPs.
\end{itemize}

%\begin{figure}[h]
%	\centering
%	\includegraphics[width=0.5\linewidth]{Figure_ideal_method_SDM.jpg}
%	\caption{{\textbf{Properties of an ideal method for SDM.}}}
%	\label{fig:figure_ideal_method_SDM}
%\end{figure}

% ----------------------------

We now leverage these five desirable properties to compare the different existing methods for solving MDPs, assessing their advantages and disadvantages.

% >> APPLICABILITY

% > Subsymbolic more applicable than symbolic.
%   For some complex MDPs, it may be very hard to obtain a logical representation of their dynamics, in the form of objects and their relations (ver abajo). Poner ejemplo MDP donde sería complicado obtener esta representación simbólica.

% > Model-free more applicable than model-based.
%   We cannot provide an action model if 1) we do not know the environment dynamics or 2) the dynamics are too complex to encode them either symbolically or subsymbolically (ver Doc).

% Subsymbolic easily manages extensions of MDP. Ejemplos Deep Q-Learning. Symbolic has a harder time doing that. We believe that's because extensions to CP require specific formal representations and ways of reasoning with that type of knowledge.
%Regarding applicability, SDM methods that represent their knowledge subsymbolically can often be applied to a wider set of different scenarios than symbolic methods.

Regarding applicability, most subsymbolic methods manage many different types of MDPs either off the shelf or with some minor modifications. For example, Deep Q-Learning naturally manages non-determinism and has been extended to also manage partial observability and noise \cite{hausknecht2015deep}, and continuous state and action spaces \cite{gu2016continuous}. 
On the other hand, the symbolic community has historically focused on solving the CP case so, even though there exist symbolic methods capable of managing aspects such as non-determinism {(e.g., probabilistic planners)} and partial observability \cite{draper1994probabilistic}, their support is currently more limited than the one provided by subsymbolic methods. We believe the reason behind this is that, in order to manage these extensions to the CP case, symbolic techniques require more complex methods for representing knowledge (e.g., PPDDL instead of PDDL for non-determinism) and reasoning with it, which must be designed by humans, as opposed to subsymbolic techniques where a DNN usually \textit{takes care of everything} and naturally manages all these aspects. Therefore, subsymbolic methods exhibit better applicability, in general, than symbolic ones.
Additionally, it may be the case that we do not have access to an action model (e.g., the MDP dynamics are unknown) and we cannot learn it with the methods described in Section \ref{subsection:action_model_learning} because either we lack the necessary data or the dynamics are too complex to learn. In these scenarios, methods that synthesize an MDP solution with a model, such as AP, cannot be employed. Therefore, these methods exhibit worse applicability than those that learn their solution from data without a model, such as model-free RL.

Regarding ease of use, SDM methods that require an action model (e.g., AP) are harder to use than those which do not (e.g., model-free RL). This is due to the fact that, as previously commented, in some situations the action model cannot be learned and needs to be designed by a human expert, thus requiring additional human effort.
Additionally, symbolic methods generally require more human effort, and thus provide worse ease of use, than subsymbolic ones. Symbolic methods require knowledge to be represented symbolically, in a logic-based language such as PDDL. Nonetheless, 
{some MDPs may be hard (or even impossible) to describe using FOL or some other logic.}
%some MDPs may be hard  to describe in terms of a set of objects and their relations/interactions, so obtaining a logical representation for them may require extensive human effort.
On the other hand, subsymbolic methods often impose no restrictions on how knowledge must be represented, so the user can choose whichever representation scheme it likes (including symbolic ones).
%whichever representation scheme (including symbolic ones) that minimizes the effort needed to encode this knowledge.
For example, subsymbolic model-based RL methods such as AlphaGo and Dyna are indifferent to how the action model is represented, as long as it allows them to obtain the next state $s'$ and reward $r$ associated with a given state-action $(s,a)$ pair.

% EFFICIENCY

% Methods that employ more prior knowledge leverage this extra knowledge to achieve higher efficiency.
% > Model-based more efficient than model-free
%   Ej.: poner comparación efficiency model-based vs model-free RL que aparece en la versión antigua de la review. Ver si hay algún trabajo que compare efficiency de NSP (ej.: MCTS) con model-free RL.
% Posible referencia para inefficiency of model-free RL: Trust region policy optimization by Schulman et al
% > Symbolic more efficient than subsymbolic?
%   Ver si hay trabajos que comparen efficiency RRL con "RL normal" y SP con NSP.
% Extra knowledge -> more efficiency. Examples: SP algorithms that exploit symbolic action model description to compute heuristics (see if there is comparison between NSP (e.g., MCTS) and FF, FD...) (ver también review antigua). Example2: reward machines. Use inner (symbolic) structure of rewards to achieve better efficiency than "normal" RL where the reward function is a black-box.

The efficiency of SDM methods is directly proportional to the amount of task knowledge that they leverage. On the one hand, methods that employ an action model are usually more efficient than those which do not. For example, it has been shown that model-based RL achieves higher data-efficiency than model-free RL \cite{kaiser2020model}. 
On the other hand, symbolic representations often provide more knowledge than their subsymbolic counterparts and, hence, result in higher efficiency. For instance, SP techniques make use of powerful domain-independent heuristics to speed up search which, in order to be computed, require a symbolic description of the action model,
%\cite{keyder2014improving} or of states and goals \cite{frances2017purely},
so they cannot be exploited by AP methods with purely subsymbolic representations. 
% Another good example are Reward Machines \cite{icarte2022reward}... -> They use a "symbolic" representation for the reward function, but instead of a logic language like PDDL, they use a finite state machine with transitions depending on propositional symbols.

% INTERPRETABILITY (VER BLOC DE NOTAS MÓVIL)
% > Model-based more interpretable than model-free?
%   La idea es que un method that synthesize the solution with a model sigue un algoritmo, un proceso secuencial, que es más fácil de interpretar que una política reactiva que mappea "directamente" un input (observation) a un output (action). BUSCAR REFERENCIAS (explainability in NSP).
% Symbolic more interpretable than subsymbolic
%   Symbolic representations (e.g., PDDL) more amenable to interpretation than black-boxes (in the form of action models or policies learned by a neural network). Poner referencia a explainable planning.

Regarding interpretability, SDM methods that synthesize their solution with a model (e.g., AP) tend to be more interpretable than those which do not (e.g., model-free RL).
This is mainly due to the fact that methods in the first category solve the MDP via an iterative process, carrying out a series of steps which can then be analyzed in order to provide insight that is out of reach for model-free methods. For instance, \cite{magnaguagno2017web} provides a visualization tool of the search tree of CP algorithms, showing the heuristic value for different stages of the search. This is useful for understanding how different planners solve the task, step by step.
Additionally, symbolic methods are more interpretable than subsymbolic ones, as they represent their knowledge in a logic-based language that is comprehensible to humans (or, at least, human experts). For example, PDDL 
action models explicitly state what are the effects of each action, whereas subsymbolic models do not.
Another example is the solution obtained by symbolic methods. 
{For instance, the solution obtained by a CP algorithm is encoded as a sequence of grounded actions and, hence, is more interpretable than the one obtained by a DRL method (which is encoded in the weights of a DNN).}

Finally, generalizability depends on the specific scope of each SDM method. Some methods (e.g., CP algorithms) focus on solving a single MDP, whereas others (e.g., Generalized Planning \cite{jimenez2019review} and many DRL algorithms) aim to solve a set of different MDPs. 
Given the same scope, policies that carry out an iterative process over an action model in order to select the action to execute, i.e., \textit{deliberative} policies, often generalize better than those which do not, i.e., \textit{reactive} policies.
For instance, \cite{tamar2016value} shows that deliberative policies learned by VINs generalize better than reactive policies when applied to new problems of the same domain.
Besides, representing knowledge symbolically also helps achieve good generalization. Two examples are RRL, where FOL-based representations are leveraged in order to generalize to tasks with different number of objects, and DNNs with relational inductive biases \cite{zambaldi2019deep}, which generalize better than purely subsymbolic DNNs.

% Reactive policies (e.g., singlef forward pass of DNN) generalize worse than policies that, in order to select the action to execute, carry out an iterative process over an action model. This iterative process can more easily adapt to changes and different situations than a reactive policy/value-function (algorithms generalize better than functions) -> algorithmic reasoning

% Resumir resultados. Ventajas y desventajas (según estas 5 propiedades) de symbolic vs subsymbolic y learn from data vs synthesize with a model.

% Method for obtaining the solution
% Synthesize with a model: efficiency, interpretability, generalizability.
% Learn from data: applicability, ease of use,

% Knowledge representation
% Symbolic: efficiency, interpretability, generalizability.
% Subsymbolic: applicability, ease of use, 

To summarize, SDM methods that synthesize their MDP solution with a model exhibit better efficiency, interpretability and generalizability, whereas methods that learn their solution from data, without a model, are more widely applicable and easier to use. Analogously, symbolic methods present better efficiency, interpretability and generalizability, whereas subsymbolic methods present better applicability and ease of use. As a result of this analysis, we conclude that the two competing paradigms for obtaining the MDP solution (synthesizing it with a model versus learning it from data) and representing knowledge (symbolically versus subsymbolically) are complementary, as the shortcomings of each one correspond to the strengths of the other. Therefore, we argue that an ideal method for SDM should integrate these different approaches into the same architecture: it should both learn and plan in order to obtain its solution, and combine the symbolic and subsymbolic representations for its knowledge.
As shown in Figure \ref{fig:diagram_2D_methods},
neurosymbolic methods perform this integration, which is why they are situated in the center of the diagram.
%neurosymbolic methods combine the two existing paradigms for obtaining the solution and representing knowledge.
%neurosymbolic AI is currently the only approach that combines both paradigms for solving the MDP and representing knowledge.
Therefore, we believe neurosymbolic AI poses a very promising approach towards achieving an ideal method for SDM, one that would exhibit the five desirable properties detailed in this section. 

\section{Future Directions}
\label{section:future_directions}
%\textit{\textbf{Num max. de páginas: 3}}

This review has helped us to identify promising future directions to further advance the field of SDM via the integration of symbolic and subsymbolic AI. We now mention some of these proposals:

\begin{itemize}
    \item \textbf{Interpretability.}
    The interpretability of AI systems has become a big concern in recent years due to the ever-increasing ubiquity of such systems, especially in safety-critical applications. Despite advancements in the fields of DL and DRL to design more interpretable architectures such as Graph Neural Networks, the symbolic approach still has the upper hand when it comes to interpretability. Neurosymbolic methods allow to integrate the capabilities of DL and DRL to extract complex patterns from data with the interpretability of classical symbolic representations such as PDDL. They pose a promising approach towards building systems that not only solve tasks in an effective and efficient manner, but which are also able to explain their decisions (actions) to a human supervisor. Additionally, it would be interesting to explore representations not only interpretable by human experts, such as PDDL, but also by any human user, such as natural language.
    
    \item \textbf{Human-machine collaboration.}
    The current paradigm for problem solving entails building autonomous systems based on DL and DRL which try to rely on the user as little as possible. However, in most real-world scenarios, humans have access to crucial domain-specific knowledge which can greatly facilitate the learning process of the agent. The main difficulty in providing this information to the system comes from the fact that the knowledge representation commonly employed in modern AI, based on DNNs, is very different to that used by humans. Thus, in order to achieve effective human-machine collaboration, it is essential to reconcile these two representations. One possibility is provided once again by neurosymbolic methods, since these techniques utilize symbolic representations that are understandable by humans (experts). An alternative approach is to implement a communication interface capable of translating between the subsymbolic representation of the AI system and one understandable by humans (e.g., natural language). Regardless of the chosen method, the integration of both types of representation will make it possible to build AI systems that effectively communicate and cooperate with humans. For example, users will be able to easily provide prior knowledge to the agent and guide its behaviour during the task-solving process. Additionally, it would also be useful that the agent itself could query the human for assistance when needed, e.g., in order to escape a dead-end situation.
    
    \item \textbf{Goal Reasoning.}
    Goal Reasoning allows to create AI systems which are capable of reasoning about their own goals. This is similar to the way we humans think, as our actions are influenced by a set of goals, intentions, desires, etc. which are not fixed but rather vary with time. Thus, Goal Reasoning provides an ideal framework to design AI agents which collaborate with humans. Ideally, such a system should be able to function autonomously, being capable of formulating, selecting and pursuing the necessary goals to achieve the corresponding task. At the same time, the human user should be able to understand the behaviour and intentions of the system and, at any given moment, modify the goals the agent is pursuing. To build such a system, the neurosymbolic approach is a good fit. A neurosymbolic goal reasoning method could use DNNs to build a latent representation of the domain which allowed it to infer interesting goals, encoded in a symbolic representation. Then, a symbolic reasoning method, e.g., an automated planner, could reason about which goals should be pursued at each moment and the specific method for achieving them.

    \item \textbf{Symbolic value-equivalent action models.}
    Value-equivalent action models only encode those aspects of the environment dynamics which result useful for the task at hand. This is an essential requirement to learn action models of complex environments, such as those often encountered in real-world tasks, as it is infeasible to accurately depict every aspect of their dynamics. However, so far all value-equivalent action models employ a subsymbolic knowledge representation. We believe this value-equivalence principle would prove even more useful in the case of symbolic action models. Subsymbolic action models employ DNNs to encode their knowledge and, since neural networks are universal approximators, they can accurately represent any aspect of the environment regardless of how complex it is (although learning such a complex representation would require huge amounts of data and would have the risk of overfitting, thus the need for value-equivalent models). 
    { On the other hand, some complex environments may be hard (or even impossible) to accurately represent using a symbolic action model, in terms of a set of distinct objects and their relations/interactions.  
    Therefore, when confronted with such kind of environments, we should instead try to obtain a symbolic description of only those aspects of the dynamics that are needed to solve the task at hand, i.e., a symbolic value-equivalent model.}
    %On the other hand, symbolic action models use a much more coarse-grained representation based on abstract concepts, i.e., objects and their relations, which is not suitable to represent and reason about certain aspects of the environment, such as complex interdependencies between groups of numeric variables.  
    %Thus, if our goal is to effectively learn a symbolic action model of a complex environment, we must adapt the idea of value-equivalence to the symbolic realm, i.e., learn a symbolic value-equivalent model. Such a method would obtain an abstract, symbolic representation of the environment dynamics suitable for the task at hand, while ignoring those aspects which cannot be properly depicted in a symbolic manner.
    
    \item \textbf{Neurosymbolic action models.}
    The main idea behind symbolic value-equivalent action models is to find a symbolic description of the environment that is \textit{good enough} for the task at hand, even if it completely ignores some aspects about the dynamics. However, we might wonder if such a description exists in the first place. For some environments, a symbolic description of their dynamics may prove highly inaccurate and leave out aspects that result crucial to solve the task. For example, it would be infeasible to obtain a high-quality description of the dynamics of a complex physics simulator in PDDL and, thus, the resulting symbolic action model would not be of great use to solve any task in this domain. In these cases, those crucial aspects for which a good symbolic description cannot be obtained should be encoded subsymbolically. This neurosymbolic action model would employ a symbolic representation for those elements of the environment which can be properly encoded in such a way, and a subsymbolic representation for the rest. This approach could, in theory, combine the best of both worlds: the abstraction and interpretability of the symbolic representation with the accuracy and wide applicability of the subsymbolic representation. 
    
\end{itemize}

\section{Conclusions}
\label{section:conclusions}
% \textit{\textbf{Num max. de páginas: 2}}

{ In this work, we have provided an overview of the main symbolic, subsymbolic and hybrid methods for SDM, focusing on the case of finite MDPs. We have presented both techniques for obtaining solutions of MDPs and for learning information about their structure. We have classified methods for solving MDPs in those that obtain their solution via a reasoning process over a model of the world, known as AP, and those that learn their solution from data, known as RL. Additionally, we have described a new type of methods, capable of \textit{learning to plan}, which have emerged in recent years. 
We have classified methods for learning the structure of MDPs in those that learn an action model, i.e., a model of the MDP dynamics required by many MDP-solving methods such as AP and model-based RL, and those that learn additional structural aspects which are not encoded in the action model but may be useful to solve the MDP. We have coined the name \textit{domain exploitation} to refer to this second group of methods.
Afterwards, we have discussed what properties an ideal method for SDM should exhibit. Based on these properties, we have analysed the advantages and disadvantages of the different MDP-solving approaches presented in this work. As a result of our analysis, we have argued that an ideal method for SDM should integrate both paradigms for solving the MDP (the \textit{synthesize with a model} of AP versus the \textit{learn from data} of RL) and representing knowledge (symbolically versus subsymbolically). Since neurosymbolic AI actually performs this integration, we have concluded that it poses a promising approach towards achieving an ideal method for SDM.  Finally, we have proposed several lines of future work in order to advance the field of SDM via the unification of symbolic and subsymbolic AI.}

\section{Acknowledgements}
We want to thank Dr. Masataro Asai for his valuable feedback when writing this review.

\bibliographystyle{apalike}
\bibliography{References}

\begin{thebibliography}{}

\bibitem[Abbeel and Ng, 2004]{abbeel2004learning}
Abbeel, P. and Ng, A. (2004).
\newblock Learning first-order {M}arkov models for control.
\newblock {\em NeurIPS}, 17.

\bibitem[Acharya et~al., 2023]{acharya2023neurosymbolic}
Acharya, K., Raza, W., Dourado, C., Velasquez, A., and Song, H.~H. (2023).
\newblock Neurosymbolic reinforcement learning and planning: A survey.
\newblock {\em IEEE Trans. Artif. Intell.}

\bibitem[Aha, 2018]{aha2018goal}
Aha, D.~W. (2018).
\newblock Goal reasoning: Foundations, emerging applications, and prospects.
\newblock {\em AI Mag.}, 39(2):3--24.

\bibitem[Arora et~al., 2018]{arora2018review}
Arora, A., Fiorino, H., Pellier, D., M{\'e}tivier, M., and Pesty, S. (2018).
\newblock A review of learning planning action models.
\newblock {\em Knowl. Eng. Rev.}, 33:e20.

\bibitem[Arora and Doshi, 2021]{arora2021survey}
Arora, S. and Doshi, P. (2021).
\newblock A survey of inverse reinforcement learning: Challenges, methods and progress.
\newblock {\em Artif. Intell.}, 297:103500.

\bibitem[Asadi et~al., 2018]{asadi2018towards}
Asadi, K., Cater, E., Misra, D., and Littman, M.~L. (2018).
\newblock Towards a simple approach to multi-step model-based reinforcement learning.
\newblock {\em arXiv}.

\bibitem[Asai et~al., 2022]{asai2022classical}
Asai, M., Kajino, H., Fukunaga, A., and Muise, C. (2022).
\newblock Classical planning in deep latent space.
\newblock {\em JAIR}, 74:1599--1686.

\bibitem[Balduccini, 2011]{balduccini2011learning}
Balduccini, M. (2011).
\newblock Learning and using domain-specific heuristics in asp solvers.
\newblock {\em AI Commun.}, 24(2):147--164.

\bibitem[Battaglia et~al., 2016]{battaglia2016interaction}
Battaglia, P., Pascanu, R., Lai, M., Jimenez~Rezende, D., et~al. (2016).
\newblock Interaction networks for learning about objects, relations and physics.
\newblock {\em NeurIPS}, 29.

\bibitem[Battaglia et~al., 2018]{battaglia2018relational}
Battaglia, P.~W., Hamrick, J.~B., Bapst, V., Sanchez-Gonzalez, A., Zambaldi, V., Malinowski, M., Tacchetti, A., Raposo, D., Santoro, A., Faulkner, R., et~al. (2018).
\newblock Relational inductive biases, deep learning, and graph networks.
\newblock {\em arXiv}.

\bibitem[Bertsekas, 2019]{bertsekas2019reinforcement}
Bertsekas, D. (2019).
\newblock {\em Reinforcement learning and optimal control}.
\newblock Athena Scientific.

\bibitem[Blum and Furst, 1997]{blum1997fast}
Blum, A.~L. and Furst, M.~L. (1997).
\newblock Fast planning through planning graph analysis.
\newblock {\em Artif. Intell.}, 90:281--300.

\bibitem[Bonet and Geffner, 2003]{bonet2003labeled}
Bonet, B. and Geffner, H. (2003).
\newblock Labeled rtdp: Improving the convergence of real-time dynamic programming.
\newblock In {\em ICAPS}, volume~3, pages 12--21.

\bibitem[Botea et~al., 2005]{botea2005macro}
Botea, A., Enzenberger, M., M{\"u}ller, M., and Schaeffer, J. (2005).
\newblock Macro-ff: Improving ai planning with automatically learned macro-operators.
\newblock {\em JAIR}, 24:581--621.

\bibitem[Brown et~al., 2020]{brown2020language}
Brown, T., Mann, B., Ryder, N., Subbiah, M., Kaplan, J.~D., Dhariwal, P., Neelakantan, A., Shyam, P., Sastry, G., Askell, A., et~al. (2020).
\newblock Language models are few-shot learners.
\newblock {\em NeurIPS}, 33:1877--1901.

\bibitem[Browne et~al., 2012]{browne2012survey}
Browne, C.~B., Powley, E., Whitehouse, D., Lucas, S.~M., Cowling, P.~I., Rohlfshagen, P., Tavener, S., Perez, D., Samothrakis, S., and Colton, S. (2012).
\newblock A survey of monte carlo tree search methods.
\newblock {\em IEEE Trans. Comput. Intell. AI Games}, 4(1):1--43.

\bibitem[Bundy and Wallen, 1984]{bundy1984breadth}
Bundy, A. and Wallen, L. (1984).
\newblock Breadth-first search.
\newblock {\em Catalogue of artificial intelligence tools}, pages 13--13.

\bibitem[Cappart et~al., 2021]{cappart2021combinatorial}
Cappart, Q., Ch{\'{e}}telat, D., Khalil, E.~B., Lodi, A., Morris, C., and Velickovic, P. (2021).
\newblock Combinatorial optimization and reasoning with graph neural networks.
\newblock In {\em IJCAI}, pages 4348--4355.

\bibitem[Castillo et~al., 2006]{castillo2006efficiently}
Castillo, L.~A., Fern{\'a}ndez-Olivares, J., Garcia-Perez, O., and Palao, F. (2006).
\newblock Efficiently handling temporal knowledge in an htn planner.
\newblock In {\em ICAPS}, pages 63--72.

\bibitem[Cazenave, 2006]{cazenave2006optimizations}
Cazenave, T. (2006).
\newblock Optimizations of data structures, heuristics and algorithms for path-finding on maps.
\newblock In {\em 2006 IEEE Symp. Comp. Intell. Games}, pages 27--33.

\bibitem[Chakraborti et~al., 2020]{chakraborti2020emerging}
Chakraborti, T., Sreedharan, S., and Kambhampati, S. (2020).
\newblock The emerging landscape of explainable automated planning \& decision making.
\newblock In {\em IJCAI}, pages 4803--4811.

\bibitem[Chang et~al., 2017]{chang2017compositional}
Chang, M., Ullman, T.~D., Torralba, A., and Tenenbaum, J.~B. (2017).
\newblock A compositional object-based approach to learning physical dynamics.
\newblock In {\em ICLR}. OpenReview.net.

\bibitem[Charpentier et~al., 2021]{charpentier2021reinforcement}
Charpentier, A., Elie, R., and Remlinger, C. (2021).
\newblock Reinforcement learning in economics and finance.
\newblock {\em Comput. Econ.}, pages 1--38.

\bibitem[Chen et~al., 2021]{chen2021learning}
Chen, K., Srikanth, N.~S., Kent, D., Ravichandar, H., and Chernova, S. (2021).
\newblock Learning hierarchical task networks with preferences from unannotated demonstrations.
\newblock In {\em CoRL}, pages 1572--1581.

\bibitem[Chiappa et~al., 2017]{chiappa2017recurrent}
Chiappa, S., Racani{\`{e}}re, S., Wierstra, D., and Mohamed, S. (2017).
\newblock Recurrent environment simulators.
\newblock In {\em ICLR}. OpenReview.net.

\bibitem[Choi and Kim, 2011]{choi2011map}
Choi, J. and Kim, K.-E. (2011).
\newblock Map inference for bayesian inverse reinforcement learning.
\newblock {\em NeurIPS}, 24.

\bibitem[Chrisman, 1992]{chrisman1992reinforcement}
Chrisman, L. (1992).
\newblock Reinforcement learning with perceptual aliasing: The perceptual distinctions approach.
\newblock In {\em AAAI}, volume 1992, pages 183--188.

\bibitem[Coles and Smith, 2007]{coles2007marvin}
Coles, A.~I. and Smith, A.~J. (2007).
\newblock Marvin: A heuristic search planner with online macro-action learning.
\newblock {\em JAIR}, 28:119--156.

\bibitem[Deisenroth and Rasmussen, 2011]{deisenroth2011pilco}
Deisenroth, M. and Rasmussen, C.~E. (2011).
\newblock Pilco: A model-based and data-efficient approach to policy search.
\newblock In {\em ICML}, pages 465--472.

\bibitem[Depeweg et~al., 2017]{depeweg2017learning}
Depeweg, S., Hern{\'a}ndez-Lobato, J., Doshi-Velez, F., and Udluft, S. (2017).
\newblock Learning and policy search in stochastic dynamical systems with bayesian neural networks.
\newblock In {\em ICLR}. OpenReview.net.

\bibitem[Diuk et~al., 2008]{diuk2008object}
Diuk, C., Cohen, A., and Littman, M.~L. (2008).
\newblock An object-oriented representation for efficient reinforcement learning.
\newblock In {\em ICML}, pages 240--247.

\bibitem[Dong et~al., 2019]{dong2019neural}
Dong, H., Mao, J., Lin, T., Wang, C., Li, L., and Zhou, D. (2019).
\newblock Neural logic machines.
\newblock In {\em ICLR}. OpenReview.net.

\bibitem[Draper et~al., 1994]{draper1994probabilistic}
Draper, D., Hanks, S., and Weld, D.~S. (1994).
\newblock Probabilistic planning with information gathering and contingent execution.
\newblock In {\em AIPS}, pages 31--36.

\bibitem[D{\v{z}}eroski et~al., 2001]{dvzeroski2001relational}
D{\v{z}}eroski, S., De~Raedt, L., and Driessens, K. (2001).
\newblock Relational reinforcement learning.
\newblock {\em Mach. Learn.}, 43:7--52.

\bibitem[Edelkamp, 2001]{edelkamp2001planning}
Edelkamp, S. (2001).
\newblock Planning with pattern databases.
\newblock In {\em Proc. ECP}, volume~1, pages 13--24.

\bibitem[Edelkamp, 2007]{edelkamp2007automated}
Edelkamp, S. (2007).
\newblock Automated creation of pattern database search heuristics.
\newblock {\em Lect. Notes Comput. Sci.}, 4428:35.

\bibitem[Farquhar et~al., 2018]{farquhar2018treeqn}
Farquhar, G., Rockt\:{a}schel, T., Igl, M., and Whiteson, S. (2018).
\newblock Treeqn and atreec: Differentiable tree planning for deep reinforcement learning.
\newblock In {\em ICLR}.

\bibitem[Feng and Hansen, 2002]{feng2002symbolic}
Feng, Z. and Hansen, E.~A. (2002).
\newblock Symbolic heuristic search for factored {M}arkov decision processes.
\newblock In {\em AAAI/IAAI}, pages 455--460.

\bibitem[Fern et~al., 2004]{fern2004learning}
Fern, A., Yoon, S.~W., and Givan, R. (2004).
\newblock Learning domain-specific control knowledge from random walks.
\newblock In {\em ICAPS}, pages 191--199.

\bibitem[Feyzabadi and Carpin, 2017]{feyzabadi2017planning}
Feyzabadi, S. and Carpin, S. (2017).
\newblock Planning using hierarchical constrained {M}arkov decision processes.
\newblock {\em Auton. Robots}, 41:1589--1607.

\bibitem[Fikes et~al., 1972]{fikes1972learning}
Fikes, R.~E., Hart, P.~E., and Nilsson, N.~J. (1972).
\newblock Learning and executing generalized robot plans.
\newblock {\em Artif. Intell.}, 3:251--288.

\bibitem[Forestier et~al., 2022]{forestier2022intrinsically}
Forestier, S., Portelas, R., Mollard, Y., and Oudeyer, P.-Y. (2022).
\newblock Intrinsically motivated goal exploration processes with automatic curriculum learning.
\newblock {\em JMLR}, 23(1):6818--6858.

\bibitem[Fox and Long, 1998]{fox1998automatic}
Fox, M. and Long, D. (1998).
\newblock The automatic inference of state invariants in tim.
\newblock {\em JAIR}, 9:367--421.

\bibitem[Fuentetaja and De~la Rosa, 2012]{fuentetaja2012planning}
Fuentetaja, R. and De~la Rosa, T. (2012).
\newblock A planning-based approach for generating planning problems.
\newblock In {\em Workshops at AAAI}.

\bibitem[Garcez et~al., 2022]{garcez2022neural}
Garcez, A.~d., Bader, S., Bowman, H., Lamb, L.~C., de~Penning, L., Illuminoo, B., Poon, H., and Zaverucha, C.~G. (2022).
\newblock Neural-symbolic learning and reasoning: a survey and interpretation.
\newblock {\em Neuro-Symbolic Artificial Intelligence: The State of the Art}, 342.

\bibitem[Garc{\i}a and Fern{\'a}ndez, 2015]{garcia2015comprehensive}
Garc{\i}a, J. and Fern{\'a}ndez, F. (2015).
\newblock A comprehensive survey on safe reinforcement learning.
\newblock {\em JMLR}, 16:1437--1480.

\bibitem[Garnelo et~al., 2016]{garnelo2016towards}
Garnelo, M., Arulkumaran, K., and Shanahan, M. (2016).
\newblock Towards deep symbolic reinforcement learning.
\newblock {\em arXiv}.

\bibitem[Garnelo and Shanahan, 2019]{garnelo2019reconciling}
Garnelo, M. and Shanahan, M. (2019).
\newblock Reconciling deep learning with symbolic artificial intelligence: representing objects and relations.
\newblock {\em Curr. Opin. Behav. Sci.}, 29:17--23.

\bibitem[Gehring et~al., 2022]{gehring2022reinforcement}
Gehring, C., Asai, M., Chitnis, R., Silver, T., Kaelbling, L., Sohrabi, S., and Katz, M. (2022).
\newblock Reinforcement learning for classical planning: Viewing heuristics as dense reward generators.
\newblock In {\em ICAPS}, volume~32, pages 588--596.

\bibitem[Georgievski and Aiello, 2015]{georgievski2015htn}
Georgievski, I. and Aiello, M. (2015).
\newblock Htn planning: Overview, comparison, and beyond.
\newblock {\em Artif. Intell.}, 222:124--156.

\bibitem[Gerevini and Schubert, 1998]{gerevini1998inferring}
Gerevini, A. and Schubert, L. (1998).
\newblock Inferring state constraints for domain-independent planning.
\newblock In {\em AAAI}, pages 905--912.

\bibitem[Ghallab et~al., 2016]{ghallab2016automated}
Ghallab, M., Nau, D., and Traverso, P. (2016).
\newblock {\em Automated planning and acting}.
\newblock Cambridge University Press.

\bibitem[Grimm et~al., 2020]{grimm2020value}
Grimm, C., Barreto, A., Singh, S., and Silver, D. (2020).
\newblock The value equivalence principle for model-based reinforcement learning.
\newblock {\em NeurIPS}, 33:5541--5552.

\bibitem[Gu et~al., 2016]{gu2016continuous}
Gu, S., Lillicrap, T., Sutskever, I., and Levine, S. (2016).
\newblock Continuous deep q-learning with model-based acceleration.
\newblock In {\em ICML}, pages 2829--2838.

\bibitem[Guez et~al., 2019]{guez2019investigation}
Guez, A., Mirza, M., Gregor, K., Kabra, R., Racani{\`e}re, S., Weber, T., Raposo, D., Santoro, A., Orseau, L., Eccles, T., et~al. (2019).
\newblock An investigation of model-free planning.
\newblock In {\em ICML}, pages 2464--2473.

\bibitem[Guez et~al., 2018]{guez2018learning}
Guez, A., Weber, T., Antonoglou, I., Simonyan, K., Vinyals, O., Wierstra, D., Munos, R., and Silver, D. (2018).
\newblock Learning to search with {MCTS}nets.
\newblock In {\em ICML}, pages 1822--1831.

\bibitem[Gunning and Aha, 2019]{gunning2019darpa}
Gunning, D. and Aha, D. (2019).
\newblock Darpa’s explainable artificial intelligence (xai) program.
\newblock {\em AI Mag.}, 40:44--58.

\bibitem[Hans et~al., 2008]{hans2008safe}
Hans, A., Schneega{\ss}, D., Sch{\"a}fer, A.~M., and Udluft, S. (2008).
\newblock Safe exploration for reinforcement learning.
\newblock In {\em ESANN}, pages 143--148.

\bibitem[Hansen and Zilberstein, 2001]{hansen2001lao}
Hansen, E.~A. and Zilberstein, S. (2001).
\newblock Lao*: A heuristic search algorithm that finds solutions with loops.
\newblock {\em Artif. Intell.}, 129(1-2):35--62.

\bibitem[Hart et~al., 1968]{hart1968formal}
Hart, P.~E., Nilsson, N.~J., and Raphael, B. (1968).
\newblock A formal basis for the heuristic determination of minimum cost paths.
\newblock {\em IEEE Trans. Syst. Sci. Cybern.}, 4(2):100--107.

\bibitem[Haslum et~al., 2019]{haslum2019introduction}
Haslum, P., Lipovetzky, N., Magazzeni, D., and Muise, C. (2019).
\newblock An introduction to the planning domain definition language.
\newblock {\em Synth. Lect. Artif. Intell. Mach. Learn.}, 13:1--187.

\bibitem[Hausknecht and Stone, 2015]{hausknecht2015deep}
Hausknecht, M. and Stone, P. (2015).
\newblock Deep recurrent q-learning for partially observable mdps.
\newblock In {\em AAAI}.

\bibitem[Heger, 1994]{heger1994consideration}
Heger, M. (1994).
\newblock Consideration of risk in reinforcement learning.
\newblock In {\em Mach. Learn.}, pages 105--111. Elsevier.

\bibitem[Helmert, 2006]{helmert2006fast}
Helmert, M. (2006).
\newblock The fast downward planning system.
\newblock {\em JAIR}, 26:191--246.

\bibitem[Helmert, 2009]{helmert2009concise}
Helmert, M. (2009).
\newblock Concise finite-domain representations for pddl planning tasks.
\newblock {\em Artif. Intell.}, 173:503--535.

\bibitem[Hester and Stone, 2013]{hester2013texplore}
Hester, T. and Stone, P. (2013).
\newblock Texplore: real-time sample-efficient reinforcement learning for robots.
\newblock {\em Mach. Learn.}, 90:385--429.

\bibitem[Hickling et~al., 2023]{hickling2023explainability}
Hickling, T., Zenati, A., Aouf, N., and Spencer, P. (2023).
\newblock Explainability in deep reinforcement learning: A review into current methods and applications.
\newblock {\em ACM Comput. Surv.}, 56(5):1--35.

\bibitem[Hoey et~al., 1999]{hoey1999spudd}
Hoey, J., St-Aubin, R., Hu, A., and Boutilier, C. (1999).
\newblock Spudd: stochastic planning using decision diagrams.
\newblock In {\em UAI}, pages 279--288.

\bibitem[Hoffmann, 2001]{hoffmann2001ff}
Hoffmann, J. (2001).
\newblock Ff: The fast-forward planning system.
\newblock {\em AI Mag.}, 22(3):57--57.

\bibitem[Hoffmann et~al., 2004]{hoffmann2004ordered}
Hoffmann, J., Porteous, J., and Sebastia, L. (2004).
\newblock Ordered landmarks in planning.
\newblock {\em JAIR}, 22:215--278.

\bibitem[Hogg et~al., 2008]{hogg2008htn}
Hogg, C., Munoz-Avila, H., and Kuter, U. (2008).
\newblock Htn-maker: Learning htns with minimal additional knowledge engineering required.
\newblock In {\em AAAI}, pages 950--956.

\bibitem[H{\"o}ller and Bercher, 2021]{holler2021landmark}
H{\"o}ller, D. and Bercher, P. (2021).
\newblock Landmark generation in htn planning.
\newblock In {\em AAAI}, volume~35, pages 11826--11834.

\bibitem[Howey et~al., 2004]{howey2004val}
Howey, R., Long, D., and Fox, M. (2004).
\newblock Val: Automatic plan validation, continuous effects and mixed initiative planning using pddl.
\newblock In {\em IEEE Int. Conf. Tools Artif. Intell.}, pages 294--301.

\bibitem[Jim{\'e}nez et~al., 2012]{jimenez2012review}
Jim{\'e}nez, S., De~La~Rosa, T., Fern{\'a}ndez, S., Fern{\'a}ndez, F., and Borrajo, D. (2012).
\newblock A review of machine learning for automated planning.
\newblock {\em Knowl. Eng. Rev.}, 27(4):433--467.

\bibitem[Jim{\'e}nez et~al., 2008]{jimenez2008pela}
Jim{\'e}nez, S., Fern{\'a}ndez, F., and Borrajo, D. (2008).
\newblock The {PELA} architecture: integrating planning and learning to improve execution.
\newblock In {\em AAAI}. {AAAI} Press.

\bibitem[Jim{\'e}nez et~al., 2019]{jimenez2019review}
Jim{\'e}nez, S., Segovia-Aguas, J., and Jonsson, A. (2019).
\newblock A review of generalized planning.
\newblock {\em Knowl. Eng. Rev.}, 34:e5.

\bibitem[Jin et~al., 2022]{jin2022creativity}
Jin, M., Ma, Z., Jin, K., Zhuo, H.~H., Chen, C., and Yu, C. (2022).
\newblock Creativity of ai: Automatic symbolic option discovery for facilitating deep reinforcement learning.
\newblock In {\em AAAI}, volume~36, pages 7042--7050.

\bibitem[Kaiser et~al., 2020]{kaiser2020model}
Kaiser, {\L}., Babaeizadeh, M., Mi{\l}os, P., Osi{\'n}ski, B., Campbell, R.~H., Czechowski, K., Erhan, D., Finn, C., Kozakowski, P., Levine, S., et~al. (2020).
\newblock Model based reinforcement learning for atari.
\newblock In {\em ICLR}. OpenReview.net.

\bibitem[Kansky et~al., 2017]{kansky2017schema}
Kansky, K., Silver, T., M{\'e}ly, D.~A., Eldawy, M., L{\'a}zaro-Gredilla, M., Lou, X., Dorfman, N., Sidor, S., Phoenix, S., and George, D. (2017).
\newblock Schema networks: Zero-shot transfer with a generative causal model of intuitive physics.
\newblock In {\em ICML}, pages 1809--1818.

\bibitem[Karpas and Domshlak, 2009]{karpas2009cost}
Karpas, E. and Domshlak, C. (2009).
\newblock Cost-optimal planning with landmarks.
\newblock In {\em IJCAI}, pages 1728--1733.

\bibitem[Katz and Sohrabi, 2020]{katz2020generating}
Katz, M. and Sohrabi, S. (2020).
\newblock Generating data in planning: Sas planning tasks of a given causal structure.
\newblock {\em HSDIP}, page~41.

\bibitem[Katz et~al., 2021]{katz2021scenario}
Katz, M., Srinivas, K., Sohrabi, S., Feblowitz, M., Udrea, O., and Hassanzadeh, O. (2021).
\newblock Scenario planning in the wild: A neuro-symbolic approach.
\newblock {\em FinPlan}, 15.

\bibitem[Keyder et~al., 2014]{keyder2014improving}
Keyder, E., Hoffmann, J., and Haslum, P. (2014).
\newblock Improving delete relaxation heuristics through explicitly represented conjunctions.
\newblock {\em JAIR}, 50:487--533.

\bibitem[Khansari-Zadeh and Billard, 2011]{khansari2011learning}
Khansari-Zadeh, S.~M. and Billard, A. (2011).
\newblock Learning stable nonlinear dynamical systems with gaussian mixture models.
\newblock {\em IEEE Trans. Robot.}, 27(5):943--957.

\bibitem[Kingma and Welling, 2014]{Kingma2014auto}
Kingma, D. and Welling, M. (2014).
\newblock Auto-encoding variational bayes international.
\newblock In {\em ICLR}.

\bibitem[Kipf et~al., 2020]{kipf2020contrastive}
Kipf, T.~N., van~der Pol, E., and Welling, M. (2020).
\newblock Contrastive learning of structured world models.
\newblock In {\em ICLR}. OpenReview.net.

\bibitem[Kober et~al., 2013]{kober2013reinforcement}
Kober, J., Bagnell, J.~A., and Peters, J. (2013).
\newblock Reinforcement learning in robotics: A survey.
\newblock {\em Int. J. Rob. Res.}, 32(11):1238--1274.

\bibitem[Koller et~al., 2007]{koller2007introduction}
Koller, D., Friedman, N., D{\v{z}}eroski, S., Sutton, C., McCallum, A., Pfeffer, A., Abbeel, P., Wong, M.-F., Meek, C., Neville, J., et~al. (2007).
\newblock {\em Introduction to statistical relational learning}.
\newblock MIT press.

\bibitem[Konidaris and Barto, 2007]{konidaris2007building}
Konidaris, G.~D. and Barto, A.~G. (2007).
\newblock Building portable options: Skill transfer in reinforcement learning.
\newblock In {\em IJCAI}, volume~7, pages 895--900.

\bibitem[Koppejan and Whiteson, 2011]{koppejan2011neuroevolutionary}
Koppejan, R. and Whiteson, S. (2011).
\newblock Neuroevolutionary reinforcement learning for generalized control of simulated helicopters.
\newblock {\em Evol. Intell.}, 4:219--241.

\bibitem[Korf, 1985]{korf1985macro}
Korf, R.~E. (1985).
\newblock Macro-operators: A weak method for learning.
\newblock {\em Artif. Intell.}, 26:35--77.

\bibitem[Kramer, 1996]{kramer1996structural}
Kramer, S. (1996).
\newblock Structural regression trees.
\newblock In {\em AAAI}, pages 812--819.

\bibitem[Krizhevsky et~al., 2017]{krizhevsky2017imagenet}
Krizhevsky, A., Sutskever, I., and Hinton, G.~E. (2017).
\newblock Imagenet classification with deep convolutional neural networks.
\newblock {\em ACM Commun.}, 60:84--90.

\bibitem[Landajuela et~al., 2021]{landajuela2021discovering}
Landajuela, M., Petersen, B.~K., Kim, S., Santiago, C.~P., Glatt, R., Mundhenk, N., Pettit, J.~F., and Faissol, D. (2021).
\newblock Discovering symbolic policies with deep reinforcement learning.
\newblock In {\em ICML}, pages 5979--5989.

\bibitem[Laversanne-Finot et~al., 2018]{laversanne2018curiosity}
Laversanne-Finot, A., Pere, A., and Oudeyer, P.-Y. (2018).
\newblock Curiosity driven exploration of learned disentangled goal spaces.
\newblock In {\em CoRL}, pages 487--504.

\bibitem[LeCun et~al., 2015]{lecun2015deep}
LeCun, Y., Bengio, Y., and Hinton, G. (2015).
\newblock Deep learning.
\newblock {\em nature}, 521:436--444.

\bibitem[Li, 2018]{li2018deep}
Li, Y. (2018).
\newblock Deep reinforcement learning.
\newblock {\em arXiv}.

\bibitem[Littman, 1996]{littman1996algorithms}
Littman, M.~L. (1996).
\newblock {\em Algorithms for sequential decision-making}.
\newblock Brown University.

\bibitem[Long and Fox, 1999]{long1999efficient}
Long, D. and Fox, M. (1999).
\newblock Efficient implementation of the plan graph in stan.
\newblock {\em JAIR}, 10:87--115.

\bibitem[Lovejoy, 1991]{lovejoy1991survey}
Lovejoy, W.~S. (1991).
\newblock A survey of algorithmic methods for partially observed {M}arkov decision processes.
\newblock {\em Ann. Oper. Res.}, 28(1):47--65.

\bibitem[Lyu et~al., 2019]{lyu2019sdrl}
Lyu, D., Yang, F., Liu, B., and Gustafson, S. (2019).
\newblock Sdrl: interpretable and data-efficient deep reinforcement learning leveraging symbolic planning.
\newblock In {\em AAAI}, volume~33, pages 2970--2977.

\bibitem[Machado et~al., 2017]{machado2017laplacian}
Machado, M.~C., Bellemare, M.~G., and Bowling, M. (2017).
\newblock A laplacian framework for option discovery in reinforcement learning.
\newblock In {\em ICML}, pages 2295--2304.

\bibitem[Magnaguagno et~al., 2017]{magnaguagno2017web}
Magnaguagno, M.~C., FRAGA~PEREIRA, R., M{\'o}re, M.~D., and Meneguzzi, F.~R. (2017).
\newblock Web planner: A tool to develop classical planning domains and visualize heuristic state-space search.
\newblock In {\em ICAPS UISP Workshop}.

\bibitem[Marcus, 2018]{marcus2018deep}
Marcus, G. (2018).
\newblock Deep learning: A critical appraisal.
\newblock {\em arXiv}.

\bibitem[McGovern and Sutton, 1998]{mcgovern1998macro}
McGovern, A. and Sutton, R.~S. (1998).
\newblock Macro-actions in reinforcement learning: An empirical analysis.
\newblock {\em Computer Science Department Faculty Publication Series}, page~15.

\bibitem[Mnih et~al., 2016]{mnih2016asynchronous}
Mnih, V., Badia, A.~P., Mirza, M., Graves, A., Lillicrap, T., Harley, T., Silver, D., and Kavukcuoglu, K. (2016).
\newblock Asynchronous methods for deep reinforcement learning.
\newblock In {\em ICML}, pages 1928--1937.

\bibitem[Mnih et~al., 2013]{mnih2013playing}
Mnih, V., Kavukcuoglu, K., Silver, D., Graves, A., Antonoglou, I., Wierstra, D., and Riedmiller, M. (2013).
\newblock Playing atari with deep reinforcement learning.
\newblock {\em arXiv}.

\bibitem[Moerland et~al., 2020]{moerland2020framework}
Moerland, T.~M., Broekens, J., and Jonker, C.~M. (2020).
\newblock A framework for reinforcement learning and planning.
\newblock {\em arXiv}.

\bibitem[Moerland et~al., 2023]{moerland2023model}
Moerland, T.~M., Broekens, J., Plaat, A., Jonker, C.~M., et~al. (2023).
\newblock Model-based reinforcement learning: A survey.
\newblock {\em Found. Trends Mach. Learn.}, 16(1):1--118.

\bibitem[Molineaux et~al., 2010]{molineaux2010goal}
Molineaux, M., Klenk, M., and Aha, D. (2010).
\newblock Goal-driven autonomy in a navy strategy simulation.
\newblock In {\em AAAI}, volume~24, pages 1548--1554.

\bibitem[Mourao et~al., 2008]{mourao2008using}
Mourao, K., Petrick, R.~P., and Steedman, M. (2008).
\newblock Using kernel perceptrons to learn action effects for planning.
\newblock In {\em CogSys}, pages 45--50.

\bibitem[Natarajan and Kolobov, 2022]{natarajan2022planning}
Natarajan, M. and Kolobov, A. (2022).
\newblock {\em Planning with {M}arkov decision processes: An AI perspective}.
\newblock Synth. Lect. Artif. Intell. Mach. Learn. Springer Nature.

\bibitem[Nau et~al., 2003]{nau2003shop2}
Nau, D.~S., Au, T.-C., Ilghami, O., Kuter, U., Murdock, J.~W., Wu, D., and Yaman, F. (2003).
\newblock Shop2: An htn planning system.
\newblock {\em JAIR}, 20:379--404.

\bibitem[Nejati et~al., 2006]{nejati2006learning}
Nejati, N., Langley, P., and Konik, T. (2006).
\newblock Learning hierarchical task networks by observation.
\newblock In {\em ICML}, pages 665--672.

\bibitem[Ng et~al., 2000]{ng2000algorithms}
Ng, A.~Y., Russell, S., et~al. (2000).
\newblock Algorithms for inverse reinforcement learning.
\newblock In {\em ICML}, volume~1, page~2.

\bibitem[N{\'u}{\~n}ez-Molina et~al., 2022]{nunez2022learning}
N{\'u}{\~n}ez-Molina, C., Fern{\'a}ndez-Olivares, J., and P{\'e}rez, R. (2022).
\newblock Learning to select goals in automated planning with deep-q learning.
\newblock {\em Expert Syst. Appl.}, 202:117265.

\bibitem[N{\'u}{\~n}ez-Molina et~al., 2023]{nunez2023nesig}
N{\'u}{\~n}ez-Molina, C., Mesejo, P., and Fern{\'a}ndez-Olivares, J. (2023).
\newblock Nesig: A neuro-symbolic method for learning to generate planning problems.
\newblock {\em arXiv}.

\bibitem[Oates and Cohen, 1996]{oates1996searching}
Oates, T. and Cohen, P.~R. (1996).
\newblock Searching for planning operators with context-dependent and probabilistic effects.
\newblock In {\em AAAI}, pages 863--868.

\bibitem[Oh et~al., 2017]{junhyuk2017value}
Oh, J., Singh, S., and Lee, H. (2017).
\newblock Value prediction network.
\newblock {\em Advances in neural information processing systems}, 30.

\bibitem[Partalas et~al., 2008]{partalas2008reinforcement}
Partalas, I., Vrakas, D., and Vlahavas, I. (2008).
\newblock Reinforcement learning and automated planning: A survey.
\newblock In {\em Artificial Intelligence for Advanced Problem Solving Techniques}, pages 148--165. IGI Global.

\bibitem[Pascanu et~al., 2017]{pascanu2017learning}
Pascanu, R., Li, Y., Vinyals, O., Heess, N., Buesing, L., Racani{\`e}re, S., Reichert, D., Weber, T., Wierstra, D., and Battaglia, P. (2017).
\newblock Learning model-based planning from scratch.
\newblock {\em arXiv}.

\bibitem[Pasula et~al., 2007]{pasula2007learning}
Pasula, H.~M., Zettlemoyer, L.~S., and Kaelbling, L.~P. (2007).
\newblock Learning symbolic models of stochastic domains.
\newblock {\em JAIR}, 29:309--352.

\bibitem[Pereira et~al., 2020]{pereira2020landmark}
Pereira, R.~F., Oren, N., and Meneguzzi, F. (2020).
\newblock Landmark-based approaches for goal recognition as planning.
\newblock {\em Artif. Intell.}, 279:103217.

\bibitem[Plaat et~al., 2023]{plaat2023high}
Plaat, A., Kosters, W., and Preuss, M. (2023).
\newblock High-accuracy model-based reinforcement learning, a survey.
\newblock {\em Artif. Intell. Rev.}, pages 1--33.

\bibitem[Pozanco et~al., 2018]{pozanco2018learning}
Pozanco, A., Fern{\'a}ndez, S., and Borrajo, D. (2018).
\newblock Learning-driven goal generation.
\newblock {\em AI Commun.}, 31(2):137--150.

\bibitem[Ram{\i}rez and Geffner, 2009]{ramirez2009plan}
Ram{\i}rez, M. and Geffner, H. (2009).
\newblock Plan recognition as planning.
\newblock In {\em IJCAI}, pages 1778--1783.

\bibitem[Rintanen, 2008]{rintanen2008regression}
Rintanen, J. (2008).
\newblock Regression for classical and nondeterministic planning.
\newblock In {\em ECAI}, pages 568--572. IOS Press.

\bibitem[Rovner et~al., 2019]{rovner2019counterexample}
Rovner, A., Sievers, S., and Helmert, M. (2019).
\newblock Counterexample-guided abstraction refinement for pattern selection in optimal classical planning.
\newblock In {\em ICAPS}, volume~29, pages 362--367.

\bibitem[Russell and Norvig, 2020]{russell2020artificial}
Russell, S.~J. and Norvig, P. (2020).
\newblock {\em Artificial Intelligence: {A} Modern Approach (4th Edition)}.
\newblock Pearson.

\bibitem[Sacerdoti, 1975]{sacerdoti1975nonlinear}
Sacerdoti, E.~D. (1975).
\newblock The nonlinear nature of plans.
\newblock Technical report, Stanford Research Inst. Menlo Park CA.

\bibitem[Safaei and Ghassem-Sani, 2007]{safaei2007incremental}
Safaei, J. and Ghassem-Sani, G. (2007).
\newblock Incremental learning of planning operators in stochastic domains.
\newblock In {\em SOFSEM}, pages 644--655.

\bibitem[Sanner et~al., 2010]{sanner2010relational}
Sanner, S. et~al. (2010).
\newblock Relational dynamic influence diagram language (rddl): Language description.
\newblock {\em Unpublished ms. Australian National University}, 32:27.

\bibitem[Sch{\"a}pers et~al., 2018]{schapers2018asp}
Sch{\"a}pers, B., Niemueller, T., Lakemeyer, G., Gebser, M., and Schaub, T. (2018).
\newblock Asp-based time-bounded planning for logistics robots.
\newblock In {\em ICAPS}, volume~28, pages 509--517.

\bibitem[Schrittwieser et~al., 2020]{schrittwieser2020mastering}
Schrittwieser, J., Antonoglou, I., Hubert, T., Simonyan, K., Sifre, L., Schmitt, S., Guez, A., Lockhart, E., Hassabis, D., Graepel, T., et~al. (2020).
\newblock Mastering atari, go, chess and shogi by planning with a learned model.
\newblock {\em Nature}, 588(7839):604--609.

\bibitem[Segura-Muros et~al., 2021]{segura2021discovering}
Segura-Muros, J.~{\'A}., P{\'e}rez, R., and Fern{\'a}ndez-Olivares, J. (2021).
\newblock Discovering relational and numerical expressions from plan traces for learning action models.
\newblock {\em Appl. Intell.}, 51:7973--7989.

\bibitem[Shakya et~al., 2023]{shakya2023reinforcement}
Shakya, A.~K., Pillai, G., and Chakrabarty, S. (2023).
\newblock Reinforcement learning algorithms: A brief survey.
\newblock {\em Expert Syst. Appl.}, page 120495.

\bibitem[Shen et~al., 2020]{shen2020learning}
Shen, W., Trevizan, F., and Thi{\'e}baux, S. (2020).
\newblock Learning domain-independent planning heuristics with hypergraph networks.
\newblock In {\em ICAPS}, volume~30, pages 574--584.

\bibitem[Shen and Simon, 1989]{shen1989rule}
Shen, W.~M. and Simon, H.~A. (1989).
\newblock Rule creation and rule learning through environmental exploration.
\newblock In {\em IJCAI}, pages 675--680. Morgan Kaufmann.

\bibitem[Shi et~al., 2015]{shi2015convolutional}
Shi, X., Chen, Z., Wang, H., Yeung, D.-Y., Wong, W.-K., and Woo, W.-c. (2015).
\newblock Convolutional lstm network: A machine learning approach for precipitation nowcasting.
\newblock {\em NeurIPS}, 28.

\bibitem[Silver et~al., 2017]{silver2017predictron}
Silver, D., Hasselt, H., Hessel, M., Schaul, T., Guez, A., Harley, T., Dulac-Arnold, G., Reichert, D., Rabinowitz, N., Barreto, A., et~al. (2017).
\newblock The predictron: End-to-end learning and planning.
\newblock In {\em ICML}, pages 3191--3199.

\bibitem[Silver et~al., 2018]{silver2018general}
Silver, D., Hubert, T., Schrittwieser, J., Antonoglou, I., Lai, M., Guez, A., Lanctot, M., Sifre, L., Kumaran, D., Graepel, T., et~al. (2018).
\newblock A general reinforcement learning algorithm that masters chess, shogi, and go through self-play.
\newblock {\em Science}, 362(6419):1140--1144.

\bibitem[Singh et~al., 1994]{singh1994reinforcement}
Singh, S., Jaakkola, T., and Jordan, M. (1994).
\newblock Reinforcement learning with soft state aggregation.
\newblock {\em NeurIPS}, 7.

\bibitem[Sohrabi et~al., 2016]{sohrabi2016plan}
Sohrabi, S., Riabov, A.~V., and Udrea, O. (2016).
\newblock Plan recognition as planning revisited.
\newblock In {\em IJCAI}, pages 3258--3264.

\bibitem[Srinivas et~al., 2018]{srinivas2018universal}
Srinivas, A., Jabri, A., Abbeel, P., Levine, S., and Finn, C. (2018).
\newblock Universal planning networks: Learning generalizable representations for visuomotor control.
\newblock In {\em ICML}, pages 4732--4741.

\bibitem[Stolle and Precup, 2002]{stolle2002learning}
Stolle, M. and Precup, D. (2002).
\newblock Learning options in reinforcement learning.
\newblock In {\em SARA}, pages 212--223.

\bibitem[Sutton, 1991]{sutton1991dyna}
Sutton, R.~S. (1991).
\newblock Dyna, an integrated architecture for learning, planning, and reacting.
\newblock {\em ACM Sigart Bulletin}, 2(4):160--163.

\bibitem[Sutton and Barto, 2018]{sutton2018reinforcement}
Sutton, R.~S. and Barto, A.~G. (2018).
\newblock {\em Reinforcement learning: An introduction}.
\newblock MIT press.

\bibitem[Sutton et~al., 1999]{sutton1999between}
Sutton, R.~S., Precup, D., and Singh, S. (1999).
\newblock Between mdps and semi-mdps: A framework for temporal abstraction in reinforcement learning.
\newblock {\em Artif. Intell.}, 112(1-2):181--211.

\bibitem[Sutton et~al., 2008]{sutton2008dyna}
Sutton, R.~S., Szepesv{\'{a}}ri, C., Geramifard, A., and Bowling, M.~H. (2008).
\newblock Dyna-style planning with linear function approximation and prioritized sweeping.
\newblock In {\em UAI}, pages 528--536. {AUAI} Press.

\bibitem[Tadepalli et~al., 2004]{tadepalli2004relational}
Tadepalli, P., Givan, R., and Driessens, K. (2004).
\newblock Relational reinforcement learning: An overview.
\newblock In {\em ICML workshop on relational reinforcement learning}, pages 1--9.

\bibitem[Tamar et~al., 2016]{tamar2016value}
Tamar, A., Wu, Y., Thomas, G., Levine, S., and Abbeel, P. (2016).
\newblock Value iteration networks.
\newblock {\em NeurIPS}, 29.

\bibitem[Tarjan, 1972]{tarjan1972depth}
Tarjan, R. (1972).
\newblock Depth-first search and linear graph algorithms.
\newblock {\em SIAM J. Comput.}, 1(2):146--160.

\bibitem[Tate, 1977]{tate1977generating}
Tate, A. (1977).
\newblock Generating project networks.
\newblock In {\em IJCAI}, pages 888--893.

\bibitem[Thomaz et~al., 2006]{thomaz2006reinforcement}
Thomaz, A.~L., Breazeal, C., et~al. (2006).
\newblock Reinforcement learning with human teachers: Evidence of feedback and guidance with implications for learning performance.
\newblock In {\em AAAI}, volume~6, pages 1000--1005.

\bibitem[Torralba et~al., 2021]{torralba2021automatic}
Torralba, A., Seipp, J., and Sievers, S. (2021).
\newblock Automatic instance generation for classical planning.
\newblock In {\em ICAPS}, volume~31, pages 376--384.

\bibitem[Toyer et~al., 2018]{toyer2018action}
Toyer, S., Trevizan, F., Thi{\'e}baux, S., and Xie, L. (2018).
\newblock Action schema networks: Generalised policies with deep learning.
\newblock In {\em AAAI}, volume~32.

\bibitem[Trevizan and Veloso, 2014]{trevizan2014depth}
Trevizan, F.~W. and Veloso, M.~M. (2014).
\newblock Depth-based short-sighted stochastic shortest path problems.
\newblock {\em Artif. Intell.}, 216:179--205.

\bibitem[{\'u}s~Virseda et~al., 2013]{us2013learning}
{\'u}s~Virseda, J., Borrajo, D., and Alc{\'a}zar, V. (2013).
\newblock Learning heuristic functions for cost-based planning.
\newblock {\em Planning and Learning}, 4.

\bibitem[Vallati et~al., 2015]{vallati20152014}
Vallati, M., Chrpa, L., Grze{\'s}, M., McCluskey, T.~L., Roberts, M., Sanner, S., et~al. (2015).
\newblock The 2014 international planning competition: Progress and trends.
\newblock {\em Ai Mag.}, 36(3):90--98.

\bibitem[Vaswani et~al., 2017]{vaswani2017attention}
Vaswani, A., Shazeer, N., Parmar, N., Uszkoreit, J., Jones, L., Gomez, A.~N., Kaiser, {\L}., and Polosukhin, I. (2017).
\newblock Attention is all you need.
\newblock {\em NeurIPS}, 30.

\bibitem[Wahlstr{\"o}m et~al., 2015]{wahlstrom2015pixels}
Wahlstr{\"o}m, N., Sch{\"o}n, T.~B., and Deisenroth, M.~P. (2015).
\newblock From pixels to torques: Policy learning with deep dynamical models.
\newblock {\em arXiv}.

\bibitem[Walsh and Littman, 2008]{walsh2008efficient}
Walsh, T.~J. and Littman, M.~L. (2008).
\newblock Efficient learning of action schemas and web-service descriptions.
\newblock In {\em AAAI}, volume~8, pages 714--719.

\bibitem[Wang et~al., 2018]{wang2018deep}
Wang, W.~Y., Li, J., and He, X. (2018).
\newblock Deep reinforcement learning for nlp.
\newblock In {\em ACL: Tutorial Abstracts}, pages 19--21.

\bibitem[Wang, 1996]{wang1996learning}
Wang, X. (1996).
\newblock {\em Learning planning operators by observation and practice}.
\newblock PhD thesis, Carnegie Mellon University.

\bibitem[Watkins, 1989]{watkins1989learning}
Watkins, C. J. C.~H. (1989).
\newblock {\em Learning from delayed rewards}.
\newblock PhD thesis, King's College.

\bibitem[Weber et~al., 2012]{weber2012learning}
Weber, B., Mateas, M., and Jhala, A. (2012).
\newblock Learning from demonstration for goal-driven autonomy.
\newblock In {\em AAAI}, volume~26, pages 1176--1182.

\bibitem[Williams, 1992]{williams1992simple}
Williams, R.~J. (1992).
\newblock Simple statistical gradient-following algorithms for connectionist reinforcement learning.
\newblock {\em Reinforcement learning}, pages 5--32.

\bibitem[Yang et~al., 2007]{yang2007learning}
Yang, Q., Wu, K., and Jiang, Y. (2007).
\newblock Learning action models from plan examples using weighted max-sat.
\newblock {\em Artif. Intell.}, 171(2-3):107--143.

\bibitem[Yoon and Kambhampati, 2007]{yoon2007towards}
Yoon, S. and Kambhampati, S. (2007).
\newblock Towards model-lite planning: A proposal for learning \& planning with incomplete domain models.
\newblock In {\em ICAPS Workshop on Artificial Intelligence Planning and Learning}.

\bibitem[Yoon et~al., 2006]{yoon2006learning}
Yoon, S.~W., Fern, A., and Givan, R. (2006).
\newblock Learning heuristic functions from relaxed plans.
\newblock In {\em ICAPS}, volume~2, page~3.

\bibitem[Yoon et~al., 2007]{yoon2007ff}
Yoon, S.~W., Fern, A., and Givan, R. (2007).
\newblock Ff-replan: A baseline for probabilistic planning.
\newblock In {\em ICAPS}, volume~7, pages 352--359.

\bibitem[Younes and Littman, 2004]{younes2004ppddl1}
Younes, H.~L. and Littman, M.~L. (2004).
\newblock Ppddl1. 0: An extension to pddl for expressing planning domains with probabilistic effects.
\newblock {\em Techn. Rep. CMU-CS-04-162}, 2:99.

\bibitem[Yu et~al., 2023]{yu2023reinforcement}
Yu, C., Zheng, X., Zhuo, H.~H., Wan, H., and Luo, W. (2023).
\newblock Reinforcement learning with knowledge representation and reasoning: A brief survey.
\newblock {\em arXiv}.

\bibitem[Zambaldi et~al., 2019]{zambaldi2019deep}
Zambaldi, V., Raposo, D., Santoro, A., Bapst, V., Li, Y., Babuschkin, I., Tuyls, K., Reichert, D., Lillicrap, T., Lockhart, E., et~al. (2019).
\newblock Deep reinforcement learning with relational inductive biases.
\newblock In {\em ICLR}.

\bibitem[Zhuo and Yang, 2014]{zhuo2014action}
Zhuo, H.~H. and Yang, Q. (2014).
\newblock Action-model acquisition for planning via transfer learning.
\newblock {\em Artif. Intell.}, 212:80--103.

\end{thebibliography}

\end{document}